\documentclass{article}

\usepackage{microtype}
\usepackage{graphicx}
\usepackage{subfigure}
\usepackage{booktabs} 

\usepackage{hyperref}

\usepackage[accepted]{icml2024}

\usepackage{amsmath}
\usepackage{amssymb}
\usepackage{mathtools}
\usepackage{amsthm}

\usepackage[capitalize,noabbrev]{cleveref}

\theoremstyle{plain}

\theoremstyle{definition}

\theoremstyle{remark}

\usepackage[textsize=tiny]{todonotes}

\usepackage[utf8]{inputenc} 
\usepackage[T1]{fontenc}    
\usepackage{hyperref}       
\usepackage{url}            
\usepackage{booktabs}       
\usepackage{amsfonts}       
\usepackage{nicefrac}       
\usepackage{microtype}      
\usepackage{xcolor}         
\definecolor{red-figure2}{RGB}{151,40,30}
\definecolor{purple-figure2}{RGB}{93,57,150}
\usepackage{tikz}
\usetikzlibrary{bayesnet}
\usepackage{tabularx}
\usepackage{makecell}
\usepackage{multirow}

\usepackage{comment}
\usepackage{amssymb}
\usepackage[T1]{tipa}
\usepackage{tipx}
\usepackage{mathtools}
\usepackage{booktabs}
\usepackage{bbm}
\usepackage{amsmath}
\usepackage{graphicx}
\usepackage{listings}
\usepackage{float}
\usepackage{longtable}
\usepackage{soul}
\usepackage{xcolor}
\usepackage{caption}
\usepackage{algorithm}
\usepackage{algpseudocode}
\usepackage{threeparttable}

\definecolor{brickred}{rgb}{0.8, 0.25, 0.33}
\newcommand{\redbf}[1]{\bf{\textcolor{brickred}{#1}}}

\usepackage[english]{babel}

\let\emptyset\varnothing

\usepackage{mathtools}

\newcommand\iidsim{\stackrel{\mathclap{\normalfont\mbox{\tiny{i.i.d.}}}}{\,\,\sim\,\,}}
\definecolor{coopermj}{rgb}{0.0,0.6,0.0}

\icmltitlerunning{InterpreTabNet: Distilling Predictive Signals from Tabular Data by Salient Feature Interpretation}

\begin{document}

\twocolumn[
\icmltitle{InterpreTabNet: Distilling Predictive Signals from Tabular Data\\ by Salient Feature Interpretation}



\icmlsetsymbol{equal}{*}

\begin{icmlauthorlist}
\icmlauthor{Jacob Si}{uoft}
\icmlauthor{Wendy Yusi Cheng*}{uoft}
\icmlauthor{Michael Cooper*}{uoft,vector}
\icmlauthor{Rahul G. Krishnan}{uoft,vector}
\end{icmlauthorlist}

\icmlaffiliation{uoft}{University of Toronto}
\icmlaffiliation{vector}{Vector Institute}

\icmlcorrespondingauthor{Jacob Si}{jacobyhsi@cs.toronto.edu}

\icmlkeywords{Machine Learning, ICML}

\vskip 0.3in
]



\printAffiliationsAndNotice{\icmlEqualContribution} 

\begin{abstract}
  Tabular data are omnipresent in various sectors of industries. Neural networks for tabular data such as TabNet have been proposed to make predictions while leveraging the attention mechanism for interpretability. However, the inferred attention masks are often dense, making it challenging to come up with rationales about the predictive signal. To remedy this, we propose InterpreTabNet, a variant of the TabNet model that models the attention mechanism as a latent variable sampled from a Gumbel-Softmax distribution. This enables us to regularize the model to learn distinct concepts in the attention masks via a KL Divergence regularizer. It prevents overlapping feature selection by promoting sparsity which maximizes the model's efficacy and improves interpretability to determine the important features when predicting the outcome. To assist in the interpretation of feature interdependencies from our model, we employ a large language model (GPT-4) and use prompt engineering to map from the learned feature mask onto natural language text describing the learned signal. Through comprehensive experiments on real-world datasets, we demonstrate that InterpreTabNet outperforms previous methods for interpreting tabular data while attaining competitive accuracy.
\end{abstract}

\section{Introduction}

Machine learning methods for tabular data enjoy broad applications in diverse settings like healthcare \citep{diabetes}, insurance \citep{insurance}, and finance \citep{bank}. While predictive performance is key in these settings, practitioners often aim to translate predictive models into intelligible insights. For example, a medical practitioner working with tabular Electronic Health Records may be interested in determining features that contribute to a patient's diagnosis. Furthermore, an insurance underwriter working with tabular client data focuses on determining crucial factors that influence a client's risk profile. 

Despite commendable advancements made by existing models such as TabNet \citep{arik_pfister_2020}, there remains a discernible gap in achieving an integration of accuracy and interpretability. TabNet's ability to generate learnable masks for salient feature interpretation is limited as its interpretation is ambiguous. The considerable overlap between multiple masks makes it challenging for a user to discern the salient features used by the model for reasoning at each decision step. Other means of interpreting tabular models, such as attention weights \citep{attention} and SHAP values \citep{shap} have been criticized for their inconsistency in providing meaningful insights \citep{shap_disadvantage} and the computational intensity required to apply them to complex datasets \citep{attention_disad}. Additionally, tree-boosting methods such as XGBoost \citep{xgboost} and LightGBM \citep{lightgbm} exhibited limitations in their interpretability aspects when analyzed through the lens of SHAP values. These models tend to distribute the prediction contribution across an extensive range of features \citep{shap_dis}, leading to a less sparse representation of feature importance, making it difficult to identify important features.

The objective of our work is to distill the predictive signals from tabular data by enhancing the interpretability of the established TabNet architecture while maintaining competitive accuracy on practical datasets. To do so, we introduce InterpreTabNet, a modified variant of the TabNet neural architecture, enabling us to sparsify the identity of the predictive signals. Our work is premised on the hypothesis that we can map the predictive signals from the TabNet model onto a collection of sparse attribution masks that encode instance-wise feature significance. The sparsity of our masks leads to quick and easy identification of the salient features in the data. Having achieved this, we then enable post-hoc, text-based interpretability, using large language models (LLMs) \citep{gpt4} to draw upon rich prior knowledge related to the application domain \citep{choi2022lmpriors} and provide textual summaries of our sparse masks.

\begin{figure*}[htbp]
    \centering
    \includegraphics[width=0.85\textwidth, height=5.5cm]{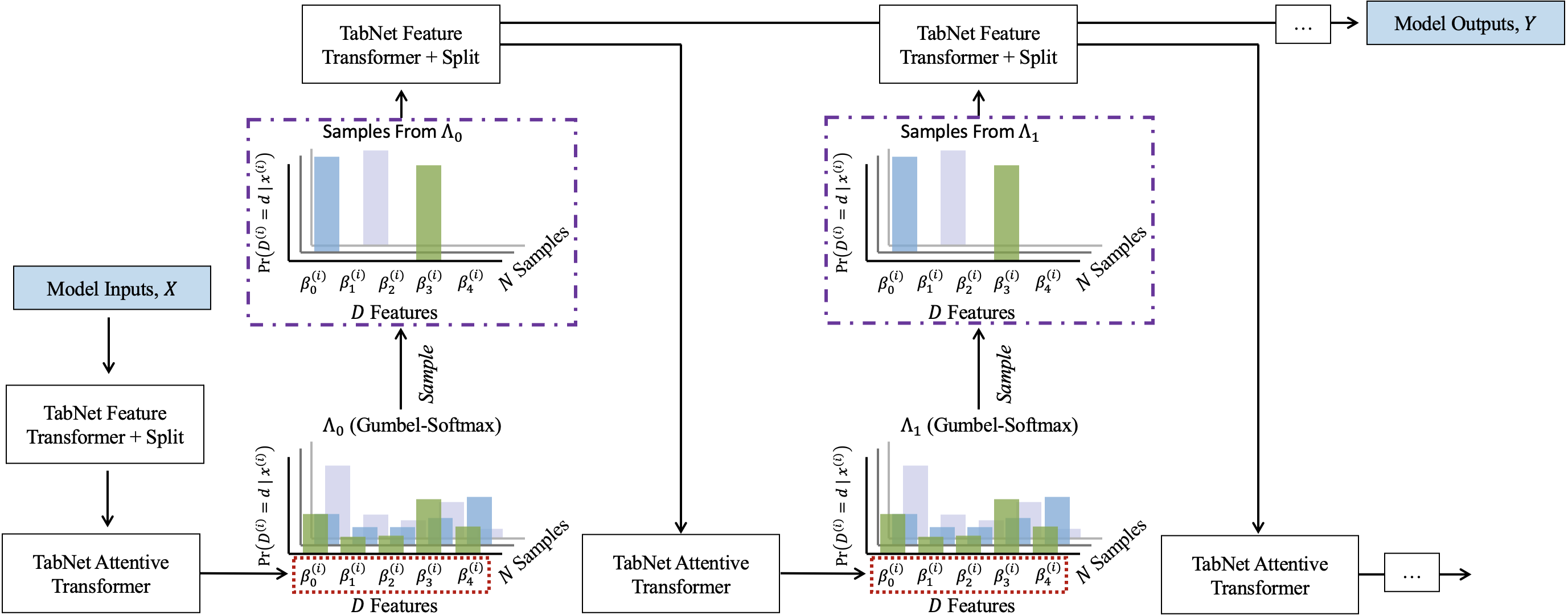}
    \caption{The InterpreTabNet architecture presents a variational formulation of the TabNet encoder. In our formulation, the weights of the attention masks produced by the TabNet encoder at each step $k$ are treated as the parameters, $\beta_0^{(i)}, ..., \beta_{D-1}^{(i)}$, of a Gumbel-Softmax distribution, $\Lambda_k$, unique to each instance (shown by the \textcolor{red-figure2}{red dotted rectangle}). This distribution is then sampled to produce a single feature that is highlighted for each feature at each step (\textcolor{purple-figure2}{purple dot-dashed rectangle}). This figure shows $k=2$ steps of the encoder architecture, over $D=5$ features, for $N=3$ samples.}
    \label{fig:architecture}
    \vspace{-2mm}
\end{figure*}

Our work makes the following contributions:
\vspace{-2mm}
\begin{enumerate}
    \item We devise \emph{a regularization scheme that maximizes diversity between masks in the TabNet architecture}. This is in contrast to the default "sparsity regularizer" employed by TabNet \citep{entropy}; its reliance on entropy often leads to the reuse of features across attention masks within the architecture. Empirically, under our regularization scheme, the model learns to extract salient features and generate sparse masks, reducing these challenges implicit in interpreting the feature masks generated by TabNet. Furthermore, our method suffers from only a modest tradeoff between accuracy and interpretability: we find that our approach performs comparably to the other baselines in accuracy but outperforms them considerably in terms of interpretability.
    \item Our regularization scheme relies on maximizing the KL divergence \citep{kullback1951information} between the distributions from which each TabNet attention mask is implicitly sampled. Whereas the original TabNet formulation does not explicitly characterize these distributions, we leverage tools from variational inference to \emph{model the attention weights within TabNet as samples drawn from a Gumbel-Softmax distribution}. By reformulating the attention weights within TabNet as a latent variable model, we can directly control properties of the mask distributions (such as the KL divergence) using regularized gradient-based optimization. Our architecture can be found in \cref{fig:architecture}.
    \item We show that by leveraging rich linguistic priors in a large language model we can capture the rich interdependencies between features that are needed to interpret model predictions in complex settings. We demonstrate how language models can relate the learned feature masks in our model to form detailed hypotheses about what is being learned at each step of the TabNet decision-making pipeline.
\end{enumerate}

\section{Related Works}

{\bf Learning from Tabular Data}. Early works on deep learning architecture for tabular data, such as TabNet, use a sequential attention mechanism for tabular data analysis \citep{arik_pfister_2020}. Their prominent strength is the capability to outperform other neural networks and decision trees on tabular datasets while yielding some level of interpretability for feature selections. However, TabNet's self-attention transformers' inability to capture diversifying latent variables leads to suboptimal feature selection. To address this limitation, diversity-promoting regularizers and latent models attempted to solve this problem \citep{xie_zhu_xing_2017} \citep{pmlr-v48-xiea16}. Subsequent works on tabular data include Net-DNF \citep{katzir2020net}, SubTab \citep{ucar2021subtab}, and TabTransformer \citep{tabtransformer}. Net-DNF \citep{katzir2020net} introduced an inductive bias that aligns model structures with disjunctive normal form (DNF) and emphasizes localized decisions. SubTab \citep{ucar2021subtab} transformed tabular data into a multi-view representation learning task, enhancing latent representation. Furthermore, TabTransformer \citep{tabtransformer} is a deep tabular data modelling architecture built upon self-attention-based Transformers.

{\bf Latent Variable Models}. Latent variable models like VAEs \citep{vae} and their variations demonstrate attractive abilities to model complex distributions and produce latent values. DirVAE has more interpretable latent values with no collapsing issues \citep{joo_lee_park_moon_2019}, while the cVAE \citep{kristiadi} models random latent variables and observed data, which gains control of the data generation process on the VAE. Additionally, the cVAE generates diverse but realistic output representations using stochastic inference \citep{sohn_lee_yan_1970}. Transformer-based cVAE exhibits excellent representation learning capability and controllability \citep{fang_zeng_liu_bo_dong_chen_2021}. We draw inspiration from these VAE extensions and incorporate the cVAE into TabNet's architecture to capture and reconstruct discrete data. Recent work in approximate inference for categorical data includes Categorical Reparameterization with Gumbel-Softmax \citep{jang2016categorical}. In our paper, we leverage the Gumbel-Softmax distribution as a key component of our methodology to strike a balance between interpretability and performance.

{\bf Model Interpretability.} Methods from interpretability aim to surface information about \textit{why} a machine learning model is making certain predictions to the user. Broadly, there are two families of methods in model interpretability. \textit{Intrinsic interpretability} refers to the scenario in which the user can directly leverage the parameters learned by the model to understand the rationale underlying the predictions. Linear models \citep{gauss1877theoria}, decision trees, Transformers (by means of their learned attention weights), and TabNet \citep{arik_pfister_2020}, are all, to varying degrees, intrinsincally interpretable methods. In contrast, methods from \textit{post-hoc interpretability} tackle the scenario in which the model may be black-box: these methods instead attempt to approximate the decision-making process underlying the model, which is then surfaced to the user. Methods like SHAP \citep{shap}, LIME \citep{lime}, and Grad-CAM \citep{selvaraju2017grad} are methods for post-hoc interpretability. The central tradeoff between intrinsic and post-hoc interpretability is this: while an intrinsically interpretable model is (definitionally) faithful to its underlying decision rule, it may be necessary to make simplifying assumptions in the design of the model. Conversely, while post-hoc interpretability methods can interpret models of arbitrary complexity, the interpretable decision rule surfaced by such procedures is only an approximate one \citep{du2019techniques}. Our approach draws upon insights from both classes of methods: we leverage tools from variational inference to improve upon the intrinsic interpretability of TabNet, and we employ a large language model to provide a richer contextual interpretation of the learned features post-hoc.

\section{The InterpreTabNet Model} 

Let $(X, Y) \iidsim \mathcal{X} \times \mathcal{Y}$ represent the covariates and an outcome that we want to model, respectively. As we are operating in the tabular data regime, assume that $X \in \mathbb{R}^{N \times D}$, where each $d \in [1, ..., D]$ corresponds to a single discrete feature in the data. Then, each $x^{(i)}, y^{(i)}$ represents $D$-vector and label corresponding to a particular example.\footnote{Unless otherwise stated, our notation uses uppercase letters to refer to distribution-level quantities, such as the distribution over the covariates, and lowercase letters to refer to specific samples drawn from those distributions.} Let $P(\cdot | \cdot)$ denote true probability density functions, and $Q(\cdot | \cdot)$ denote variational approximations of those densities.

\begin{figure*}[htb!]
    \centering
    \begin{minipage}[b]{0.47\textwidth} 
        \centering
        \includegraphics[width=\textwidth]{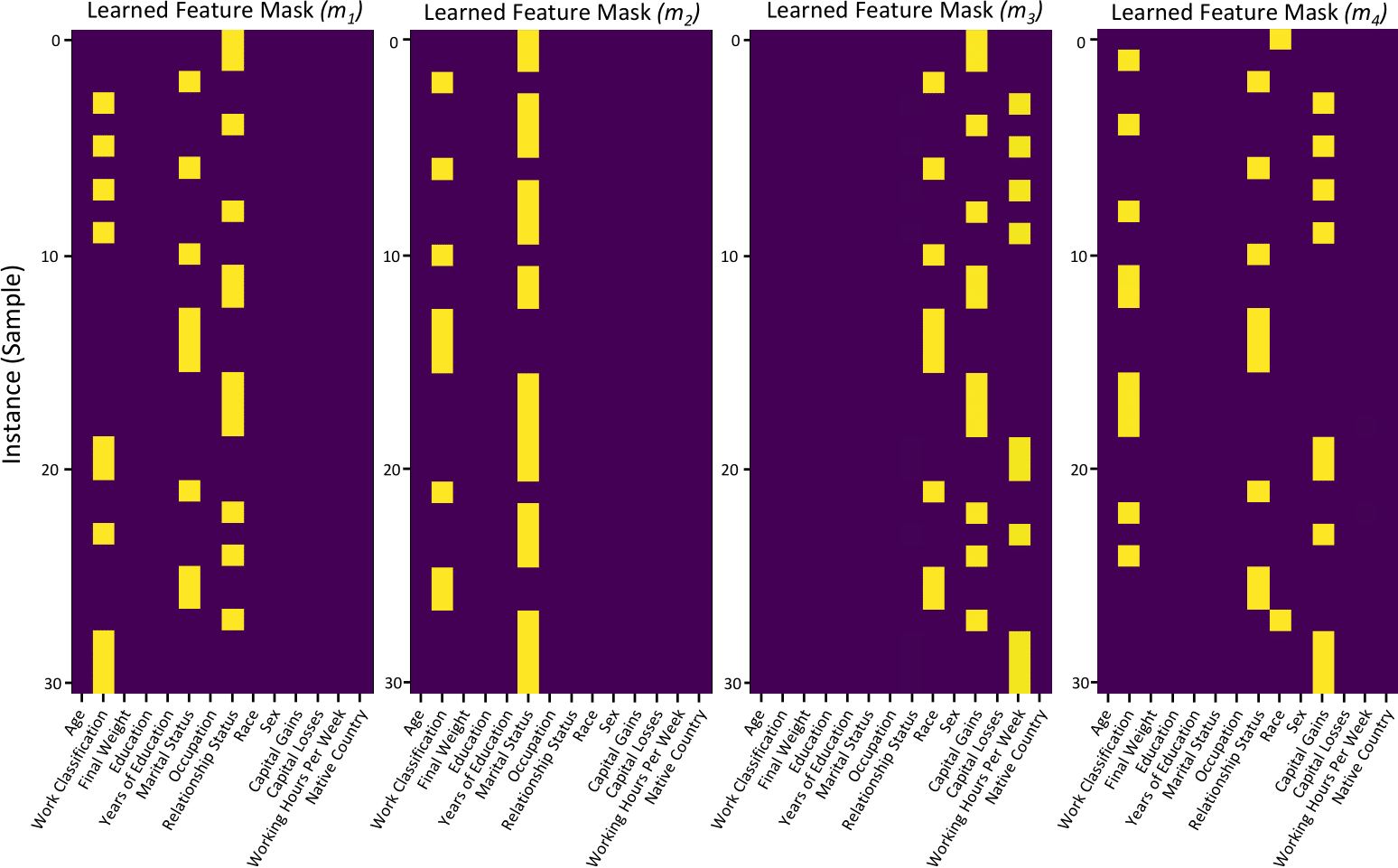}
        \caption*{(a) InterpreTabNet Feature Mask ($r_M^* = 9$)} 
    \end{minipage}
    \hspace{0.2cm} 
    \begin{minipage}[b]{0.47\textwidth} 
        \centering
        \includegraphics[width=\textwidth]{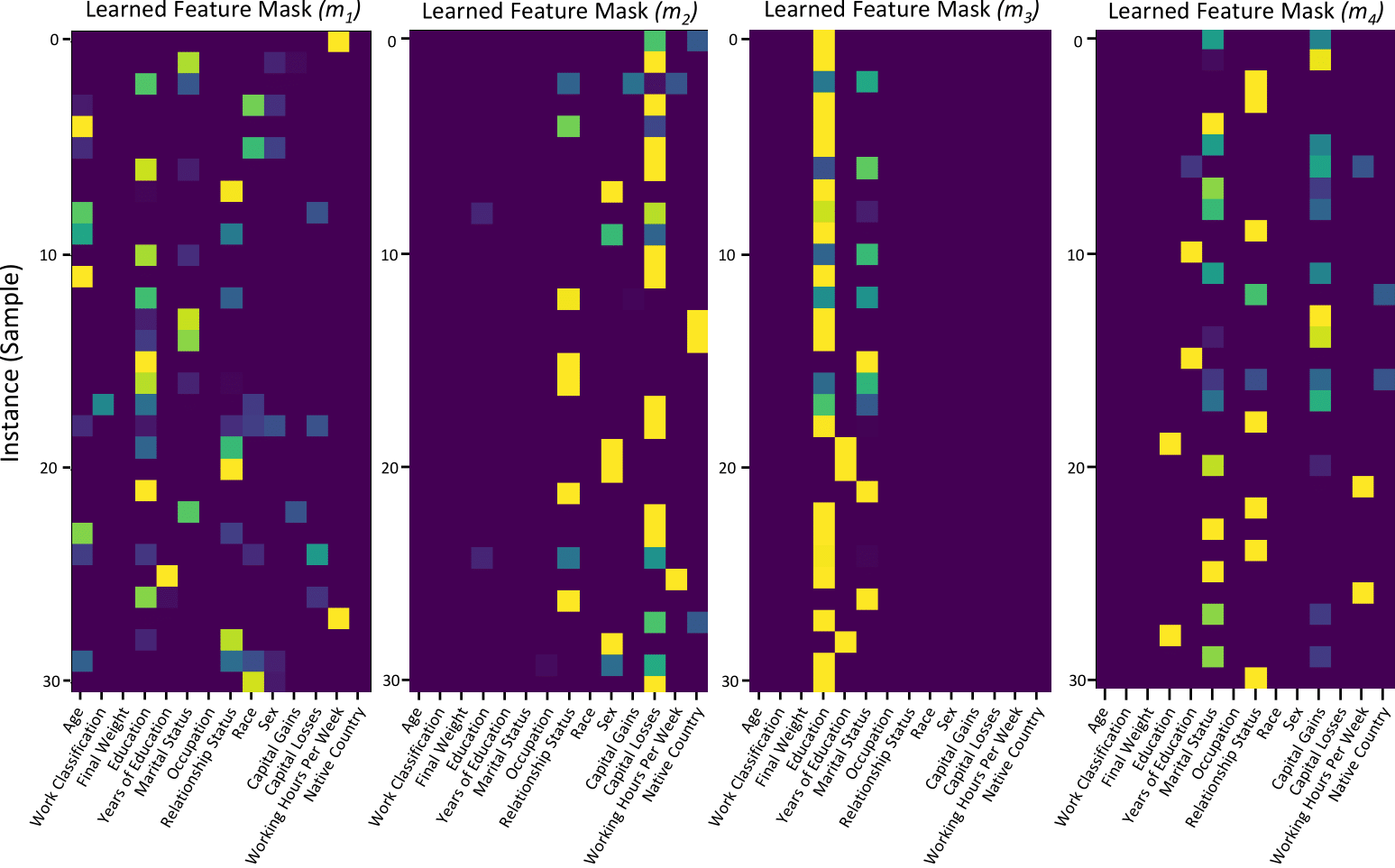}
        \caption*{(b) Original TabNet Feature Mask} 
    \end{minipage}
    \begin{minipage}{0.47\textwidth} 
        \centering
        \includegraphics[width=\textwidth]{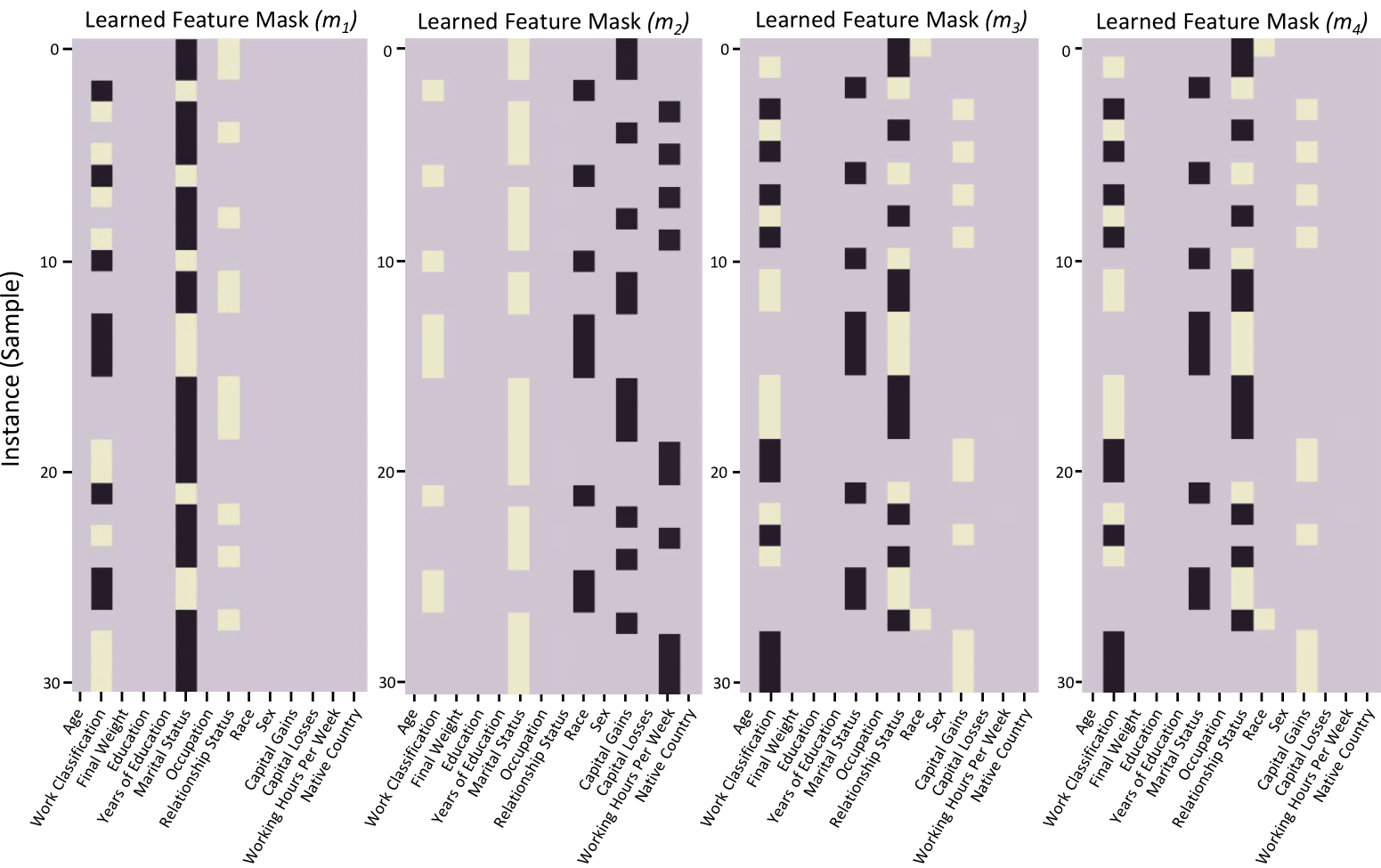}
        \caption*{(c) Stacked InterpreTabNet Feature Masks} %
    \end{minipage}
    
    \caption{X/Y-axis labels denote the features and test samples for each respective mask at $N_{steps} = 4$ decision steps of the Adult Census Income dataset. \textbf{Left (a)}: Learned masks associated with InterpreTabNet. Observe how for each example, there is no overlap in the attention learned across different masks with high salience. This mutual exclusivity of attention across masks makes for easier visual interpretation of the learned signal that InterpreTabNet leverages in its predictions. \textbf{Right (b)}: Learned masks associated with TabNet. Observe how, for each example, there exist overlaps in the attention learned for each mask with no clear salience. This makes the masks challenging to interpret, as there is no obvious way to reconcile attention that is distributed across multiple masks in this manner. \textbf{Bottom (c)}: Stacked InterpreTabNet Feature Masks between subsequent feature masks (Left to Right: Masks 0 \& 1, 1 \& 2, 2 \& 3, 3 \& 0) outlining no overlap and sparsity in feature selection. More details can be found in Section \ref{qual_anal}.}
    \label{fig:mask_comparison}
    \vspace{-2mm}
\end{figure*}

\subsection{High-Level Approach}

The TabNet encoder architecture models the predictive signal, $P(y\,|\,x)$, as a nonlinear combination of the covariates, $x$, and a sequence of $k$ {\it learned attention masks}. Each feature importance mask, $m_k$, depicts the feature selected at the $k$-th decision step. We learn each mask by applying the TabNet Transformer in the encoder to the covariates and previous attention mask at each step of a multi-step decision process. Since the nonlinear combination is modeled using a multi-layer perceptron \citep{mlp}, inference within TabNet's encoder can be expressed as:

{\small
\begin{equation}
    \Pr(y \,|\, x) = f_{\psi}^{\text{(MLP)}}\left(\sum_{k=0}^{K-1} f_{\psi}^{\text{(TabNet\_Transformer)}}(m_{k}, x) \right),
\end{equation}
}

where $m_{k} = \emptyset$ if $k = 0$, and where $\psi$ is a general-purpose variable to denote the parameters that are associated with a given MLP or TabNet Transformer sub-model. Our goal is to construct a version of this model wherein each mask is a latent variable in a deep generative model. Then we can learn the model via amortized variational inference by inferring $m_{k}$ using some parametric distribution $Q$ that admits backpropagation by means of the reparameterization trick. By specifying the form of the distribution mask samples are from, we can directly adjust the properties of this latent variable by regularizing the loss function. Specifically, as our objective is to promote sparsity among the masks, we will then aim to maximize the KL divergence between subsequent masks of the decision steps.

In the following sections, we demonstrate how we sample the masks in our architecture from a Gumbel-Softmax distribution \citep{jang2016categorical}. We choose Gumbel-Softmax as a natural sampling distribution for the masks because the salience of a feature can be treated as a categorical variable: for each example $i$ in mask $k$, a feature $j$ can either be "selected" ($m_{k_{ij}}=1$), or "not selected" ($m_{k_{ij}}=0$). The Gumbel-Softmax distribution offers a continuous relaxation of a categorical distribution, thus facilitating the application of the reparameterization trick under our method.

\subsection{Mask Sampling Process}
The mask sampling process for InterpreTabNet is the following, where $Y$ represents the predicted outcome, $z$ represents the concatenation of all the $m_k$ mask samples from a Gumbel-Softmax distribution, $X$ represents the data, and $\Lambda$ represents a Gumbel-Softmax distribution.

\vspace{-6mm}
{\small
\begin{align*}
    P(m_k|X) &\sim \Lambda_k (\text{TabNet\_Transformer}(X)), \\
    & \quad \text{for } k = 0 \nonumber \\
    P(m_{k+1}|m_k, X) &\sim \Lambda_k (\text{TabNet\_Transformer}(m_k, X)), \\
    & \quad \text{for } k \in [1, \ldots, K-1]. \nonumber
\end{align*}
}
\vspace{-6mm}

Unlike TabNet, InterpreTabNet does not only leverage its feature importance masks to make predictions; instead, the feature importance masks serve as the emissions of a stochastic process that we regularize in order to promote sparsity. InterpreTabNet utilizes these masks from the first iteration onwards (after the zeroth iteration) as latent variables. These latent variables serve as a rich source of embedded knowledge, allowing the model to improve its generalizations by acting as a stochastic process. Furthermore, sampling this latent variable from the Gumbel-Softmax distribution will act as a crucial component in improving interpretability (details explored in Section \ref{sparsity_reg}).

Let us represent the collection of all $k$ masks, $[m_0, ..., m_{k-1}]$ as a single latent variable, $z \in \mathbb{R}^{N \times k}$, drawn from a Gumbel-Softmax distribution. Drawing samples $z$ from a categorical distribution with class probabilities $\pi$ is as follows.

\vspace{-5mm}
{\small
\begin{equation*}
z = \mathrm{one\_hot}\left(\underset{i}{\arg\max} (\beta_i + \log \pi_i) \right)
\end{equation*}
}
\vspace{-5mm}

where $\beta_0, ..., \beta_{D-1}$ are i.i.d samples drawn from a standard Gumbel distribution, Gumbel($0$,$1$). As a small technical note, the original TabNet architecture requires a ReLU function to be applied to the embeddings between blocks. Our sampling scheme ensures nonnegative mask values, so this requirement is not necessary in our architecture.

The mask sampling process is characterized as a latent variable problem. Thus, this necessitates the implementation of inference techniques for effective learning. 

\subsection{Generating Predictions with the Conditional Variational Autoencoder}

We interpret TabNet's encoder-decoder architecture as a conditional variational autoencoder (cVAE) \citep{vae, blei2017variational}. We imagine an encoder conditioned on two variables, $Y$ and $X$, which leverages the distribution $Q(z|Y,X)$ to sample the feature masks, $z$. Similarly, we imagine a decoder that conditions on the feature masks, $z$, and the data $X$, to predict a corresponding label drawn from $P(Y|z, X)$. Using this framework, we can derive a variational lower bound on this cVAE. We do so by modelling the outcome, $P(Y|X)$ as $\int{P(Y|X, z)P(z|X)dz}$, and inferring $P(z)$ through $P(z|Y)$ using $Q(z|Y)$. See \cref{graphical_model} for the graphical model. The derivation can be found in Appendix \ref{appendix_cvae}.
\begin{multline}
    \log P(Y|X) - D_{KL}[Q(z|Y, X)||P(z|Y, X)] \\
    = E[\log P(Y|z, X)] - D_{KL}[Q(z|Y, X)||P(z|X)]
\end{multline}

\begin{figure}[htbp]
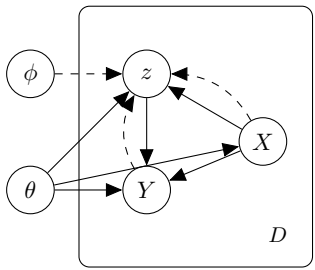

    \centering
    \tikz[scale=0.9, transform shape]{
    \node[latent] (z) {$z$} ;
    \node[latent, below=of z] (Y) {$Y$} ;
    \node[latent, right=of z, yshift=-1cm] (X) {$X$} ;
    \node[latent, left=of z] (phi) {$\phi$} ;
    \node[latent, left=of Y] (theta) {$\theta$} ;
    \plate[inner sep=0.45cm, xshift=-0.12cm, yshift=0.12cm] {plate1} {(z) (Y) (X)} {$D$};
    \edge {z} {Y} ;
    \edge {X} {Y} ;
    \edge {X} {z} ;
    \edge {theta} {z} ;
    \edge {theta} {X} ;
    \edge {theta} {Y} ;
    \path [draw,->,dashed] (X) [bend right] edge (z) ;
    \path [draw,->,dashed] (phi) edge (z) ;
    \path [draw,->,dashed] (Y) [bend left] edge (z) ;
    }
    \caption{Graphical model of InterpreTabNet with $D$ i.i.d samples. Solid lines denote the generative model $p_{\theta}(Y|z, X)p_{\theta}(z|X)$, dashed lines denote the variational approximation $q_{\phi}(z|X, Y)$ to the intractable posterior $p_{\theta}(z|X, Y)$. The variational parameters $\phi$ are learned jointly with the generative model parameters $\theta$.}
    \label{graphical_model}
\end{figure}

\subsection{Sparsity-Promoting  Regularization}
\label{sparsity_reg}

Our formulation of TabNet as a stochastic cVAE allows us to directly promote mask sparsity by using the loss function to encourage variation in the Gumbel-Softmax distributions corresponding to adjacent masks. To do so, we incorporate a KL Divergence Sparsity Regularizer ($r_M$) in the model architecture. With the KL Divergence, we aim to maximize the difference between the distribution of masks that are subsequent to one another. This would reduce the number of selected features, ensuring that the features selected are independent between masks. Additionally, with a sparser feature selection, the model can focus on fewer high-salience features. Therefore, the ELBO of the InterpreTabNet model is as follows with $r_M$ as a tunable regularizer weight.

\vspace{-6mm}
{\small
\begin{multline}
    E[\log P(Y|z, X)] - \sum_{i} D_{KL}\bigg(\big(Q\left(z_i | Y, X\right)\big)\,\,\bigg|\bigg|\,\,\big(P\left(z_i | X\right)\big)\bigg) \\
    + r_{M} \cdot \sum_{i \neq j} D_{KL} \bigg(\big(Q\left(z_i | Y, X\right)\big)\,\,\bigg|\bigg|\,\,\big(Q\left(z_j | Y, X\right)\big)\bigg)
\end{multline}
}
\vspace{-6mm}

\subsection{Sparsity Regularizer ($r_M$) Algorithm}
\label{reg_algo}

To assess the level of interpretability a feature mask provides, we divide it into two sets of criteria. Note that there are more criteria than those we enumerate here. Those enumerated here are exemplars to aid understanding.
\begin{enumerate}
    \item Number of selected features (e.g., number of important features must be at least 2-3).
    \item "Salience" of each feature (e.g., percentage of importance captured by one feature in each mask must be between 20 and 25\%).
\end{enumerate}

Within a feature mask, we would like to swiftly identify the salient features that contribute to its prediction. Thus, our aim is to \emph{minimize the number of selected features}, and only select those of \emph{high salience}, while maintaining a competitive accuracy. This would yield an interpretable mask to determine the important features.

We propose an adaptive algorithm to optimize our KL Divergence Sparsity Regularizer, $r_M$, to improve the interpretability of the feature masks. Our method involves iterative training and evaluation of the InterpreTabNet model with varying values of $r_M$ within a pre-defined range, to check the fulfilment of the above criteria. The end result is the optimal $r_M$ value corresponding to a balance between an interpretable feature mask and classification accuracy, improving the overall efficacy of our model. The algorithm and the full set of criteria can be found in Appendix \ref{appendix_algorithm}.

\section{Experiments and Discussions}
We evaluate the performance of InterpreTabNet on real-world classification tasks both quantitatively and qualitatively. Our analysis sections (Sections \ref{quant_analysis} and \ref{qual_anal}) are based on the Adult Census Income \citep{adult} dataset for simplicity purposes. See Appendix \ref{appendix_exp_other_datasets} for full results from the other datasets.\footnote{The code is available on GitHub at:\\ \url{https://github.com/jacobyhsi/InterpreTabNet}}

\begin{table*}[!t]
\centering
\caption{Test Accuracy Scores (mean and standard deviations across 20 random seed trails in \%) across Different Models and Datasets with Optimal Mask Regularizer Values ($r_M$) for InterpreTabNet. InterpreTabNet achieves substantial improvements in interpretability across all the datasets and remains competitive in terms of accuracy in most datasets.}
\label{tab:accuracy_scores_reg}
\footnotesize 
\begin{tabular}{lccccccc}
\toprule
Model / Dataset & Adult Census & Forest Cover & Poker Hand & Mushroom & Blastchar & Diabetes & Higgs \\
\midrule
InterpreTabNet & \textbf{87.42 $\pm$ 0.55} & \textbf{94.75 $\pm$ 0.53} & 99.50 $\pm$ 0.48 & 96.62 $\pm$ 0.35 & 72.96 $\pm$ 0.56 & 55.37 $\pm$ 0.47 & 53.08 $\pm$ 0.56 \\
Original TabNet & 85.55 $\pm$ 0.56 & 94.18 $\pm$ 0.63 & 99.00 $\pm$ 0.62 & 99.94 $\pm$ 0.31 & 76.22 $\pm$ 0.34 & 56.91 $\pm$ 0.53 & 52.94 $\pm$ 0.45\\
XGBoost & 86.60 $\pm$ 0.64 & 92.30 $\pm$ 0.62 & 75.57 $\pm$ 0.47 & 99.69 $\pm$ 0.39 & 77.29 $\pm$ 0.53 & \textbf{61.44 $\pm$ 0.32} & \textbf{72.70 $\pm$ 0.35} \\
LightGBM & 86.20 $\pm$ 0.43 & 86.38 $\pm$ 0.64 & 78.47 $\pm$ 0.36 & \textbf{100.00 $\pm$ 0.37} & \textbf{77.86 $\pm$ 0.46} & 60.87 $\pm$ 0.39 & 72.62 $\pm$ 0.35 \\
TabTransformer & 85.09 $\pm$ 0.39 & 82.55 $\pm$ 0.39 & \textbf{99.81 $\pm$ 0.31} & \textbf{100.00 $\pm$ 0.57} & 73.17 $\pm$ 0.56 & 44.45 $\pm$ 0.34 & 51.97 $\pm$ 0.54 \\
MLP & 79.76 $\pm$ 0.65 & 84.89 $\pm$ 0.56 & 99.70 $\pm$ 0.56 & 99.82 $\pm$ 0.56 & 75.16 $\pm$ 0.61 & 53.99 $\pm$ 0.46 & 63.17 $\pm$ 0.36 \\
\bottomrule
\end{tabular}
\end{table*}

\textbf{Datasets}. The real-world tabular datasets we use in our experiments are from the UCI Machine Learning Repository \citep{uci_ml} and OpenML \citep{OpenML2013}. These datasets were selected since they were utilized to evaluate the existing methods (baselines). Additionally, they vary in size and nature, with both categorical and continuous features, to ensure a holistic evaluation of our methodology across multiple domains and scenarios. The training/validation/testing proportion of the datasets for each split is 80/10/10\% apart from the Higgs dataset. Due to the inherently large Higgs dataset, we adhere to TabNet's method of data splitting with 500k training samples, 100k validation samples, and 100k testing samples. Details of the datasets can be found in Appendix \ref{appendix_dataset}.

\textbf{Baselines: Accuracy}. We compare our model against five other ML methods for tabular classification. These include the Original TabNet, XGBoost \citep{xgboost}, LightGBM \citep{lightgbm}, TabTransformer \citep{tabtransformer}, and multi-layer perceptrons (MLP) \citep{mlp}. For each model, we utilize the recommended hyperparameters mentioned by the authors of their respective papers. Furthermore, we also conduct a grid search within the range of the recommended hyperparameters to optimize the models, selecting the best-performing hyperparameter configuration.

\textbf{Baselines: Interpretability}. We compare our model against four other ML methods to determine which model allows the user to easily determine the important features when predicting the outcome. These include the Original TabNet, XGBoost, LightGBM, and TabTransformer. We excluded MLPs as they perform notably worse than the other models in accuracy. The interpretability figures for InterpreTabNet, Original TabNet, XGBoost, and LightGBM are feature masks whereas TabTransformer uses an attention mask. In order to compare the interpretability of feature masks between InterpreTabNet, Original TabNet, XGBoost, and LightGBM, we conduct row-wise normalization on the absolute SHAP values from XGBoost and LightGBM. This yields the same feature importance scale (relative importance of each feature within each sample's prediction) as InterpreTabNet and Original TabNet.

\subsection{Quantitative Analysis}
\label{quant_analysis}

\textbf{Performance against Baselines.} The performance of our method relative to the baselines for tabular learning is shown in Table \ref{tab:accuracy_scores_reg}. We achieve the best performance in 2/7 datasets while maintaining a competitive accuracy for the remaining 5/7 datasets. Our most notable contribution is achieving a significant improvement in interpretability.

\textbf{Model Faithfulness (Quantitative).} We conducted synthetic dataset experiments to ensure that InterpreTabNet is faithful to its predictions. We used the same synthetic data generation model in TabNet \citep{arik_pfister_2020} and INVASE \citep{yoon2018invase}. InterpreTabNet outperforms existing methods in 4 out of 6 synthetic datasets, indicating that its predictions adhere to the ground truth quantitatively. See Table \ref{tbl:exp-syn} Appendix \ref{appen:syn_dataset} for more details.

\textbf{Computational Efficiency.} Our model necessitates an additional computation through the Gumbel-Softmax reparameterization and conditioning on the mask from the previous time step when compared to TabNet. Nonetheless, this extra step incurs a minimal cost, leading to a mere several-minute increase in training time. Furthermore, likewise to TabNet, our model maintains greater computational efficiency than other baseline models without necessitating an extensive search for fine-grained hyperparameters.

\subsection{Qualitative Analysis}
\label{qual_anal}

\begin{figure}[htb!]
    \centering
    \begin{minipage}[b]{0.325\columnwidth}
        \centering
        \includegraphics[width=0.73\columnwidth, height=5cm]{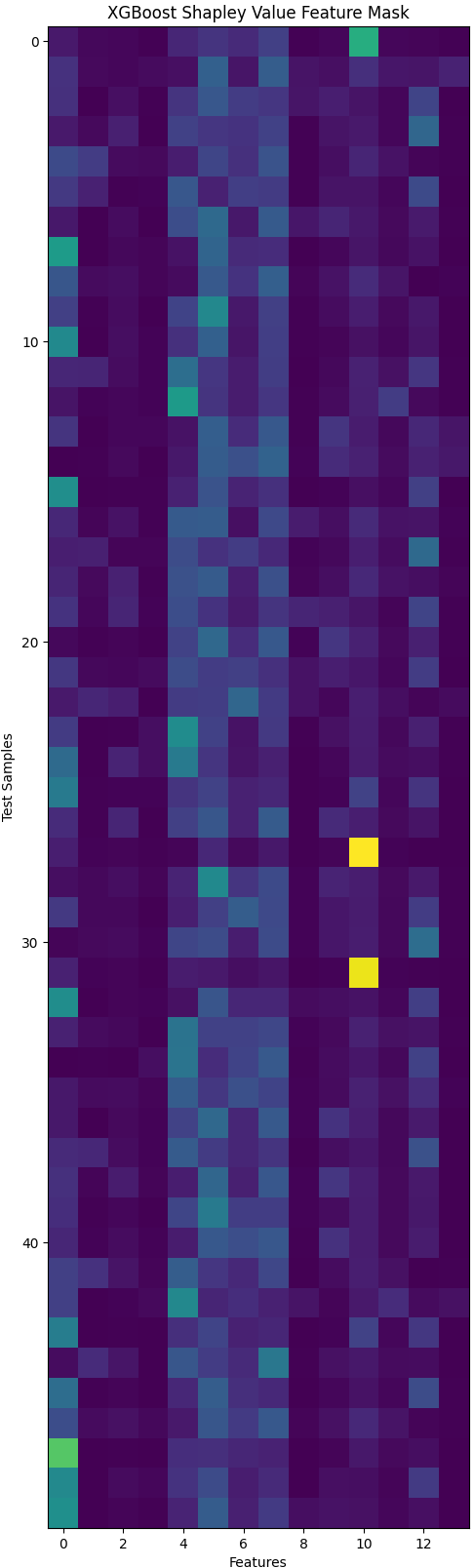}
        \caption*{(a) XGBoost\\ SHAP}
    \end{minipage}
    \begin{minipage}[b]{0.325\columnwidth}
        \centering
        \includegraphics[width=0.73\columnwidth, height=5cm]{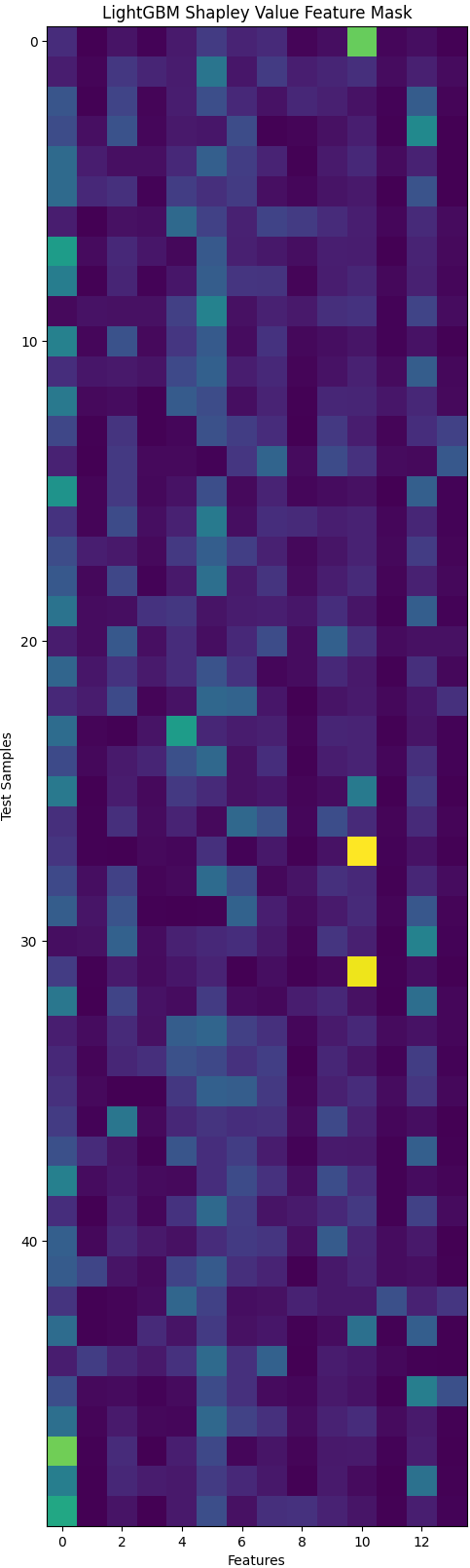}
        \caption*{(b) LightGBM\\ SHAP}
    \end{minipage}
    \begin{minipage}[b]{0.325\columnwidth}
        \centering
        \includegraphics[width=\textwidth]{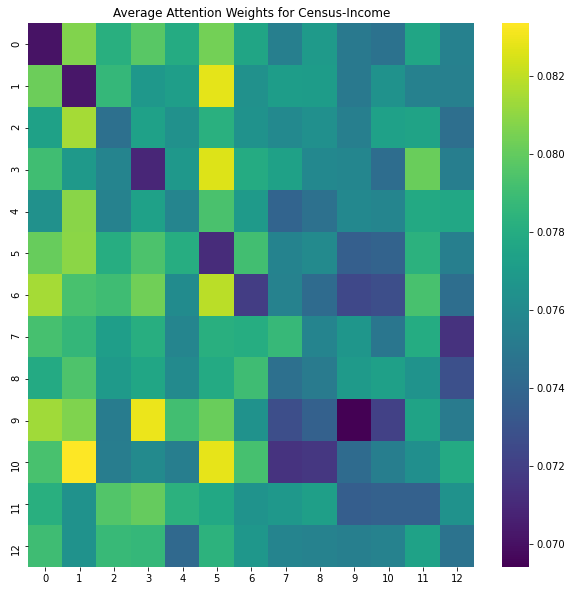}
        \caption*{(c) TabTrans.\\ Att. Wgts.}
        \label{fig:att_weights_income}
    \end{minipage}
    
    \caption{(a), (b), and (c) depicts the XGBoost SHAP Feature Mask, LightGBM SHAP Feature Mask, and TabTransformer Attention Weights for the Adult Census Income Dataset, respectively. X/Y-axis labels denote the features and test samples for the feature masks in (a) and (b) whereas only features for the attention weights in (c).}
    \label{interpretability_figs}
    \vspace{-2mm}
\end{figure}

\textbf{Interpretability Evaluation.} Figure \ref{fig:mask_comparison} highlights the learned masks associated with InterpreTabNet using a sparsity regularizer value of $r_M = 9$ compared to those of TabNet. The rows of each mask represent individual data samples, while the columns represent discrete features in the tabular data. Values of feature importance for each test sample/row sum up to $1$. Thus, bright yellow squares indicate values close to/equal to $1$, dark purple squares indicate values close to/equal to $0$, and other colored squares indicate values between $0$ and $1$. As observed in Figure \ref{fig:mask_comparison}, our InterpreTabNet model highlights mutually exclusive features of high importance that are easily interpretable. Practitioners can easily identify the salient features contributing to the outcome prediction. On the contrary, feature masks of the Original TabNet are more difficult to interpret since each mask highlights multiple features for a given data sample.

When comparing against the other baselines, InterpreTabNet leverages sequential decision-making that allows users to understand how the model's focus shifts and how different features interact over the decision process. Figure \ref{interpretability_figs} illustrates the complex pathways of model interpretation inherent in our baseline architectures like XGBoost, LightGBM, and TabTransformer. These models necessitate additional tools to render interpretative insights. Both XGBoost and LightGBM are augmented with SHAP values derived from external SHAP packages to achieve interpretability while TabTransformer relies on attention weights. The reliance on SHAP values in XGBoost and LightGBM found in Figures \ref{interpretability_figs}(a) and \ref{interpretability_figs}(b) distribute the contribution of the prediction across all features. This leads to a less sparse representation of feature importance. In practice, this means that while each feature's contribution to the prediction is identified, the significance of each feature is not as distinct. This results in an interpretation where barely any feature stands out, especially in models with a large number of features where many contributed incrementally to the final prediction. This lack of sparsity makes it challenging for practitioners to pinpoint a concise set of features for understanding and analysis. Note that we utilized feature masks to illustrate the SHAP values of XGBoost and LightGBM to ensure a consistent basis for comparison with InterpreTabNet. The attention mechanism of TabTransformer in Figure \ref{interpretability_figs}(c) provides a form of interpretability by capturing relationships between features. However, it is unable to pinpoint a set of important features. Furthermore, attention weights are typically dense, meaning that most features will get some level of attention.

\begin{figure}[htb!]
    \centering
    \begin{minipage}[b]{0.23\textwidth}
        \includegraphics[width=\textwidth]{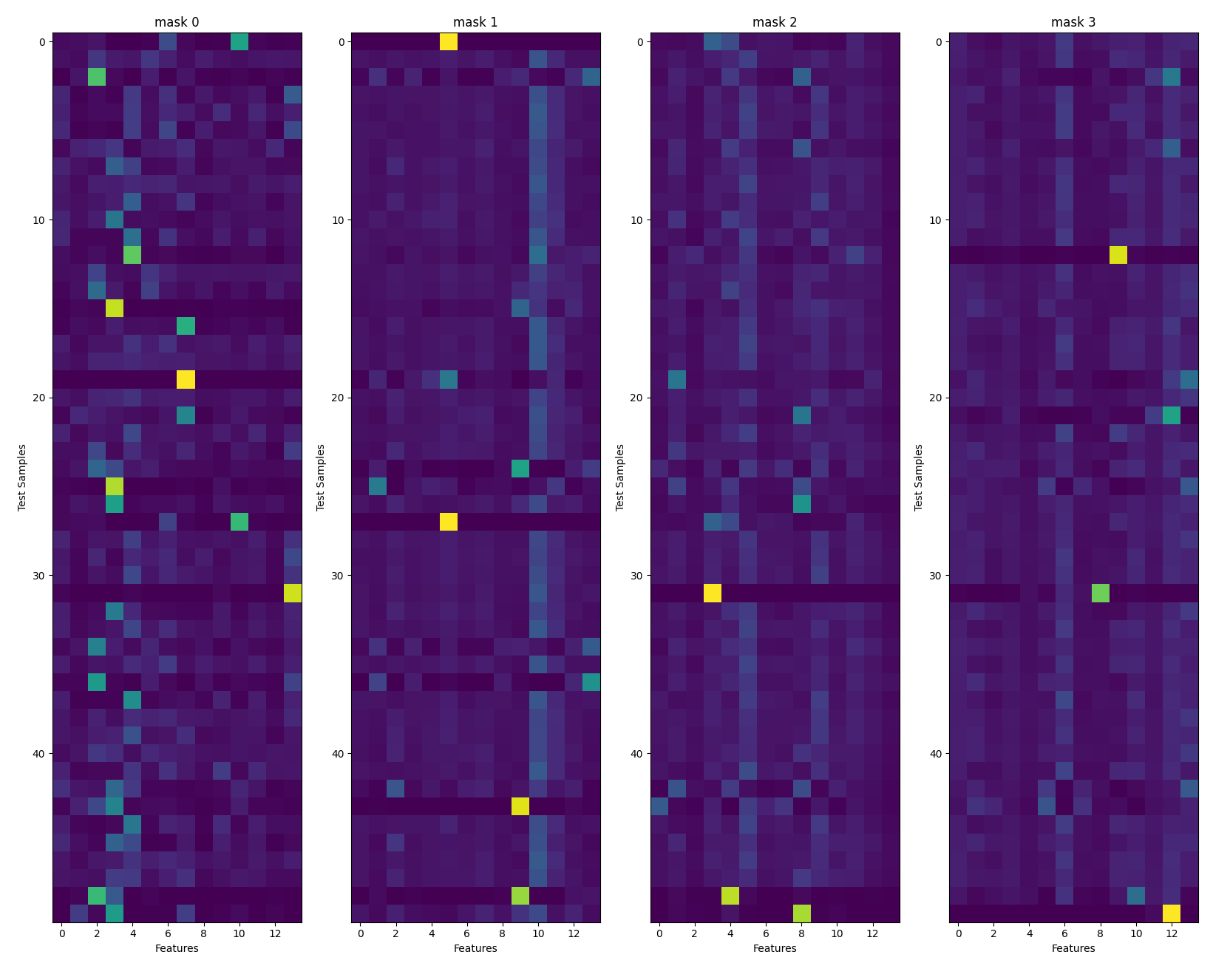}
        \caption*{(a) $r_M = 0$} 
    \end{minipage}
    \hfill
    \begin{minipage}[b]{0.23\textwidth}
        \includegraphics[width=\textwidth]{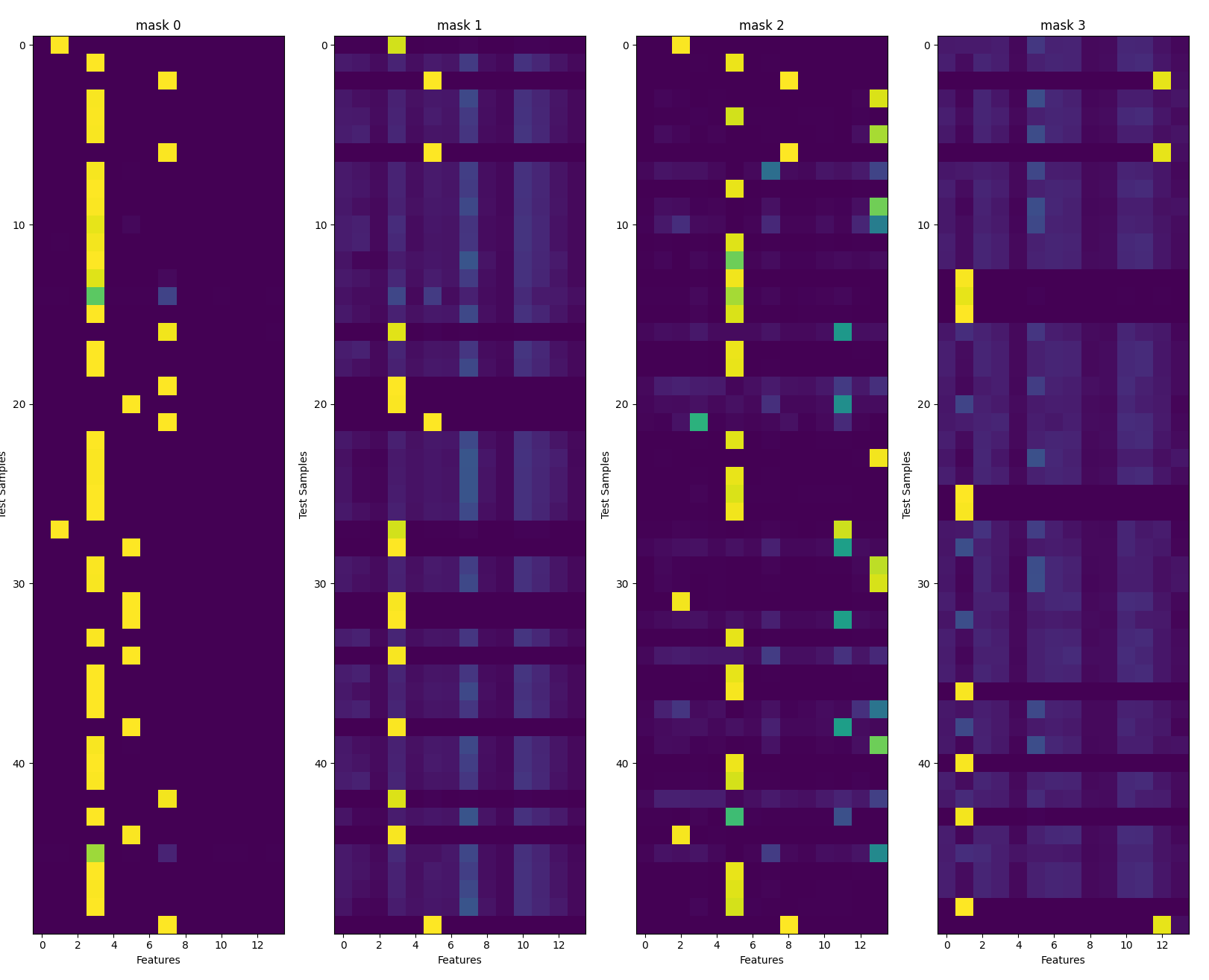}
        \caption*{(b) $r_M = 1$} 
    \end{minipage}

    \vspace{2mm}  
    
    \begin{minipage}[b]{0.23\textwidth}
        \includegraphics[width=\textwidth]{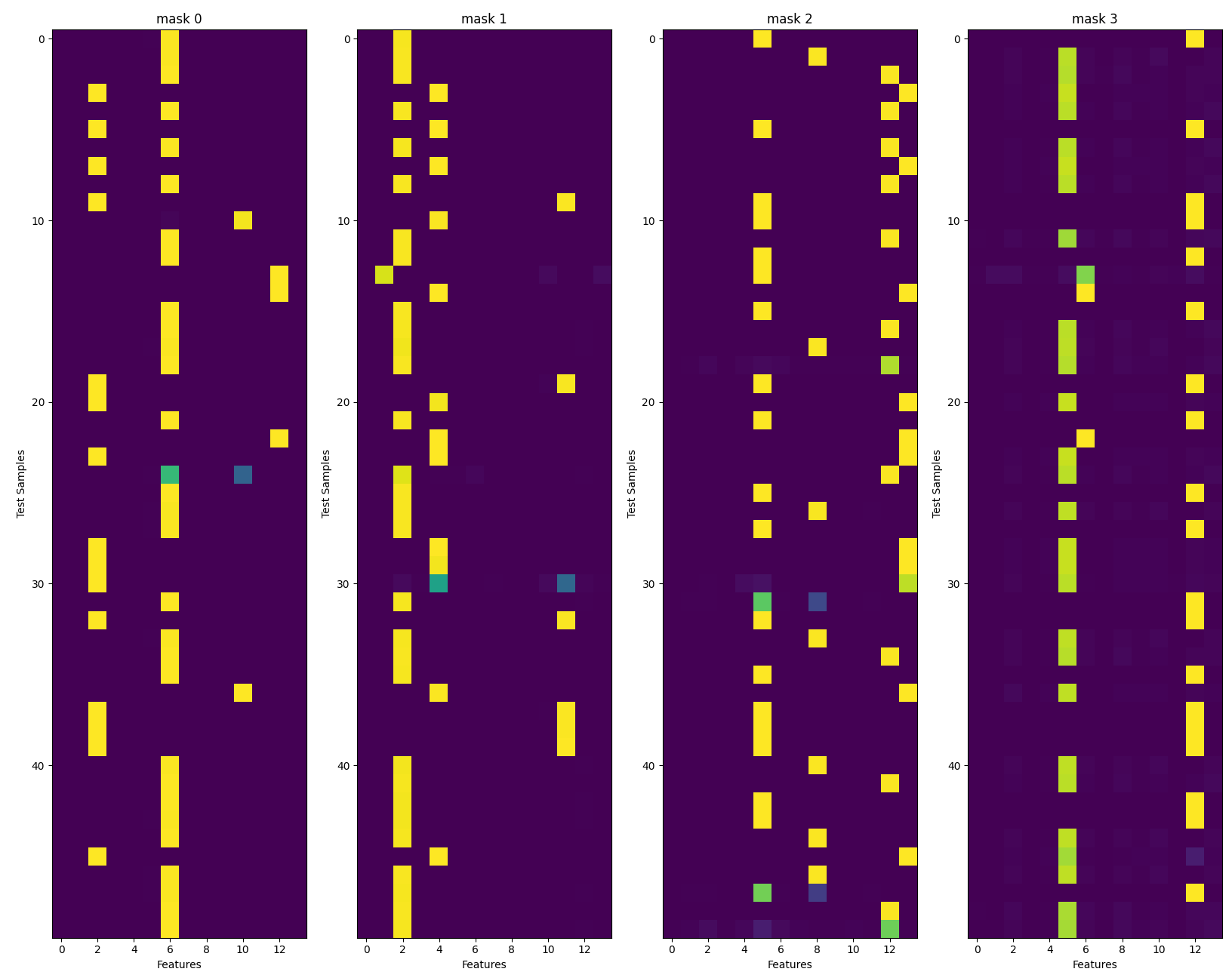}
        \caption*{(c) $r_M = 100$} 
    \end{minipage}
    \hfill
    \begin{minipage}[b]{0.23\textwidth}
        \includegraphics[width=\textwidth]{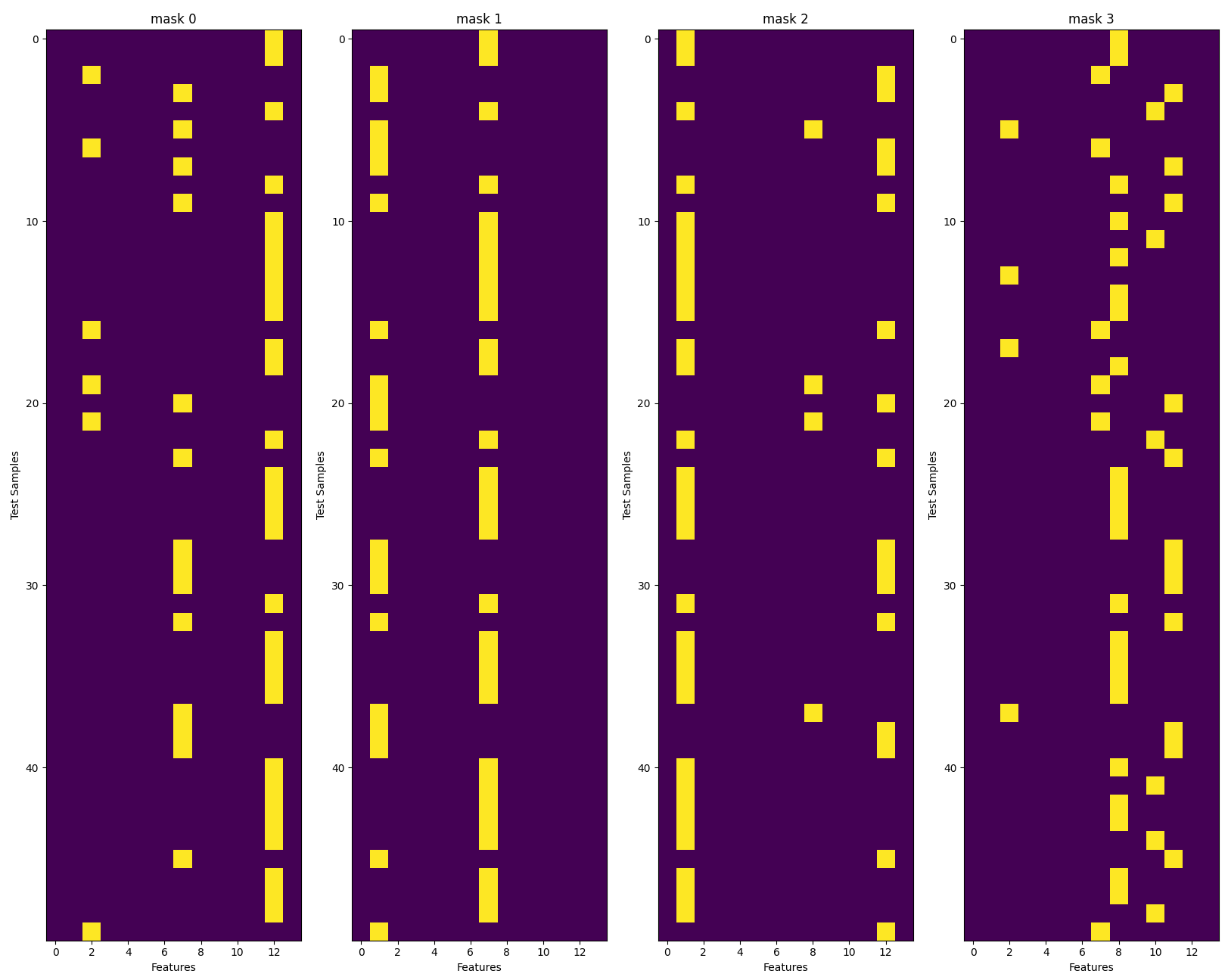}
        \caption*{(d) $r_M^* = 1000$} 
    \end{minipage}
    
    \caption{InterpreTabNet Sparsity Regularizer $r_M$ Feature Mask Ablation. As the $r_M$ value increases, both feature mask sparsity and feature importance increase, improving the interpretability of the masks.
    }
    \label{fig:2x2_comparison}
\end{figure}

\textbf{Ablation: Affect of $r_M$ Regularizer.} Figure \ref{fig:2x2_comparison} illustrates an ablation study on how varying $r_M$ values affect our masks. We notice that at low $r_M$ values, test accuracy is high but feature selection diversity is poor. This makes mask interpretation difficult since almost all features are selected in the decision-making process. On the other hand, at high $r_M$ values, the masks are sparse and are easily interpretable but at the cost of accuracy. Therefore, we ensure that our selected $r_M$ using the Sparsity Regularizer Algorithm in Section \ref{reg_algo} provides us with a compromise of a competitive accuracy while having the best interpretability against the baseline models.

\textbf{Model Faithfulness (Qualitative).} To reinforce InterpreTabNet's ability to generate faithful feature masks, we assess our synthetic data generation process qualitatively. Depicted in \cref{fig:syn3_comparison} Appendix \ref{appen:syn_dataset}, InterpreTabNet determines features 2-5 to be the most salient features as shown by the yellow bars in most of its masks, aligning with the ground truth where features 2-5 are used to generate the synthetic data's predictions. On the other hand, TabNet illustrates lower salience levels in features 2-5 when conducting its decision process for the prediction, indicating its uncertainty in reasoning when predicting the ground truth.

\newpage
\textbf{Robustness and Reliability of Feature Masks.} To assess the robustness and reliability of InterpreTabNet's feature masks, we examine if salient feature identification differs across subgroups. We divide the Adult dataset into male and female subpopulations, with the feature importance visualization depicted in Figure \ref{fig:sex_subgroups_importance} Appendix \ref{appendix_mask_subgroups}. The salient features match well with known socio-economic factors affecting income. For males, occupation and capital gains highlight the importance of job roles and investments. For females, marital status notably influences income, likely due to the socio-economic dynamics and potential household income sharing.

\textbf{Rationale of Maximizing Feature Mask Diversity.} One could argue that if TabNet's feature masks often utilize the same features across stages, it might indicate that fewer features or stages are required for accurate predictions. Maximizing diversity could potentially introduce features that do not contribute to the prediction, thus complicating the model unnecessarily. However, we observe in Figure \ref{appen_fig_diversity} Appendix \ref{appen_diversity} that even when we select the least possible decision steps, $N_{steps}=2$, TabNet utilizes every single feature aggregated across the two masks. On the contrary, InterpreTabNet selects clear salient features in its decision-making process, without introducing features that do not contribute to the prediction.

\textbf{Training Stability.} In Figure \ref{fig:training_stability} Appendix \ref{append:training_stability}, we observe that InterpreTabNet’s training loss shows a general downward trend. Although InterpreTabNet exhibits higher variability in loss reduction across epochs, it suggests a more exploratory learning process to determine the salient features. Overall, InterpreTabNet's training process is relatively smooth compared to TabNet.

\textbf{Human Evaluation Survey on Interpretability.} We conducted a small-scale human evaluation survey on 20 Ph.D. and Masters students combined, with a machine learning background. It is conducted in a blind format where the identities of the models are anonymous to ensure that our data is trusted and not biased. The survey asks: \emph{"Which figure do you think is the best method to determine the important features?"}. 

\begin{table}[htb!]
\vspace{-3mm}
\centering
\caption{Survey on Interpretability}
\label{tab:human_eval}
\footnotesize 
    \begin{tabularx}{\columnwidth}{lXX}
    \toprule
    \textbf{Model} & \textbf{Vote Percentage} & \textbf{Number of Votes} \\
    \midrule
    \textbf{InterpreTabNet} & \textbf{65\%} & \textbf{13} \\
    \addlinespace
    TabNet & 15\% & 3 \\
    \addlinespace
    XGBoost & 5\% & 1 \\
    \addlinespace
    LightGBM & 5\% & 1 \\
    \addlinespace
    TabTransformer & 10\% & 2 \\
    \midrule
    Total & 100\% & 20 \\
    \bottomrule
    \end{tabularx}
\end{table}

\newpage
In Table \ref{tab:human_eval}, 65\% of respondents prefer InterpreTabnet to highlight salient features. This validates our motivation that sparse feature selection simplifies the complexity of the data into a more understandable form for practical applications. InterpreTabNet provides a concise set of important features, making it easier for users to understand the underlying reasons for predictions, trust the model's outputs, and explain these outcomes to stakeholders.\footnote{Link to survey: https://forms.gle/87PDZo56RUtHqFSb9}

\subsection{Capturing Feature Interdependencies by Prompting LLMs}
\label{prompt}

We have generated interpretable feature masks where users can determine the important features. However, one issue is that our approach does not grasp the \emph{interrelationships among features} required to explain model predictions in complicated scenarios. Therefore, we leveraged an LLM such as GPT-4 to \emph{incorporate extensive linguistic priors into the interpretation process} that help mitigate the issues.

\begin{table}[htb!]
\vspace{-1mm}
\centering
\caption{Prompt Structure Design}
\label{tab:prompt_structure}
\footnotesize 
\begin{tabularx}{\columnwidth}{l*{1}{X}}
\toprule
\textbf{Section} & \textbf{Description} \\
\midrule
Dataset Description & The Adult Census Income dataset is considered... \\
\addlinespace
Mask Description & At the 0th step of feature selection, we observe mask 0 with the main features highlighted as 1, 5, and 7... \\
\addlinespace
In-Context Example 1 & The Poker Hand dataset is considered... \\
\addlinespace
In-Context Output 1 & Output: \{"Mask 0": "Initially, the rank of card 2 is recognized...\} \\
\addlinespace
In-Context Example 2 & The Forest Cover Type dataset is considered... \\
\addlinespace
In-Context Output 2 & Output: \{"Mask 0": "The initial feature selection identifies...\} \\
\addlinespace
\midrule
\textit{GPT-4 Output} & \textit{\{"Mask 0": "This suggests that early in the feature selection process, the model finds that work-related...\}} \\
\bottomrule
\end{tabularx}
\end{table}

\textbf{Interpreting Feature Masks with GPT-4.} To generate a precise output mapping, we provided instructions to GPT-4 where the extracted salient features are formatted into a dictionary. Each mask corresponds to an individual analysis, followed by an aggregate analysis of all masks. Furthermore, a statement to ensure that GPT-4 produced no other natural language generation was added to maintain a consistent output map.

Finally, GPT-4 was provided with in-context examples to enable prompt tuning through few-shot learning. This was conducted via 3-fold cross-validation where datasets D1 and D2 were used as part of the prompt for tuning on D3, D2, and D3 as part of the prompt for tuning on D1, and so on. Only a 3-fold CV was conducted since increasing the subsets will decrease GPT-4's performance as it was unable to process extremely long sequences of texts. 

Overall, GPT-4 improves the analysis of salient features extracted from InterpreTabNet by explaining their interdependencies. The structure of the designed prompt can be found in Table \ref{tab:prompt_structure}. The full prompts and outputs can be found in Appendix \ref{appendix_prompt} and \ref{appendix_output} respectively.

\textbf{Human Evaluation Survey on GPT-4's Analysis.} We conducted another survey on GPT-4's analysis in the same format as our survey on interpretability. The survey asks: \emph{"Which model provides a more relevant and reasonable interpretation?"} when comparing between InterpreTabNet and TabNet. In Table \ref{tab:human_eval_2}, 60\% of the respondents prefer InterpreTabnet as the model with a more interpretable output from GPT-4. The clear-cut features allow GPT-4 to easily comprehend the information, providing a clear and concise output that explains the outcomes to the stakeholders. The prompt and GPT-4 output for TabNet can be found in Appendix \ref{tabnet_prompt}.\footnote{Link to survey: https://forms.gle/ZbGeXNF1HcSEYSNP7}

\begin{table}[htb!]
\centering
\caption{Survey on LLM-Generated Interpretations from InterpreTabNet vs. TabNet}
\label{tab:human_eval_2}
\footnotesize 
    \begin{tabularx}{\columnwidth}{lXX}
    \toprule
    \textbf{Model} & \textbf{Vote Percentage} & \textbf{Number of Votes} \\
    \midrule
    \textbf{InterpreTabNet} & \textbf{60\%} & \textbf{12} \\
    \addlinespace
    TabNet & 40\% & 8 \\
    \midrule
    Total & 100\% & 20 \\
    \bottomrule
    \end{tabularx}
\end{table}

Though the results of our surveys are promising, future work can replicate these findings on a larger sample size.

\textbf{Justifying GPT-4's Interpretation Capabilities.} A potential concern is whether GPT-4 actually interprets the model's internal behavior rather than merely rephrasing the prompt input. To address this, we conduct the following experiments to demonstrate that the model exhibits a genuine understanding of the data.

We test the \textit{integrity} of our prompt with a definition check on "feature mask". Figure \ref{fig:mask_definition_check} in Appendix \ref{appendix_feature_mask_definition_check} aligns with our expectations, providing an accurate and detailed explanation. Next, we verify the \textit{reliability} of GPT-4's analysis by prompting it to interpret synthetic datasets generated in the manner from \citep{yoon2018invase}. The results in Appendix \ref{appendix_gpt4_check} indicate that the analysis is indeed robust since it does not show any signs of hallucinations or mistakes hence, verifying the integrity of GPT-4's ability to interpret feature masks.

\newpage
Additionally, we try \textit{prompt diversification} to strengthen the trust in the generated explanations while determining which prompt design yields the most insightful and accurate explanations from GPT-4. We explore explanations in different formats and varying levels of detail. The results can be found in Tables \ref{appen:prompt_format} and \ref{appen:prompt_detail} of Appendix \ref{appendix_prompt_variability}. Our analysis indicates that our original prompt structure leads to a higher level of detail and attempts to deduce deeper meanings from the prominent features, as opposed to simply categorizing them. This suggests a level of interpretive understanding by GPT-4 that went beyond basic rephrasing, thereby reinforcing the effectiveness of our prompt design.

To determine whether \textit{GPT-4's interpretations correlate with human interpretations}, we run an experiment to identify the variability in interpretations based on what an expert in the domain might care about. We summarize the variation of results in Appendix \ref{domain_expert} — what we find is that the expert interpretations do not vary significantly compared to the original interpretations given different "expert" prompts. Therefore, we can anticipate that the interpretations are "robust".

Lastly, we test if \textit{GPT-4 alone without InterpreTabNet} could identify the salient features and elucidate their relationships from the Adult dataset. However, as observed in Table \ref{appen:gpt_no_intabnet} Appendix \ref{appendix_gpt_without_interpretabnet}, GPT-4 is unable to determine the salient features even when the dataset information is provided. For the first prompt, it is unable to compile any aggregate analysis. In the second prompt, the extracted important features are not accurate as well as being dense, selecting more than 50\% (8/14) of the features.

\section{Conclusion}
We propose an interpretable variant of the TabNet neural network that is as expressive in learning the distributions of tabular data while enabling an enhanced level of interpretability. This model is designed by blending a Gumbel-Softmax distribution with a KL divergence sparsity regularizer between the attention-based feature masks to create a sparse and semantically meaningful decomposition of the predictive signals. Relative to our baselines, our model outputs more interpretable feature masks to determine salient features while maintaining its competitive accuracy across most datasets. The salient features from our masks are channeled into GPT-4 via prompts that encourage a careful analysis of the features' interdependencies. For practitioners, InterpreTabNet distills the predictive signals allowing it to stand as a practical toolkit for understanding where tabular data comes from. It bridges the often challenging gap between intricate machine learning outputs and real-world decision-making, ensuring that insights are not just extracted but also intuitively understood and readily actionable.

\newpage
\section*{Impact Statement}
This paper introduces InterpreTabNet to improve the interpretability of machine learning models that handle tabular data. Its primary societal impact lies in offering more transparent, understandable deep-learning predictions and decisions. This is crucial in high-stakes human-oriented sectors such as healthcare and finance, where decision-making impacts human lives.
Ethically, InterpreTabNet represents a step towards responsible AI, as it allows users to understand how and why specific decisions are made, enhancing trust and reducing the 'black box' nature of complex models. However, there are potential risks. An example could be an over-reliance on model interpretations, leading to neglecting other important factors not captured by the model. 

\section*{Acknowledgements}
We thank the University of Toronto for supporting this research. MC is supported by a Vector Institute Graduate Student Fellowship and a CIHR Health Systems Impact Fellowship. RGK is supported by a Tier II Canada Research Chair and Canada CIFAR AI Chair.

\bibliographystyle{abbrvnat}
\bibliography{reference}

\newpage
\appendix
\onecolumn

\begin{center}
\LARGE
\textbf{Appendix}
\end{center}

\section{Proofs and Algorithms}
\subsection{Proof: cVAE Evidence Lower Bound}
\label{appendix_cvae}

\begin{align*}
    D_{KL}[Q(z|Y, X)||P(z|Y, X)] &= \sum_{z}Q(z|Y, X)\log\frac{Q(z|Y, X)}{P(z|Y, X)} \\
    &= E[\log\frac{Q(z|Y, X)}{P(z|Y, X)}] \\
    &= E[\log Q(z|Y, X) - \log P(z|Y, X)] \\
        &\text{using Bayes' rule} \\
    &= E[\log Q(z|Y, X) - \log \frac{P(z,Y,X)}{P(Y,X)}] \\ 
    &= E[\log Q(z|Y, X) - \log \frac{P(Y|z, X)P(z, X)}{P(Y,X)}] \\ 
    &= E[\log Q(z|Y, X) - \log \frac{P(Y|z, X)P(z|X)P(X)}{P(Y,X)}] \\ 
    &= E[\log Q(z|Y, X) - \log \frac{P(Y|z, X)P(z|X)P(X)}{P(Y|X)P(X)}] \\ 
    &= E[\log Q(z|Y, X) - \log \frac{P(Y|z, X)P(z|X)}{P(Y|X)}] \\ 
    &= E[\log Q(z|Y, X) - (\log P(Y|z, X) + \log P(z|X) - \log P(Y|X))] \\
    &= E[\log Q(z|Y, X) - \log P(Y|z, X) - \log P(z|X) + \log P(Y|X)] \\
    &= E[\log Q(z|Y, X) - \log P(Y|z, X) - \log P(z|X)] + \log P(Y|X) \\
    D_{KL}[Q(z|Y, X)||P(z|Y, X)] - \log P(Y|X) &= E[\log Q(z|Y, X) - \log P(Y|z, X) - \log P(z|X)] \\
    &\text{rearranging the sign to rewrite RHS as another KL Divergence}  \\
    \log P(Y|X) - D_{KL}[Q(z|Y, X)||P(z|Y, X)] &= E[\log P(Y|z, X) - (\log Q(z|Y, X) - \log P(z|X))] \\
    \log P(Y|X) - D_{KL}[Q(z|Y, X)||P(z|Y, X)] &= E[\log P(Y|z, X)] - E[\log Q(z|Y, X) - \log P(z|X)] \\
    \log P(Y|X) - D_{KL}[Q(z|Y, X)||P(z|Y, X)] &= E[\log P(Y|z, X)] - E[\log \frac{Q(z|Y, X)}{P(z|X)}] \\
    \log P(Y|X) - D_{KL}[Q(z|Y, X)||P(z|Y, X)] &= E[\log P(Y|z, X)] - D_{KL}[Q(z|Y, X)||P(z|X)] \\
    &\text{which is the cVAE objective function} \\
\end{align*} \\

\newpage

\subsection{Algorithm: KL Divergence Sparsity Regularizer $r_M$}
\label{appendix_algorithm}

The algorithm analyzes the model's feature importance masks to validate that they meet a set criterion. This criterion is to validate that the masks are sparse and that the features the model selects are important. Upon fulfilling the criterion a specific number of times, the algorithm terminates. To increase efficiency, the algorithm also employs a recursive search to narrow down the value range around the current best $r_M$, thereby reducing computational overhead.

\begin{algorithm}[H]
\small
\caption{Our proposed algorithm for interpretability optimization. Good default settings for the tested machine learning problems are $\alpha = 0$, $\beta = [0, 10000000]$, $\delta = [0.20, 0.25]$, $\gamma = [2, 3]$ $\epsilon = [3, 5]$. For $\beta$, $\delta$ and $\gamma$, it would depend on the nature of the dataset. More samples require higher parameter values.}
\label{alg:search-best-reg-m}
\begin{algorithmic}[1]
\Require \( \alpha \): Starting range (start)
\Require \( \beta \): Ending range (end)
\Require \( \delta \): Percentage of feature importance captured by one feature in each feature mask (col\_threshold\_val)
\Require \( \gamma \): Number of columns that satisfies \( \delta \) in each feature mask (col\_threshold)
\Require \( \iota \): Number of complete-feature masks that passes the algorithm's feature selection criteria (all\_mask\_pass)
\Require \( \epsilon \): Threshold for the number of complete-feature masks that passes the algorithm's feature selection criteria (all\_mask\_pass\_thresh)
\Require \( \zeta \): Step size computed using a logarithmic scale at high levels (step\_size)
\Require \( \theta \): Dictionary storing \( r_M \)-accuracy pairs (reg\_m\_acc\_dict)
\Require \( \lambda \): Flag for recursion (is\_recursive)

\Ensure Optimal regularization parameter \( r_M^* \)

\Statex

\State Initialize \( \theta \) if \( \theta \) is None.
\State Initialize \( \iota \) if \( \iota \) is None.

\If{ \( \iota = \epsilon \) }
    \State \( r_M^* = \arg\max(\theta) \)
    \State \Return \( r_M^* \)
\EndIf

\While{ \( \alpha \leq \beta \) and \( \iota < \epsilon \) }
    \State \text{Train TabNet, Compute Accuracy and Generate Masks}
    
    \Statex  \Comment{Inner loop evaluating each feature mask here.}

    \If{ \text{Criteria for updating \( \theta \) and \( \iota \) are met} }
        \State Update \( \theta \), \( \iota \)
    \EndIf

    \If{ \( \lambda \) }
        \State \( \alpha = \alpha + \zeta \)
        \ElsIf{ \( \alpha = 0 \) }
            \State \( \alpha = 10 \)
        \Else
            \State \( \alpha *= 10 \)
    \EndIf

\EndWhile

\If{ \( r_M^* \) is Not None \& \text{Length of } \( \theta = 1 \) }
    \State Recurse with updated boundaries.
\Else
    \State \( r_M^* = \arg\max(\theta) \)
    \State \Return \( r_M^* \)
\EndIf

\end{algorithmic}
\end{algorithm}

\newpage
\section{Experimental Setup and Datasets}
\subsection{Reproducibility}
\label{reproducibility}
\textbf{Code Release.} The code for InterpreTabNet and files to reproduce the experiments are available on GitHub at:\\ \url{https://github.com/jacobyhsi/InterpreTabNet}.

\textbf{Availability of Datasets.} The datasets used in this paper are all freely accessible on OpenML. \verb|OpenML.org| and \verb|UCI Machine Learning Repository|. Download links and additional statistical details about the datasets can be found in Appendix \ref{appendix_dataset} of the paper.

\textbf{GPT-4 Version.} The GPT-4 version used in our experiments is "gpt-4-1106" with training data up to Apr 2023 and a context window of 128,000 tokens.

\subsection{Additional Dataset Information}
\label{appendix_dataset}

We evaluated our model on 7 datasets. These datasets contain 4 binary classification tasks and 3 multi-class classification tasks. We provided statistical details in Table \ref{tab:dataset_info}, and download links in Table \ref{tab:dataset_link}. 
In each of our datasets, we applied label encoding to the categorical features to transform textual values into numerical representations. Additionally, we introduced a distinct token to handle missing data within these categorical columns. This uniform preprocessing approach was applied consistently across all datasets, ensuring compatibility and reliability for subsequent machine learning analyses.\\

\begin{table}[htbp]
    \centering
    \caption{Datasets used for evaluation}
    \begin{tabularx}{\textwidth}{X l l l l l l}
        \toprule
        \textbf{Dataset} & \textbf{Task} & \textbf{\# Features} & \textbf{\# Categorical} & \textbf{\# Instances} & \textbf{\# Classes} & \textbf{\# NaNs} \\
        \midrule
        Adult Census Income & Binary & 14 & 8 & 32,560 & 2 & 0 \\
        Forest Cover Type & Multi-Class & 54 & 44 & 581,012 & 7 & 0 \\
        Poker Hand & Multi-Class & 10 & 10 & 1,025,010 & 10 & 0 \\
        Mushroom & Binary & 22 & 22 & 8,124 & 2 & 0 \\
        Blastchar & Binary & 20 & 17 & 7,043 & 2 & 0 \\
        Diabetes & Multi-Class & 49 & 39 & 101,766 & 3 & 0 \\
        Higgs & Binary & 28 & 0 & 11,000,000 & 2 & 0 \\
        \bottomrule
    \end{tabularx}
    \label{tab:dataset_info}
\end{table}

\begin{table}[htbp]
    \centering
    \caption{Dataset Links}
    \begin{tabularx}{\textwidth}{lX}
        \toprule
        \textbf{Dataset Name} & \textbf{Dataset Link} \\
        \midrule
        Adult Census Income & \url{https://archive.ics.uci.edu/dataset/2/adult} \\
        Forest Cover Type & \url{https://archive.ics.uci.edu/dataset/31/covertype} \\
        Poker Hand & \url{https://archive.ics.uci.edu/dataset/158/poker+hand} \\
        Mushroom & \url{https://archive.ics.uci.edu/dataset/73/mushroom} \\
        Blastchar & \url{https://www.kaggle.com/datasets/blastchar/telco-customer-churn} \\
        Diabetes & \url{https://archive.ics.uci.edu/dataset/296/diabetes+130-us+hospitals+for+years+1999-2008} \\
        Higgs & \url{https://archive.ics.uci.edu/dataset/280/higgs} \\
        \bottomrule
    \end{tabularx}
    \label{tab:dataset_link}
\end{table}

\newpage
\subsection{Hyperparameters}

\subsubsection*{Guidelines}
Hyperparameters such as $N_d = N_a$, $N_{steps}$, $\gamma$, and learning rate are tuned in the range per TabNet's recommendations. In terms of the sparsity regularizer for InterpreTabNet, $r_M$, we recommend a smaller range e.g. [0,10000] for datasets with a low to moderate number of features and samples (Adult dataset), and a larger range e.g. [0, 1,000,000,000,000] for  datasets with a larger number of features and samples (Higgs dataset). Within the sparsity algorithm itself (\cref{appendix_algorithm}), parameters such as the number of salient features and feature importance threshold can be adjusted to the user's preference.

\subsubsection*{Search Space}
We provided hyperparameter search spaces for all models in Table \ref{tab:hyperparameter_space}. For TabTransformer, we used the same hyperparameter space mentioned in their paper \citep{tabtransformer}. XGboost and LightGBM were designed from scratch and used common hyperparameter choices with suggestions from the official documentation \citep{xgboost} \citep{lightgbm}. For MLP, we followed the exact hyperparameter search space as \citep{tabtransformer}.

\begin{table}[htbp]
\centering
\caption{Hyperparameter spaces for all models}
\label{tab:hyperparameter_space}
\begin{tabular}{ll}
\toprule
\textbf{Model} & \textbf{Hyperparameter Space} \\
\midrule
InterpreTabNet & 
\begin{tabularx}{0.63\textwidth}{@{}X@{}}
$N_d = N_a$ (output dimension): [16, 32, 128], \\
$N_{steps}$: [3, 4, 5], \\
$\gamma$: [1.0, 1.2, 1.5, 2.0], \\
Learning Rate: [0.005, 0.01, 0.02, 0.025], \\
$r_M$: range from [0, 1,000,000,000,000]
\end{tabularx} \\ \addlinespace \midrule
Original TabNet & 
\begin{tabularx}{0.63\textwidth}{@{}X@{}}
$N_d = N_a$ (output dimension): [16, 32, 128], \\
$N_{steps}$: [3, 4, 5], \\
$\gamma$: [1.0, 1.2, 1.5, 2.0], \\
Learning Rate: [0.005, 0.01, 0.02, 0.025]
\end{tabularx} \\ \addlinespace \midrule
TabTransformer & 
\begin{tabularx}{0.63\textwidth}{@{}X@{}}
Hidden Dimension: [32, 54, 128, 256], \\
Number of Layers: [1, 2, 3, 6, 12], \\
Number of Attention Heads: [2, 4, 8], \\
MLP First Hidden Layer: ${x = m \times l, m \in \mathbb{Z} | 1 \leq m \leq 8}$, where $l$ is the input size, \\
MLP Second Hidden Layer: ${x = m \times l, m \in \mathbb{Z} | 1 \leq m \leq 3}$, where $l$ is the input size
\end{tabularx} \\ \addlinespace \midrule
XGBoost & 
\begin{tabularx}{0.63\textwidth}{@{}X@{}}
learning\_rate: [0.01, 0.1, 0.2], \\
max\_depth: [3, 4, 5, 6], \\
n\_estimators: [50, 100, 200], \\
subsample: [0.8, 0.9], \\
colsample\_bytree: [0.8, 0.9], \\
min\_child\_weight: [1, 2, 3]
\end{tabularx} \\ \addlinespace \midrule
LightGBM & 
\begin{tabularx}{0.63\textwidth}{@{}X@{}}
num\_leaves: [20, 30, 40], \\
learning\_rate: [0.05, 0.1, 0.2], \\
n\_estimators: [100, 200], \\
subsample: [0.8, 0.9], \\
colsample\_bytree: [0.8, 0.9]
\end{tabularx} \\ \addlinespace \midrule
MLP & 
\begin{tabularx}{0.63\textwidth}{@{}X@{}}
First Hidden Layer: ${x = m \times l, m \in \mathbb{Z} | 1 \leq m \leq 8}$, where $l$ is the input size, \\
Second Hidden Layer: ${x = m \times l, m \in \mathbb{Z} | 1 \leq m \leq 3}$, where $l$ is the input size
\end{tabularx} \\
\bottomrule
\end{tabular}
\end{table}

\newpage

\section{Additional Experimental Results from Other Datasets}
\label{appendix_exp_other_datasets}

\subsection{Accuracies \& Masks}
\label{appendix_best_acc_masks}

\subsubsection*{Forest Cover Type \citep{Dua:2019}}
 
\begin{table}[htbp]
    \centering
    \begin{tabular}{lc}
        \toprule
        \textbf{Model} & \textbf{Test Accuracy (\%)} \\
        \midrule
        InterpreTabNet $(r_M^* = 900)$ & \textbf{$94.75\pm0.53$} \\
        Original TabNet & $94.18\pm0.63$\\
        XGBoost & $92.30\pm0.62$ \\
        LightGBM & $86.38\pm0.64$ \\
        TabTransformer & $82.55\pm0.39$ \\
        MLP & $84.89\pm0.56$ \\
        \bottomrule
    \end{tabular}
\end{table}

\begin{figure}[htb!]
    \centering
    
    \begin{minipage}[b]{0.9\textwidth}
        \includegraphics[width=\textwidth]{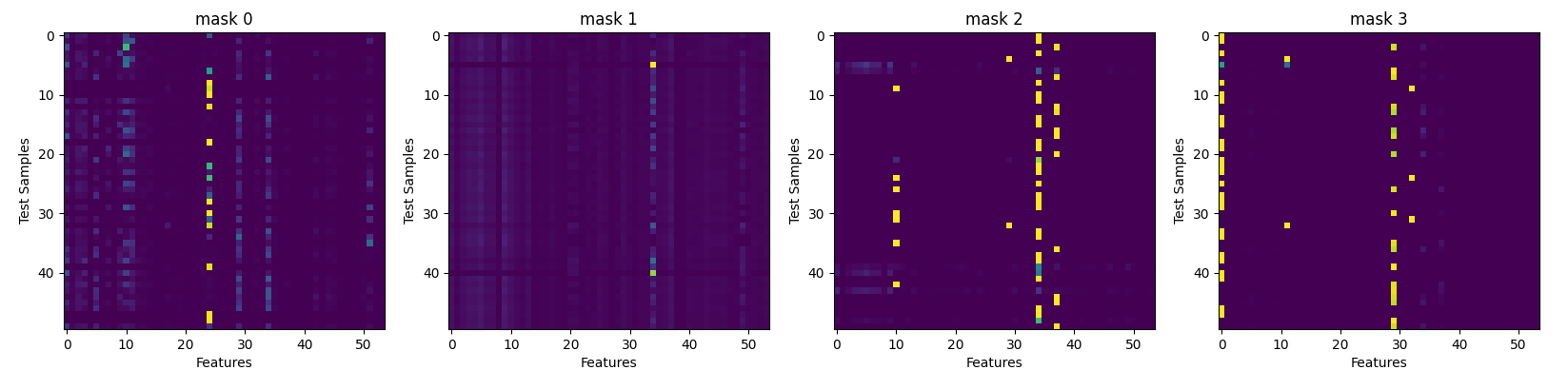}
        \caption*{(a) InterpreTabNet} 
    \end{minipage}
    \hfill
    \begin{minipage}[b]{0.9\textwidth}
        \includegraphics[width=\textwidth]{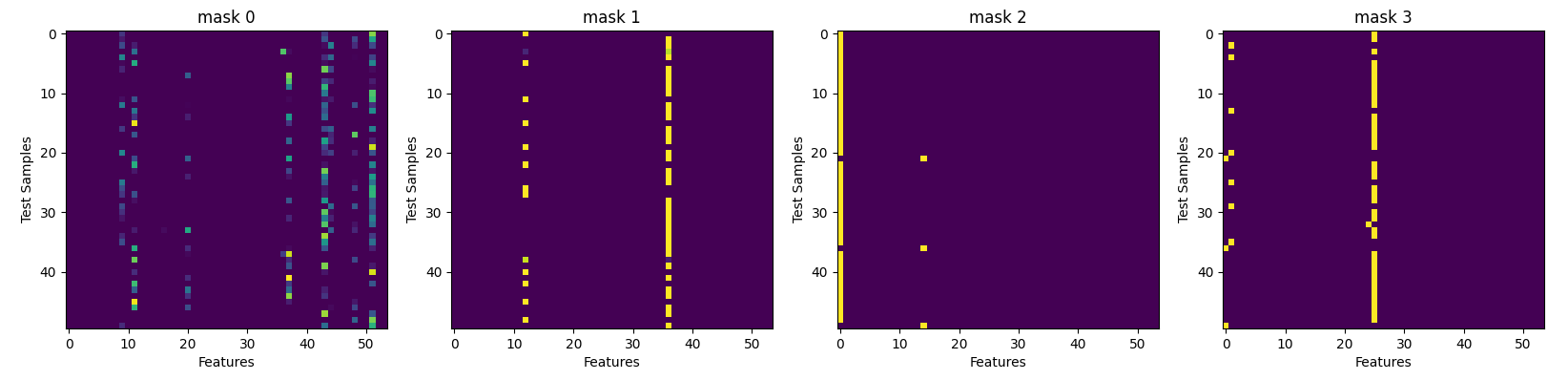}
        \caption*{(b) Original TabNet Model} 
    \end{minipage}
    
    \caption{(a) Best performing model using InterpreTabNet $r_M^* = 900$ with an accuracy of 94.75\% on the Forest Cover Type Dataset. (b) The baseline performance using the Original TabNet model, attaining an accuracy of 94.18\%.}
    \label{fig:forest_comparison}
\end{figure}

\newpage

\subsubsection*{Poker Hand \citep{poker_hand}}

\begin{center}
    \begin{tabular}{lc}
        \toprule
        \textbf{Model} & \textbf{Test Accuracy (\%)} \\
        \midrule
        InterpreTabNet $(r_M^* = 1000)$ & \textbf{$99.50\pm0.48$} \\
        Original TabNet & $99.00\pm0.62$ \\
        XGBoost & $75.57\pm0.47$ \\
        LightGBM & $78.47\pm0.36$ \\
        TabTransformer & $99.81\pm0.31$ \\
        MLP & $99.70\pm0.56$ \\
        \bottomrule
    \end{tabular}
\end{center}

\begin{figure}[htb!]
    \centering
    
    \begin{minipage}[b]{0.45\textwidth}
        \includegraphics[width=\textwidth]{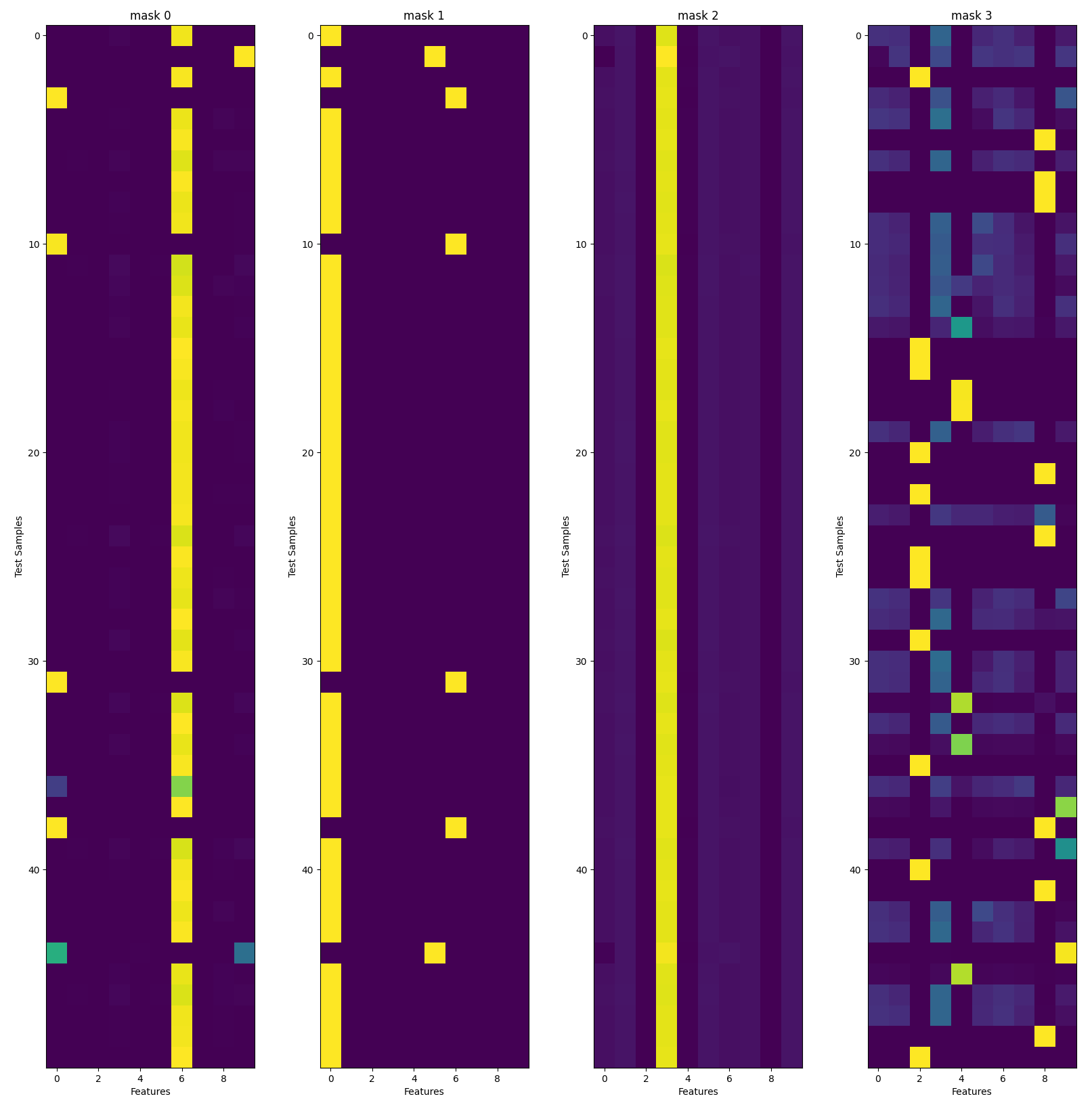}
        \caption*{(a) InterpreTabNet} 
    \end{minipage}
    \hfill
    \begin{minipage}[b]{0.45\textwidth}
        \includegraphics[width=\textwidth]{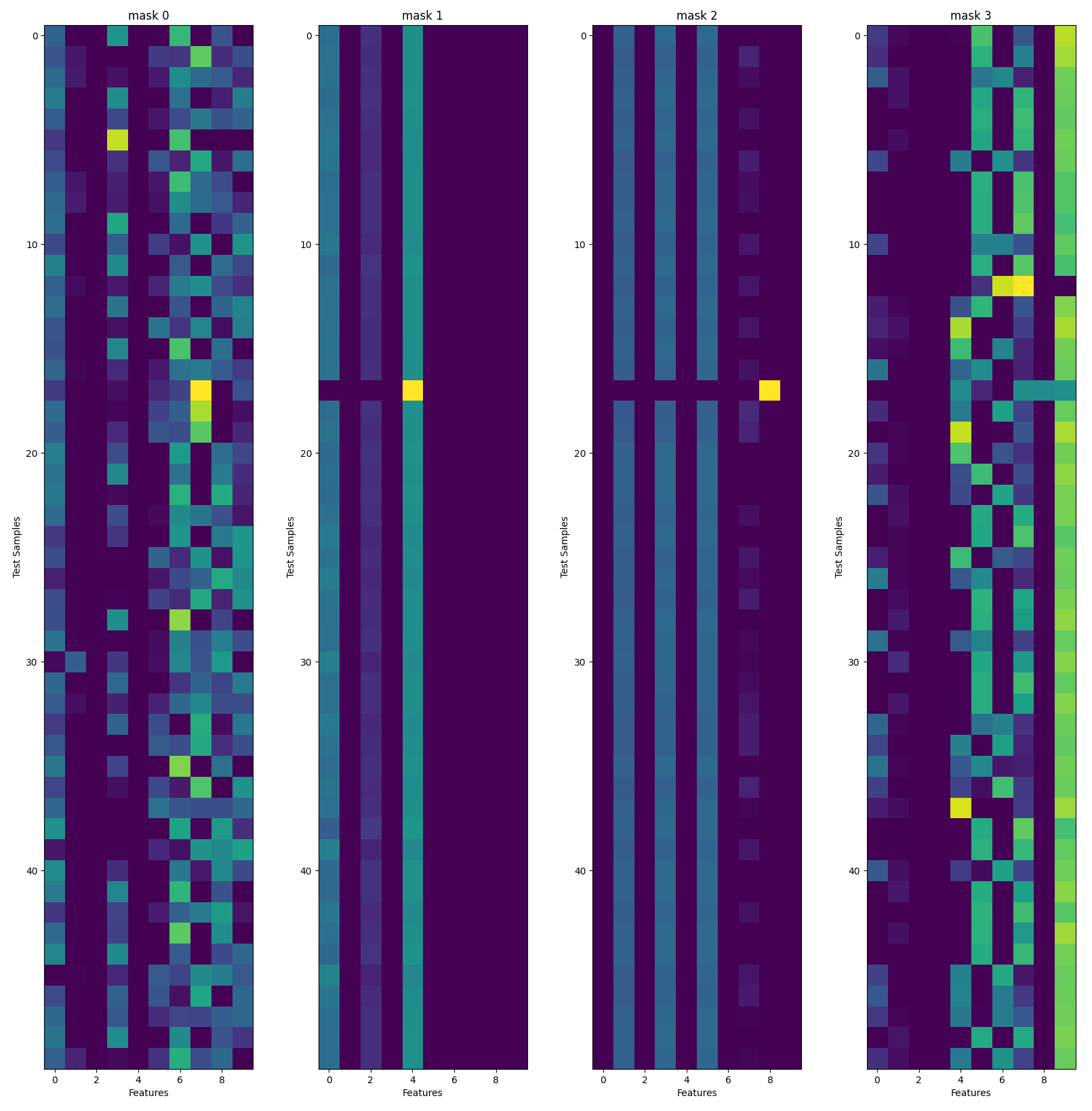}
        \caption*{(b) Original TabNet Model} 
    \end{minipage}
    
    \caption{(a) Best performing model using InterpreTabNet $r_M^* = 1000$ with an accuracy of 99.13\% on the Poker Hand Dataset. (b) The baseline performance using the Original TabNet model, attaining an accuracy of 99.23\%.}
    \label{fig:poker_comparison}
\end{figure}

\newpage

\subsubsection*{Higgs \citep{higgs}}

\begin{center}
    \begin{tabular}{lc}
        \toprule
        \textbf{Model} & \textbf{Test Accuracy (\%)} \\
        \midrule
        InterpreTabNet $(r_M^* = 10000)$ & \textbf{$53.08\pm0.56$} \\
        Original TabNet & $52.94\pm0.45$ \\
        XGBoost & $72.70\pm0.35$ \\
        LightGBM & $72.62\pm0.35$ \\
        TabTransformer & $51.97\pm0.54$ \\
        MLP & $63.17\pm0.36$ \\
        \bottomrule
    \end{tabular}
\end{center}

\begin{figure}[htb!]
    \centering
    
    \begin{minipage}[b]{0.8\textwidth}
        \includegraphics[width=\textwidth]{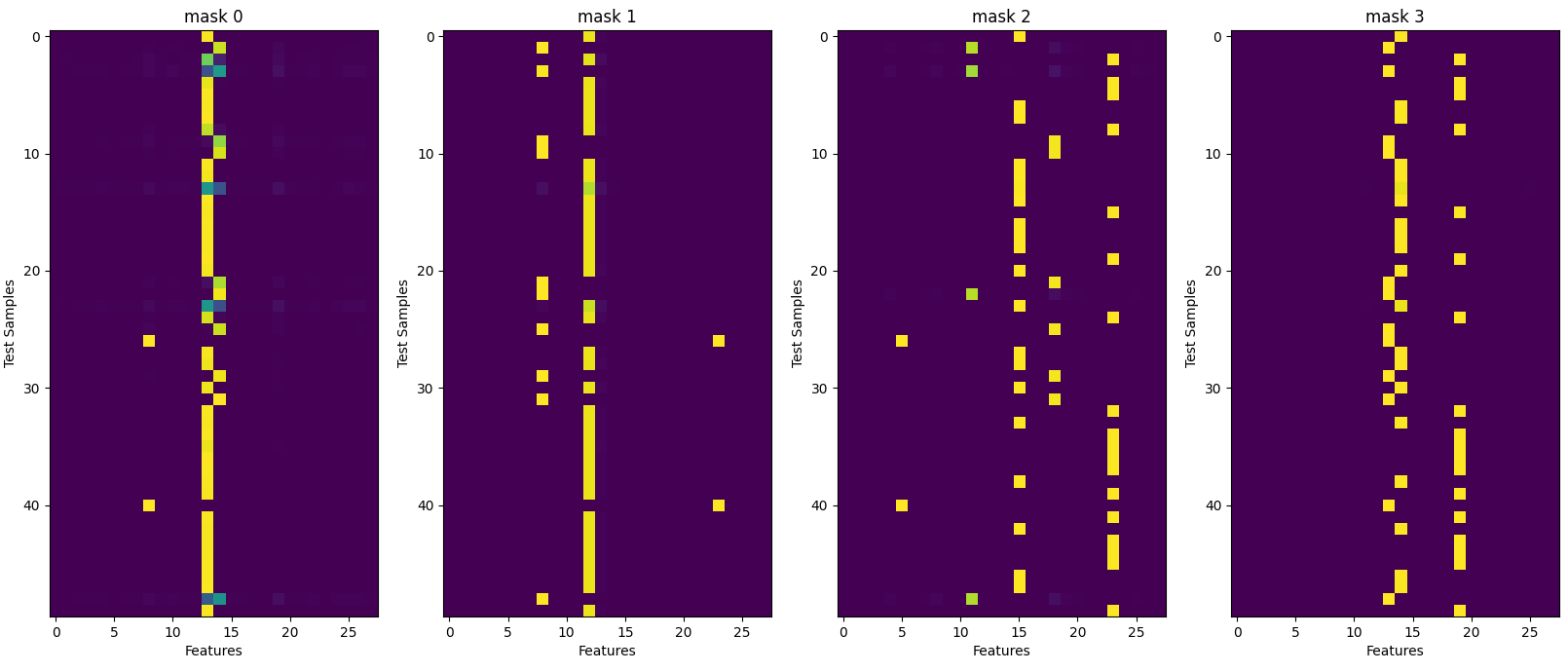}
        \caption*{(a) InterpreTabNet} 
    \end{minipage}
    \hfill
    \begin{minipage}[b]{0.8\textwidth}
        \includegraphics[width=\textwidth]{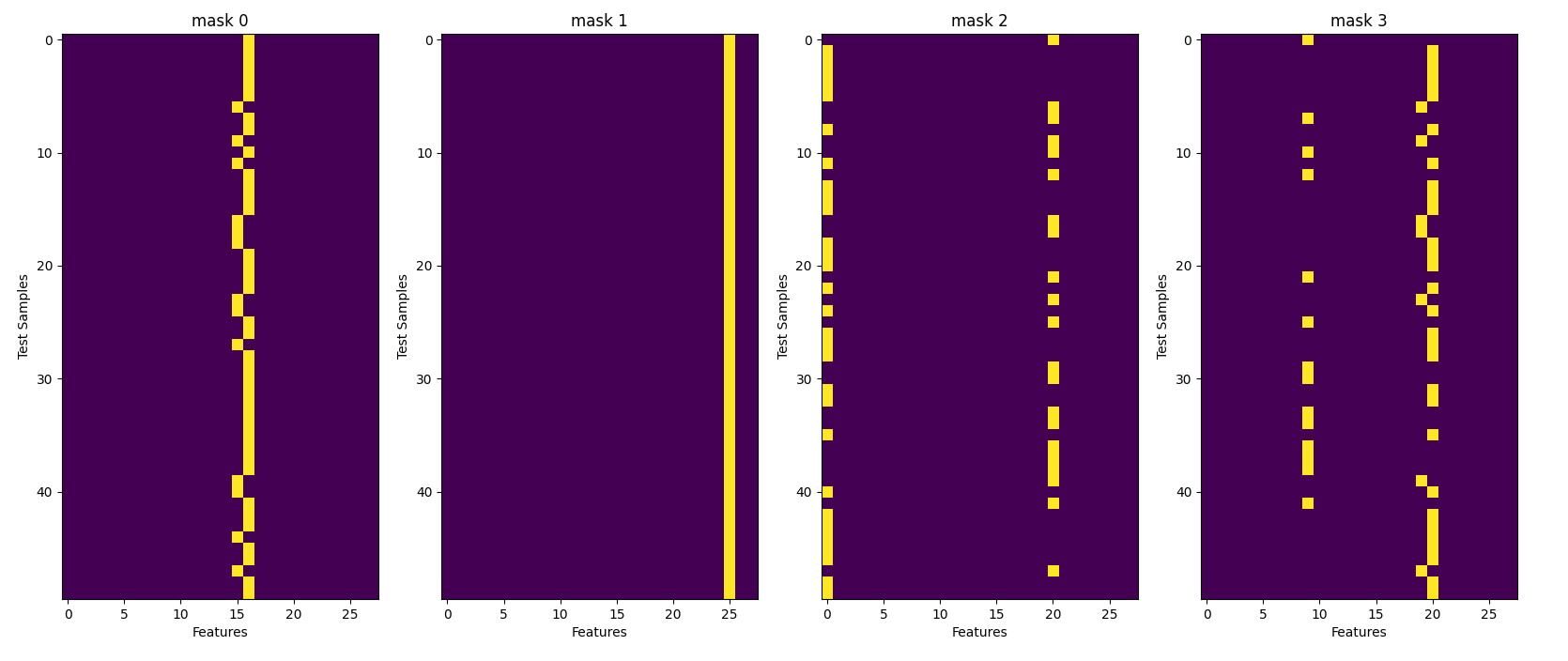}
        \caption*{(b) Original TabNet Model} 
    \end{minipage}
    
    \caption{(a) Best performing model using InterpreTabNet $r_M^* = 10000$ with an accuracy of 53.08\% on the Higgs Dataset. (b) The baseline performance using the Original TabNet model, attaining an accuracy of 60.22\%.}
    \label{fig:higgs_comparison}
\end{figure}

\subsubsection*{Mushroom \citep{mushroom}}

\begin{center}
    \begin{tabular}{lc}
        \toprule
        \textbf{Model} & \textbf{Test Accuracy (\%)} \\
        \midrule
        InterpreTabNet $(r_M^* = 10,000,000,000,000)$ & \textbf{$96.62\pm0.35$} \\
        Original TabNet & $99.94\pm0.31$ \\
        XGBoost & $99.69\pm0.39$ \\
        LightGBM & $100.00\pm0.37$ \\
        TabTransformer & $100.00\pm0.57$ \\
        MLP & $99.82\pm0.56$ \\
        \bottomrule
    \end{tabular}
\end{center}

\begin{figure}[htb!]
    \centering
    
    \begin{minipage}[b]{0.8\textwidth}
        \includegraphics[width=\textwidth]{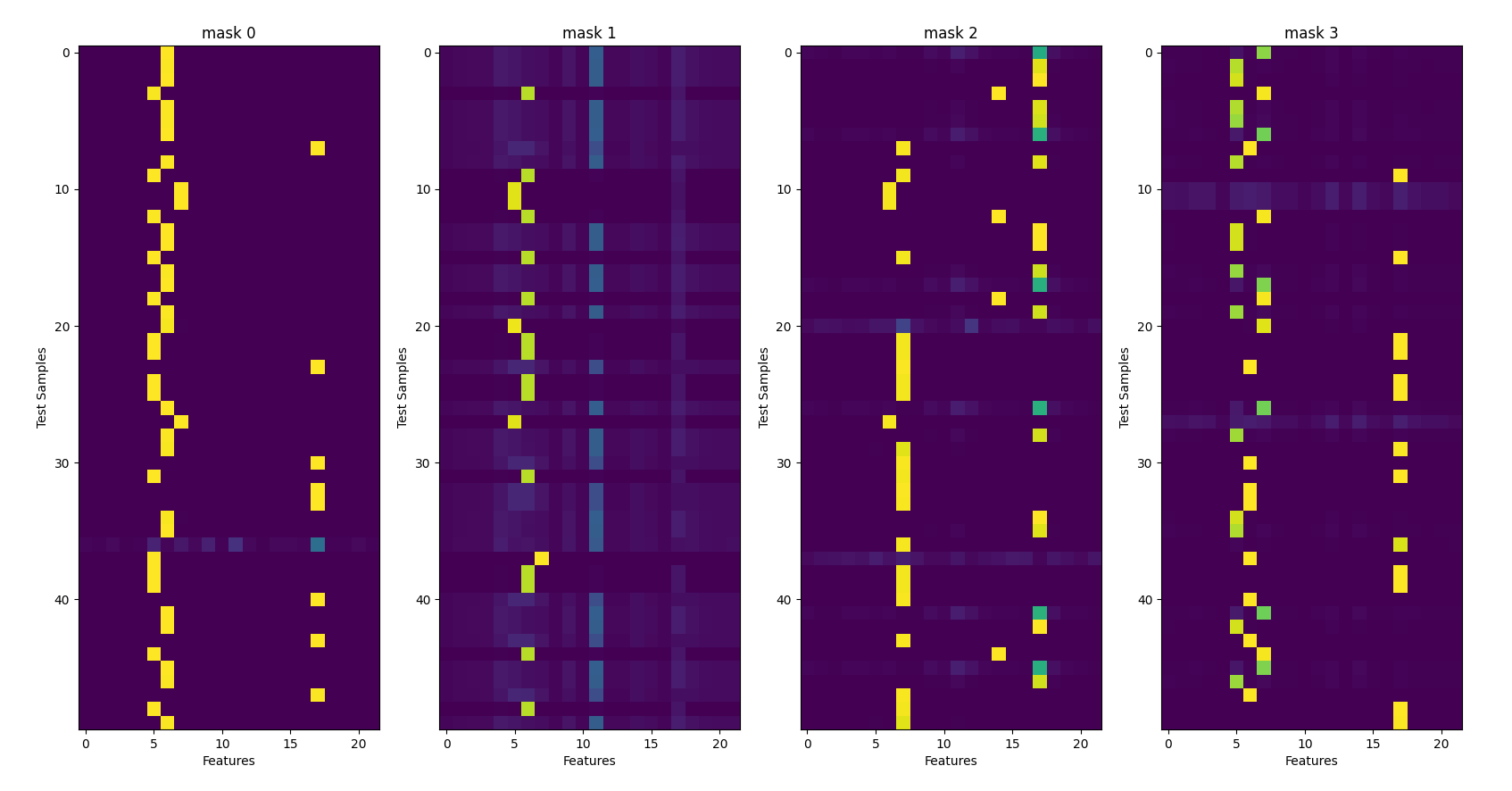}
        \caption*{(a) InterpreTabNet} 
    \end{minipage}
    \hfill
    \begin{minipage}[b]{0.8\textwidth}
        \includegraphics[width=\textwidth]{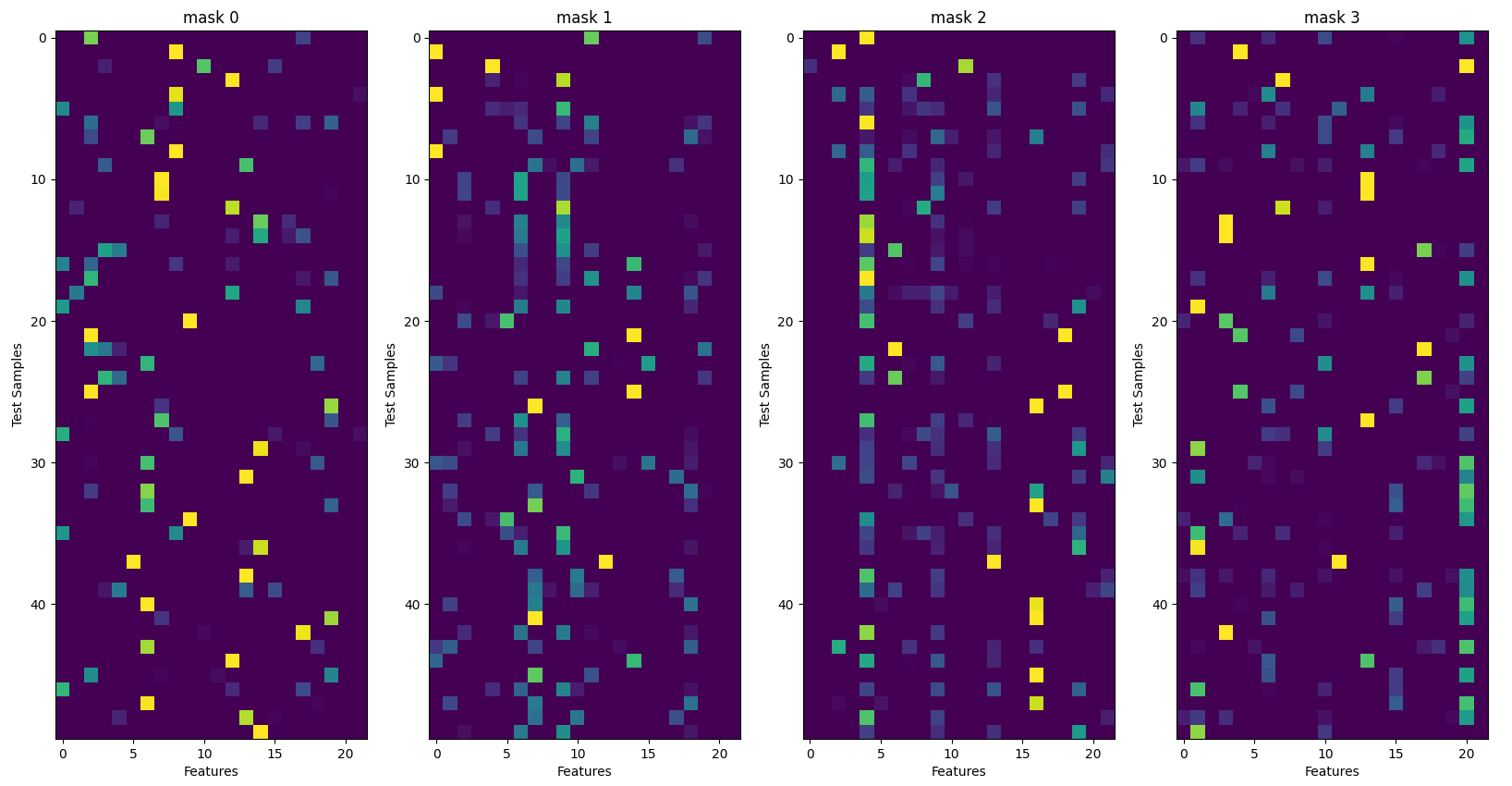}
        \caption*{(b) Original TabNet Model} 
    \end{minipage}
    
    \caption{(a) Best performing model using InterpreTabNet $r_M^* = 1,000,000,000,0000$ with an accuracy of 96.62\% on the Mushroom Dataset. (b) The baseline performance using the Original TabNet model, attaining an accuracy of 99.94\%.}
    \label{fig:mushroom_comparison}
\end{figure}

\subsubsection*{Blastchar \citep{blastchar}}

\begin{center}
    \begin{tabular}{lc}
        \toprule
        \textbf{Model} & \textbf{Test Accuracy (\%)} \\
        \midrule
        InterpreTabNet $(r_M^* = 10,000,000,000,000)$ & \textbf{$72.96\pm0.56$} \\
        Original TabNet & $76.22\pm0.34$ \\
        XGBoost & $77.29\pm0.53$ \\
        LightGBM & $77.86\pm0.46$ \\
        TabTransformer & $73.17\pm0.56$ \\
        MLP & $75.16\pm0.61$ \\
        \bottomrule
    \end{tabular}
\end{center}

\begin{figure}[htb!]
    \centering
    
    \begin{minipage}[b]{0.8\textwidth}
        \includegraphics[width=\textwidth, height=7cm]{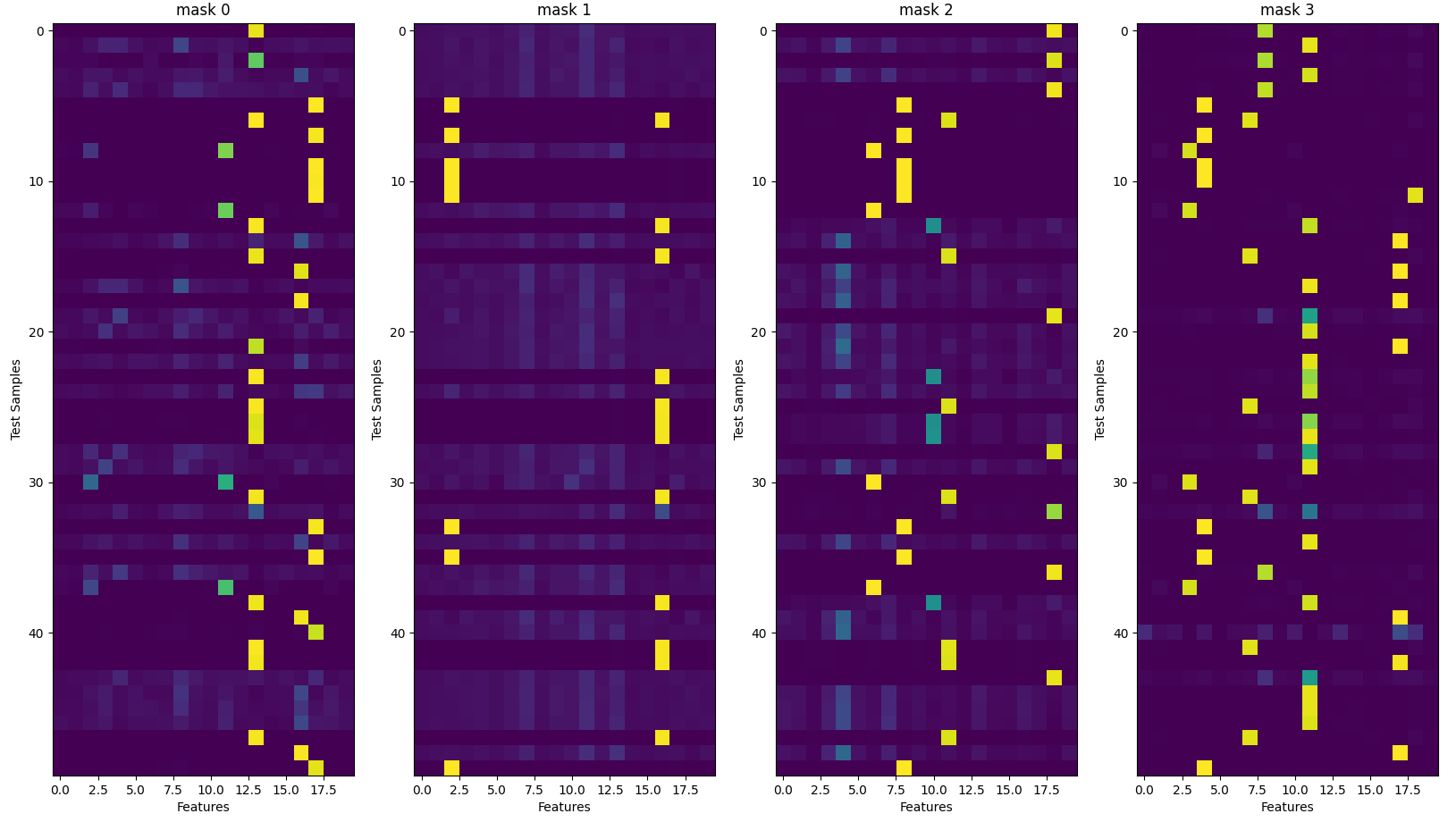}
        \caption*{(a) InterpreTabNet} 
    \end{minipage}
    \hfill
    \begin{minipage}[b]{0.8\textwidth}
        \includegraphics[width=\textwidth, height=7cm]{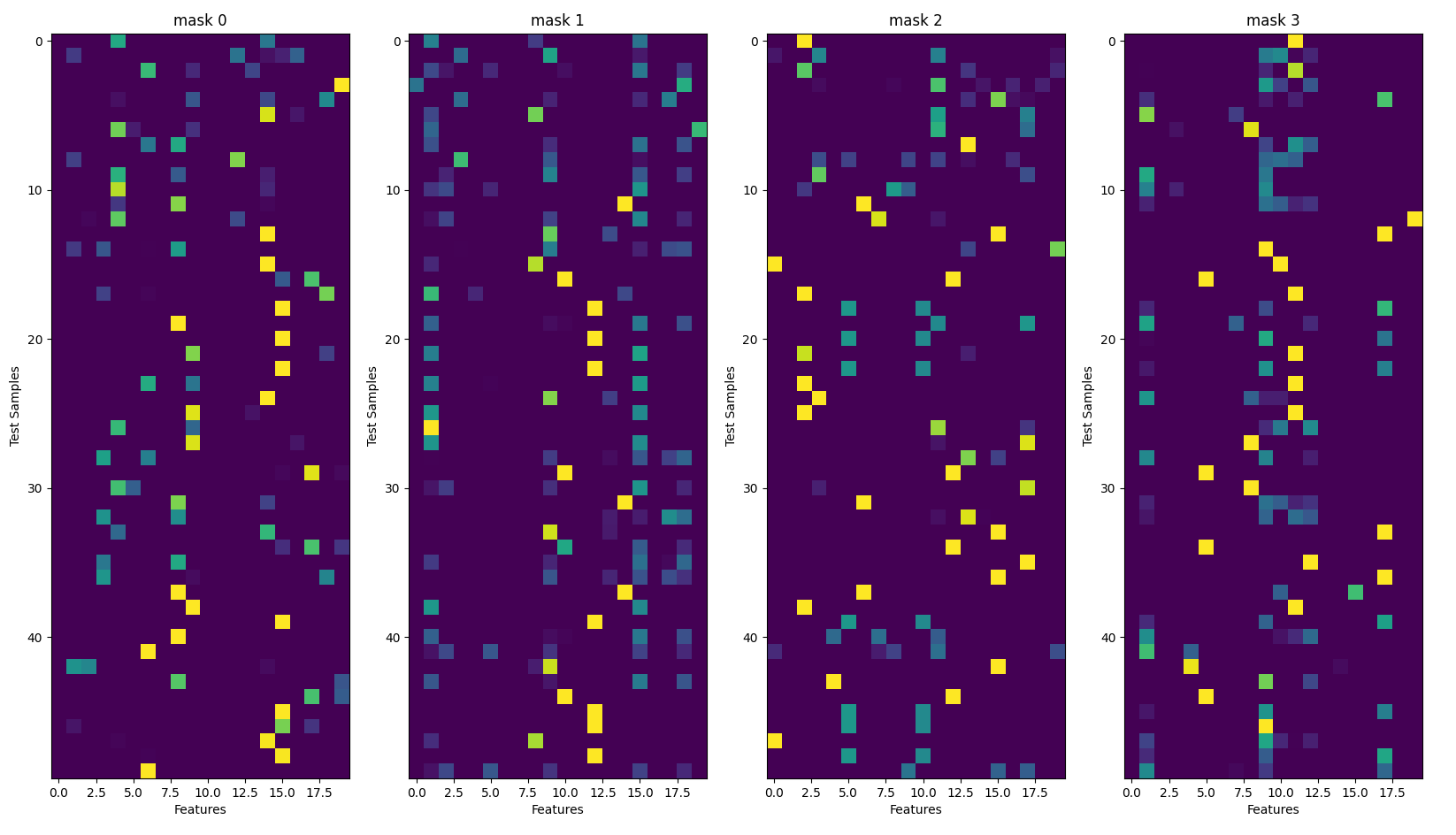}
        \caption*{(b) Original TabNet Model} 
    \end{minipage}
    
    \caption{(a) Best performing model using InterpreTabNet $r_M^* = 10,000,000,000,000$ with an accuracy of 72.96\% on the Blastchar Dataset. (b) The baseline performance using the Original TabNet model, attaining an accuracy of 76.22\%.}
    \label{fig:blastchar_comparison}
\end{figure}

\newpage
\subsubsection*{Diabetes \citep{diabetes}}

\begin{center}
    \begin{tabular}{lc}
        \toprule
        \textbf{Model} & \textbf{Test Accuracy (\%)} \\
        \midrule
        InterpreTabNet $(r_M^* = 100,000,000)$ & \textbf{$55.37\pm0.47$} \\
        Original TabNet & $56.91\pm0.53$ \\
        XGBoost & $61.44\pm0.32$ \\
        LightGBM & $60.87\pm0.39$ \\
        TabTransformer & $44.45\pm0.34$ \\
        MLP & $53.99\pm0.46$ \\
        \bottomrule
    \end{tabular}
\end{center}

\begin{figure}[htb!]
    \centering
    
    \begin{minipage}[b]{0.8\textwidth}
        \includegraphics[width=\textwidth]{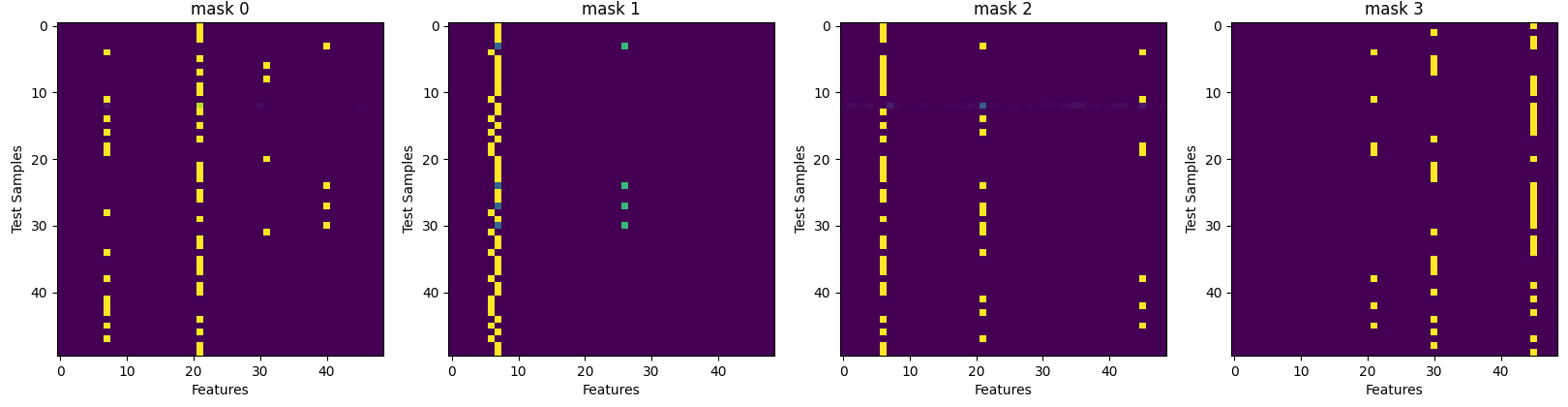}
        \caption*{(a) InterpreTabNet} 
    \end{minipage}
    \hfill
    \begin{minipage}[b]{0.8\textwidth}
        \includegraphics[width=\textwidth]{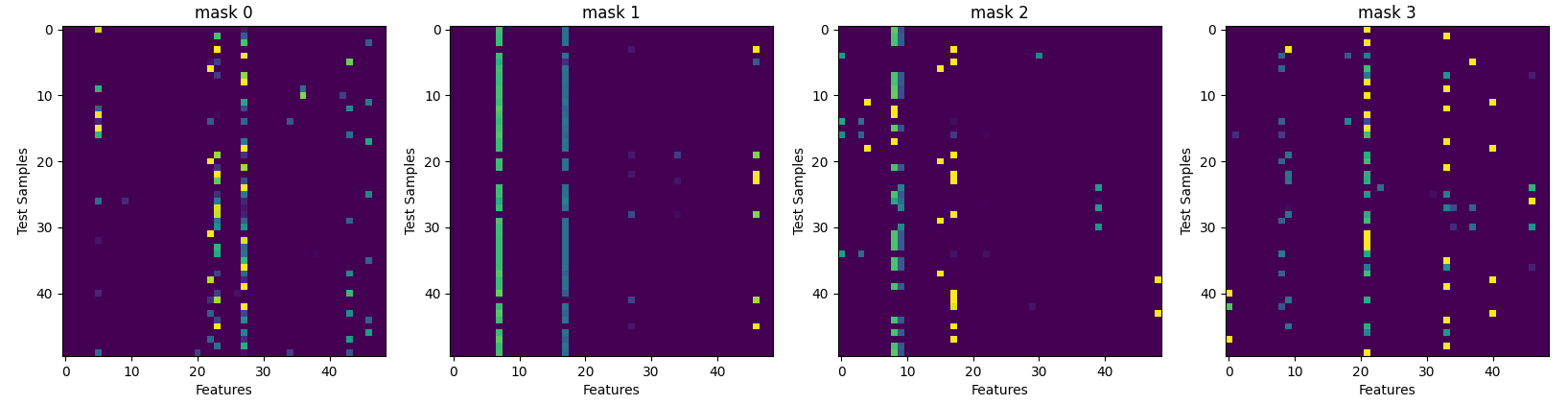}
        \caption*{(b) Original TabNet Model} 
    \end{minipage}
    
    \caption{(a) Best performing model using InterpreTabNet $r_M^* = 100,000,000$ with an accuracy of 55.38\% on the Diabetes Dataset. (b) The baseline performance using the Original TabNet model, attaining an accuracy of 56.91\%.}
    \label{fig:diabetes_comparison}
\end{figure}

\newpage
\subsection{InterpreTabNet: Prompts for All Datasets}
\label{appendix_prompt}

\subsubsection*{Adult Census Income}
Conduct aggregate analysis on the description of the following feature mask. Please output ONLY a dictionary and no other natural language generation when generating the sentence as shown in the in-context example below. Please use single-word classification that encapsulates the meaning of the features if possible.

The Adult Census Income dataset is considered which predicts whether an individual’s income exceeds \$50,000 per year. There are 50 test samples and 14 features. 

At the 0th step of feature selection, we observe mask 0 with the main features highlighted as 1, 5 and 7 which are workclass, marital status and relationship. At the 1st step of feature selection, we observe mask 1 with the main features highlighted as 1 and 5 which are workclass and marital status. At the 2nd step of feature selection, we observe mask 2 with the main features highlighted as 8, 10 and 12 which are race, capital gain and hours-per-week. Lastly, at the 3rd step of feature selection, we observe mask 3 with the main features highlighted as 1, 7 and 10 which are workclass, relationship and capital-gain.

Here are in-context examples for few-shot learning. 

The Poker Hand dataset is considered, which is a classification of poker hands into 10 distinct types based on attributes like rank of individual cards, suits, and card combinations. There are 50 test samples and 10 features. 

At the 0th step of feature selection, we observe mask 0 with the main feature highlighted as 3 which is the rank of the card 2. At the 1st step of feature selection, we observe mask 1 with the main feature highlighted as 7 which is the rank of card 3. At the 2nd step of feature selection, we observe mask 2 with the main feature highlighted as 6 which is the suit of card 3. Lastly, at the 3rd step of feature selection, we observe mask 3 with the main features highlighted as 3, 6, 7 and 10 which are the rank of cards 1 and 3 as well as the suits of cards 3 and 5.

Output: \{"Mask 0": "Initially, the rank of card 2 is recognized as a significant predictor in the classification of poker hands. The rank of individual cards seems crucial at this stage.", "Mask 1": "At this step, the rank of card 3 is highlighted, further emphasizing the importance of individual card ranks in determining poker hand types.", "Mask 2": "Here, besides the rank of card 3, the suit of the same card is considered, suggesting that both rank and suit play roles in classifying poker hands.", "Mask 3": "This step combines multiple features, both ranks (cards 1 and 3) and suits (cards 3 and 5), indicating that a combination of these factors provides a broader understanding of poker hand classifications.", "Aggregate": "The rank of cards, specifically card 2 and card 3, appears consistently across different masks, indicating their primary role in poker hand classification. Suits, particularly of card 3, also have importance but are introduced later in the feature selection. The combined importance of both ranks and suits in the final step suggests that while individual card ranks are pivotal in initial classifications, understanding the relationship between card ranks and their respective suits provides a more comprehensive insight into the poker hand types. The consistent presence of card 3's attributes (both rank and suit) underscores its pivotal role in determining poker hand categories."\}

The Forest Cover Type dataset is considered which is a classification of pixels into 7 forest cover types based on attributes such as elevation, aspect, slope, and more. There are 50 test samples and 54 features. 

At the 0th step of feature selection, we observe mask 0 with the main features highlighted as 24, 29 and 34 which are different soil types. At the 1st step of feature selection, we observe mask 1 with the main feature highlighted as 34 which is soil type. At the 2nd step of feature selection, we observe mask 2 with the main features highlighted as 10, 34 and 37 which are wilderness area and soil types. Lastly, at the 3rd step of feature selection, we observe mask 3 with the main features highlighted as 0 and 29 which are elevation, and soil type.

Output: \{"Mask 0": "The initial feature selection identifies that various soil types are significant predictors for the forest cover types. It appears that the characteristics of the soil play a vital role in determining the type of forest cover.", "Mask 1": "At this step, the model emphasizes even more on one particular soil type (34), suggesting it as a primary driver for the classification. This specific soil type may have unique characteristics making it a distinctive feature for certain forest cover types.", "Mask 2": "Here, besides the emphasis on soil type 34, another soil type (37) and wilderness area are considered. This suggests that both the nature of the terrain (wilderness) and the characteristics of the soil together contribute to the forest cover type.", "Mask 3": "This step introduces elevation as a significant predictor while still retaining a soil type. Elevation, combined with soil characteristics, might influence specific flora and subsequently determine forest cover types.", "Aggregate": "Soil type 34 is the most consistent feature across all the masks, appearing in every step, emphasizing its importance in the forest cover type classification. Another soil type, 29, is also consistent, appearing in both mask 0 and 3. The presence of different features across the steps, like wilderness area (mask 2) and elevation (mask 3), indicates that the landscape's nature and the physical height of the terrain are also important but might offer secondary or nuanced insights into forest cover types. The dominant predictor seems to be soil types, particularly soil type 34, suggesting that specific soil characteristics play a crucial role in determining forest cover types. However, aspects like elevation and the nature of the terrain (wilderness area) also contribute to the classification and shouldn't be overlooked."\}

\subsubsection*{Forest Cover Type}
Conduct aggregate analysis on the description of the following feature mask. Please output ONLY a dictionary and no other natural language generation when generating the sentence as shown in the in-context example below. Please use single-word classification that encapsulates the meaning of the features if possible.

The Forest Cover Type dataset is considered which is a classification of pixels into 7 forest cover types based on attributes such as elevation, aspect, slope, and more. There are 50 test samples and 54 features. 

At the 0th step of feature selection, we observe mask 0 with the main features highlighted as 24, 29 and 34 which are different soil types. At the 1st step of feature selection, we observe mask 1 with the main feature highlighted as 34 which is soil type. At the 2nd step of feature selection, we observe mask 2 with the main features highlighted as 10, 34 and 37 which are wilderness area and soil types. Lastly, at the 3rd step of feature selection, we observe mask 3 with the main features highlighted as 0 and 29 which are elevation, and soil type.

Here are in-context examples for few-shot learning. 

The Adult Census Income dataset is considered which predicts whether an individual’s income exceeds \$50,000 per year. There are 50 test samples and 14 features. 

At the 0th step of feature selection, we observe mask 0 with the main features highlighted as 1, 5 and 7 which are workclass, marital status and relationship. At the 1st step of feature selection, we observe mask 0 with the main features highlighted as 1 and 5 which are workclass and marital status. At the 2nd step of feature selection, we observe mask 2 with the main features highlighted as 8, 10 and 12 which are race, capital gain and hours-per-week. Lastly, at the 3rd step of feature selection, we observe mask 3 with the main features highlighted as 1, 7 and 10 which are workclass, relationship and capital-gain.

Output: \{"Mask 0": "This suggests that early in the feature selection process, the model finds that work-related factors (like the type of work) and personal relationships (both marital status and other relationships) are significant predictors for income exceeding \$50,000.", "Mask 1": "Here, the model emphasizes even more on work nature and marital status, suggesting these two as the primary drivers for the prediction. Relationship as a feature has been excluded in this step, which might indicate a possible correlation with marital status.", "Mask 2": "At this step, the model seems to be looking beyond just personal and work nature to more financial and demographic aspects. Capital gain, which is directly related to financial health, is considered along with hours-per-week, potentially pointing to a relationship between longer working hours and higher income. The inclusion of race might indicate underlying societal or economic factors at play.", "Mask 3": "This combination brings back features from both the initial and second masks. Workclass and relationship are revisited, which further underscores their importance. Capital gain remains consistent from the second step, emphasizing its importance as a predictor for income.", "Aggregate": "The features that consistently appear across multiple masks are: workclass (masks 0, 1, and 3), relationship (masks 0 and 3), and capital gain (masks 2 and 3). This highlights that these features play a pivotal role in predicting whether an individual’s income exceeds \$50,000 per year. Marital status is seen in the initial stages but disappears later, while hours-per-week and race come into play at the second step. This could indicate that while some features provide foundational predictive power, others may provide more nuanced insight or refinement to the model's predictions. Work-related aspects, personal relationships, and financial metrics seem to be the main predictors of an individual's income level in this dataset. It would be crucial to further analyze these features' exact impacts and understand any potential correlations among them for a more comprehensive understanding."\}

The Poker Hand dataset is considered, which is a classification of poker hands into 10 distinct types based on attributes like rank of individual cards, suits, and card combinations. There are 50 test samples and 10 features. 

At the 0th step of feature selection, we observe mask 0 with the main feature highlighted as 3 which is the rank of the card 2. At the 1st step of feature selection, we observe mask 1 with the main feature highlighted as 7 which is the rank of card 3. At the 2nd step of feature selection, we observe mask 2 with the main feature highlighted as 6 which is the suit of card 3. Lastly, at the 3rd step of feature selection, we observe mask 3 with the main features highlighted as 3, 6, 7 and 10 which are the rank of cards 1 and 3 as well as the suits of cards 3 and 5.

Output: \{"Mask 0": "Initially, the rank of card 2 is recognized as a significant predictor in the classification of poker hands. The rank of individual cards seems crucial at this stage.", "Mask 1": "At this step, the rank of card 3 is highlighted, further emphasizing the importance of individual card ranks in determining poker hand types.", "Mask 2": "Here, besides the rank of card 3, the suit of the same card is considered, suggesting that both rank and suit play roles in classifying poker hands.", "Mask 3": "This step combines multiple features, both ranks (cards 1 and 3) and suits (cards 3 and 5), indicating that a combination of these factors provides a broader understanding of poker hand classifications.", "Aggregate": "The rank of cards, specifically card 2 and card 3, appears consistently across different masks, indicating their primary role in poker hand classification. Suits, particularly of card 3, also have importance but are introduced later in the feature selection. The combined importance of both ranks and suits in the final step suggests that while individual card ranks are pivotal in initial classifications, understanding the relationship between card ranks and their respective suits provides a more comprehensive insight into the poker hand types. The consistent presence of card 3's attributes (both rank and suit) underscores its pivotal role in determining poker hand categories."\}

\subsubsection*{Poker Hand Prompt}
Conduct aggregate analysis on the description of the following feature mask. Please output ONLY a dictionary and no other natural language generation when generating the sentence as shown in the in-context example below. Please use single-word classification that encapsulates the meaning of the features if possible.

The Poker Hand dataset is considered, which is a classification of poker hands into 10 distinct types based on attributes like rank of individual cards, suits, and card combinations. There are 50 test samples and 10 features. 

At the 0th step of feature selection, we observe mask 0 with the main feature highlighted as 3 which is the rank of the card 2. At the 1st step of feature selection, we observe mask 1 with the main feature highlighted as 7 which is the rank of card 3. At the 2nd step of feature selection, we observe mask 2 with the main feature highlighted as 6 which is the suit of card 3. Lastly, at the 3rd step of feature selection, we observe mask 3 with the main features highlighted as 3, 6, 7 and 10 which are the rank of cards 1 and 3 as well as the suits of cards 3 and 5.

Here are in-context examples for few-shot learning. 

The Adult Census Income dataset is considered which predicts whether an individual’s income exceeds \$50,000 per year. There are 50 test samples and 14 features. At the 0th step of feature selection, we observe mask 0 with the main features highlighted as 1, 5 and 7 which are workclass, marital status and relationship. At the 1st step of feature selection, we observe mask 0 with the main features highlighted as 1 and 5 which are workclass and marital status. At the 2nd step of feature selection, we observe mask 2 with the main features highlighted as 8, 10 and 12 which are race, capital gain and hours-per-week. Lastly, at the 3rd step of feature selection, we observe mask 3 with the main features highlighted as 1, 7 and 10 which are workclass, relationship and capital-gain.

Output: \{"Mask 0": "This suggests that early in the feature selection process, the model finds that work-related factors (like the type of work) and personal relationships (both marital status and other relationships) are significant predictors for income exceeding \$50,000.", "Mask 1": "Here, the model emphasizes even more on work nature and marital status, suggesting these two as the primary drivers for the prediction. Relationship as a feature has been excluded in this step, which might indicate a possible correlation with marital status.", "Mask 2": "At this step, the model seems to be looking beyond just personal and work nature to more financial and demographic aspects. Capital gain, which is directly related to financial health, is considered along with hours-per-week, potentially pointing to a relationship between longer working hours and higher income. The inclusion of race might indicate underlying societal or economic factors at play.", "Mask 3": "This combination brings back features from both the initial and second masks. Workclass and relationship are revisited, which further underscores their importance. Capital gain remains consistent from the second step, emphasizing its importance as a predictor for income.", "Aggregate": "The features that consistently appear across multiple masks are: workclass (masks 0, 1, and 3), relationship (masks 0 and 3), and capital gain (masks 2 and 3). This highlights that these features play a pivotal role in predicting whether an individual’s income exceeds \$50,000 per year. Marital status is seen in the initial stages but disappears later, while hours-per-week and race come into play at the second step. This could indicate that while some features provide foundational predictive power, others may provide more nuanced insight or refinement to the model's predictions. Work-related aspects, personal relationships, and financial metrics seem to be the main predictors of an individual's income level in this dataset. It would be crucial to further analyze these features' exact impacts and understand any potential correlations among them for a more comprehensive understanding."\}

The Forest Cover Type dataset is considered which is a classification of pixels into 7 forest cover types based on attributes such as elevation, aspect, slope, and more. There are 50 test samples and 54 features. 

At the 0th step of feature selection, we observe mask 0 with the main features highlighted as 24, 29 and 34 which are different soil types. At the 1st step of feature selection, we observe mask 1 with the main feature highlighted as 34 which is soil type. At the 2nd step of feature selection, we observe mask 2 with the main features highlighted as 10, 34 and 37 which are wilderness area and soil types. Lastly, at the 3rd step of feature selection, we observe mask 3 with the main features highlighted as 0 and 29 which are elevation, and soil type.

Output: \{"Mask 0": "The initial feature selection identifies that various soil types are significant predictors for the forest cover types. It appears that the characteristics of the soil play a vital role in determining the type of forest cover.", "Mask 1": "At this step, the model emphasizes even more on one particular soil type (34), suggesting it as a primary driver for the classification. This specific soil type may have unique characteristics making it a distinctive feature for certain forest cover types.", "Mask 2": "Here, besides the emphasis on soil type 34, another soil type (37) and wilderness area are considered. This suggests that both the nature of the terrain (wilderness) and the characteristics of the soil together contribute to the forest cover type.", "Mask 3": "This step introduces elevation as a significant predictor while still retaining a soil type. Elevation, combined with soil characteristics, might influence specific flora and subsequently determine forest cover types.", "Aggregate": "Soil type 34 is the most consistent feature across all the masks, appearing in every step, emphasizing its importance in the forest cover type classification. Another soil type, 29, is also consistent, appearing in both mask 0 and 3. The presence of different features across the steps, like wilderness area (mask 2) and elevation (mask 3), indicates that the landscape's nature and the physical height of the terrain are also important but might offer secondary or nuanced insights into forest cover types. The dominant predictor seems to be soil types, particularly soil type 34, suggesting that specific soil characteristics play a crucial role in determining forest cover types. However, aspects like elevation and the nature of the terrain (wilderness area) also contribute to the classification and shouldn't be overlooked."\}

\subsubsection*{Mushroom Prompt}

Conduct aggregate analysis on the description of the following feature masks. Start off with an analysis of the individual masks, followed by an aggregate analysis of all masks combined. Please format the output into a dictionary as shown in the in-context examples. The output should only contain the formatted output, no other natural language generation is required.

The Mushroom dataset is considered, which is a classification of mushrooms into edible or poisonous categories based on attributes like cap shape, gill color, stalk length, and other morphological characteristics. There are 50 test samples and 22 features. 

At the 0th step of feature selection, we observe mask 0 with the main features highlighted as 5, 6 and 17 which are bruises, odor and veil-type. At the 1st step of feature selection, we observe mask 1 with the main feature highlighted as 5, 6, 11 which are bruises, odor and stalk-shape. At the 2nd step of feature selection, we observe mask 2 with the main features highlighted as 7, 14, 17 which are gill-attachment, stalk-surface-below-ring and veil-type. Lastly, at the 3rd step of feature selection, we observe mask 3 with the main features highlighted as 5, 6, 7 and 17 which are bruises, odor, gill-attachment and veil-type.

Here are in-context examples for few-shot learning. 

The Adult Census Income dataset is considered which predicts whether an individual’s income exceeds \$50,000 per year. There are 50 test samples and 14 features. At the 0th step of feature selection, we observe mask 0 with the main features highlighted as 1, 5 and 7 which are workclass, marital status and relationship. At the 1st step of feature selection, we observe mask 0 with the main features highlighted as 1 and 5 which are workclass and marital status. At the 2nd step of feature selection, we observe mask 2 with the main features highlighted as 8, 10 and 12 which are race, capital gain and hours-per-week. Lastly, at the 3rd step of feature selection, we observe mask 3 with the main features highlighted as 1, 7 and 10 which are workclass, relationship and capital-gain.

Output: \{"Mask 0": "This suggests that early in the feature selection process, the model finds that work-related factors (like the type of work) and personal relationships (both marital status and other relationships) are significant predictors for income exceeding \$50,000.", "Mask 1": "Here, the model emphasizes even more on work nature and marital status, suggesting these two as the primary drivers for the prediction. Relationship as a feature has been excluded in this step, which might indicate a possible correlation with marital status.", "Mask 2": "At this step, the model seems to be looking beyond just personal and work nature to more financial and demographic aspects. Capital gain, which is directly related to financial health, is considered along with hours-per-week, potentially pointing to a relationship between longer working hours and higher income. The inclusion of race might indicate underlying societal or economic factors at play.", "Mask 3": "This combination brings back features from both the initial and second masks. Workclass and relationship are revisited, which further underscores their importance. Capital gain remains consistent from the second step, emphasizing its importance as a predictor for income.", "Aggregate": "The features that consistently appear across multiple masks are: workclass (masks 0, 1, and 3), relationship (masks 0 and 3), and capital gain (masks 2 and 3). This highlights that these features play a pivotal role in predicting whether an individual’s income exceeds \$50,000 per year. Marital status is seen in the initial stages but disappears later, while hours-per-week and race come into play at the second step. This could indicate that while some features provide foundational predictive power, others may provide more nuanced insight or refinement to the model's predictions. Work-related aspects, personal relationships, and financial metrics seem to be the main predictors of an individual's income level in this dataset. It would be crucial to further analyze these features' exact impacts and understand any potential correlations among them for a more comprehensive understanding."\}

The Forest Cover Type dataset is considered which is a classification of pixels into 7 forest cover types based on attributes such as elevation, aspect, slope, and more. There are 50 test samples and 54 features. 

At the 0th step of feature selection, we observe mask 0 with the main features highlighted as 24, 29 and 34 which are different soil types. At the 1st step of feature selection, we observe mask 1 with the main feature highlighted as 34 which is soil type. At the 2nd step of feature selection, we observe mask 2 with the main features highlighted as 10, 34 and 37 which are wilderness area and soil types. Lastly, at the 3rd step of feature selection, we observe mask 3 with the main features highlighted as 0 and 29 which are elevation, and soil type.

Output: \{"Mask 0": "The initial feature selection identifies that various soil types are significant predictors for the forest cover types. It appears that the characteristics of the soil play a vital role in determining the type of forest cover.", "Mask 1": "At this step, the model emphasizes even more on one particular soil type (34), suggesting it as a primary driver for the classification. This specific soil type may have unique characteristics making it a distinctive feature for certain forest cover types.", "Mask 2": "Here, besides the emphasis on soil type 34, another soil type (37) and wilderness area are considered. This suggests that both the nature of the terrain (wilderness) and the characteristics of the soil together contribute to the forest cover type.", "Mask 3": "This step introduces elevation as a significant predictor while still retaining a soil type. Elevation, combined with soil characteristics, might influence specific flora and subsequently determine forest cover types.", "Aggregate": "Soil type 34 is the most consistent feature across all the masks, appearing in every step, emphasizing its importance in the forest cover type classification. Another soil type, 29, is also consistent, appearing in both mask 0 and 3. The presence of different features across the steps, like wilderness area (mask 2) and elevation (mask 3), indicates that the landscape's nature and the physical height of the terrain are also important but might offer secondary or nuanced insights into forest cover types. The dominant predictor seems to be soil types, particularly soil type 34, suggesting that specific soil characteristics play a crucial role in determining forest cover types. However, aspects like elevation and the nature of the terrain (wilderness area) also contribute to the classification and shouldn't be overlooked."\}

\subsubsection*{Blastchar Prompt}

Conduct aggregate analysis on the description of the following feature masks. Start off with an analysis of the individual masks, followed by an aggregate analysis of all masks combined. Please format the output into a dictionary as shown in the in-context examples. The output should only contain the formatted output, no other natural language generation is required.

The BlastChar Telco Customer Churn dataset is considered, which is a classification of customers into retained or churned categories based on attributes like gender, seniority, tenure, service subscriptions, contract type, billing methods, and charges, among others. There are 50 test samples and 21 features.

At the 0th step of feature selection, we observe mask 0 with the main features highlighted as 13, 16, and 17 which are StreamingTV, PaperlessBilling and PaymentMethod. At the 1st step of feature selection, we observe mask 1 with the main feature highlighted as 2 and 16 which are SeniorCitizen and PaperlessBilling. At the 2nd step of feature selection, we observe mask 2 with the main features highlighted as 8, 11, and 18 which are InternetService, DeviceProtection, and MonthlyCharges. Lastly, at the 3rd step of feature selection, we observe mask 3 with the main features highlighted as 3, 11, and 17 which are Partner, DeviceProtection, and PaymentMethod.

Here are in-context examples for few-shot learning. 

The Adult Census Income dataset is considered which predicts whether an individual’s income exceeds \$50,000 per year. There are 50 test samples and 14 features. At the 0th step of feature selection, we observe mask 0 with the main features highlighted as 1, 5 and 7 which are workclass, marital status and relationship. At the 1st step of feature selection, we observe mask 0 with the main features highlighted as 1 and 5 which are workclass and marital status. At the 2nd step of feature selection, we observe mask 2 with the main features highlighted as 8, 10 and 12 which are race, capital gain and hours-per-week. Lastly, at the 3rd step of feature selection, we observe mask 3 with the main features highlighted as 1, 7 and 10 which are workclass, relationship and capital-gain.

Output: \{"Mask 0": "This suggests that early in the feature selection process, the model finds that work-related factors (like the type of work) and personal relationships (both marital status and other relationships) are significant predictors for income exceeding \$50,000.", "Mask 1": "Here, the model emphasizes even more on work nature and marital status, suggesting these two as the primary drivers for the prediction. Relationship as a feature has been excluded in this step, which might indicate a possible correlation with marital status.", "Mask 2": "At this step, the model seems to be looking beyond just personal and work nature to more financial and demographic aspects. Capital gain, which is directly related to financial health, is considered along with hours-per-week, potentially pointing to a relationship between longer working hours and higher income. The inclusion of race might indicate underlying societal or economic factors at play.", "Mask 3": "This combination brings back features from both the initial and second masks. Workclass and relationship are revisited, which further underscores their importance. Capital gain remains consistent from the second step, emphasizing its importance as a predictor for income.", "Aggregate": "The features that consistently appear across multiple masks are: workclass (masks 0, 1, and 3), relationship (masks 0 and 3), and capital gain (masks 2 and 3). This highlights that these features play a pivotal role in predicting whether an individual’s income exceeds \$50,000 per year. Marital status is seen in the initial stages but disappears later, while hours-per-week and race come into play at the second step. This could indicate that while some features provide foundational predictive power, others may provide more nuanced insight or refinement to the model's predictions. Work-related aspects, personal relationships, and financial metrics seem to be the main predictors of an individual's income level in this dataset. It would be crucial to further analyze these features' exact impacts and understand any potential correlations among them for a more comprehensive understanding."\}

The Forest Cover Type dataset is considered which is a classification of pixels into 7 forest cover types based on attributes such as elevation, aspect, slope, and more. There are 50 test samples and 54 features. 

At the 0th step of feature selection, we observe mask 0 with the main features highlighted as 24, 29 and 34 which are different soil types. At the 1st step of feature selection, we observe mask 1 with the main feature highlighted as 34 which is soil type. At the 2nd step of feature selection, we observe mask 2 with the main features highlighted as 10, 34 and 37 which are wilderness area and soil types. Lastly, at the 3rd step of feature selection, we observe mask 3 with the main features highlighted as 0 and 29 which are elevation, and soil type.

Output: \{"Mask 0": "The initial feature selection identifies that various soil types are significant predictors for the forest cover types. It appears that the characteristics of the soil play a vital role in determining the type of forest cover.", "Mask 1": "At this step, the model emphasizes even more on one particular soil type (34), suggesting it as a primary driver for the classification. This specific soil type may have unique characteristics making it a distinctive feature for certain forest cover types.", "Mask 2": "Here, besides the emphasis on soil type 34, another soil type (37) and wilderness area are considered. This suggests that both the nature of the terrain (wilderness) and the characteristics of the soil together contribute to the forest cover type.", "Mask 3": "This step introduces elevation as a significant predictor while still retaining a soil type. Elevation, combined with soil characteristics, might influence specific flora and subsequently determine forest cover types.", "Aggregate": "Soil type 34 is the most consistent feature across all the masks, appearing in every step, emphasizing its importance in the forest cover type classification. Another soil type, 29, is also consistent, appearing in both mask 0 and 3. The presence of different features across the steps, like wilderness area (mask 2) and elevation (mask 3), indicates that the landscape's nature and the physical height of the terrain are also important but might offer secondary or nuanced insights into forest cover types. The dominant predictor seems to be soil types, particularly soil type 34, suggesting that specific soil characteristics play a crucial role in determining forest cover types. However, aspects like elevation and the nature of the terrain (wilderness area) also contribute to the classification and shouldn't be overlooked."\}

\subsubsection*{Diabetes Prompt}

Conduct aggregate analysis on the description of the following feature masks. Start off with an analysis of the individual masks, followed by an aggregate analysis of all masks combined. Please format the output into a dictionary as shown in the in-context examples. The output should only contain the formatted output, no other natural language generation is required.

The Diabetes 130-US hospitals for years 1999-2008 dataset is considered, which is a classification of patient encounters into readmitted or not readmitted categories based on attributes like the number of laboratory tests performed, the number of medications prescribed, diagnoses, and other clinical and administrative data. There are 50 test samples and 50 features.

At the 0th step of feature selection, we observe mask 0 with the main features highlighted as 7, 21, 31, and 40 which are discharge disposition id, number diagnoses, glyburide, and citoglipton. At the 1st step of feature selection, we observe mask 1 with the main feature highlighted as 6 and 7 which are admission type id and discharge disposition id. At the 2nd step of feature selection, we observe mask 2 with the main features highlighted as 6, 21, and 45 which are admission type id, number diagnoses, and metformin. Lastly, at the 3rd step of feature selection, we observe mask 3 with the main features highlighted as 21, 30, 45 which are number diagnoses, glipizide, and metformin.

Here are in-context examples for few-shot learning. 

The Adult Census Income dataset is considered which predicts whether an individual’s income exceeds \$50,000 per year. There are 50 test samples and 14 features. At the 0th step of feature selection, we observe mask 0 with the main features highlighted as 1, 5 and 7 which are workclass, marital status and relationship. At the 1st step of feature selection, we observe mask 0 with the main features highlighted as 1 and 5 which are workclass and marital status. At the 2nd step of feature selection, we observe mask 2 with the main features highlighted as 8, 10 and 12 which are race, capital gain and hours-per-week. Lastly, at the 3rd step of feature selection, we observe mask 3 with the main features highlighted as 1, 7 and 10 which are workclass, relationship and capital-gain.

Output: \{"Mask 0": "This suggests that early in the feature selection process, the model finds that work-related factors (like the type of work) and personal relationships (both marital status and other relationships) are significant predictors for income exceeding \$50,000.", "Mask 1": "Here, the model emphasizes even more on work nature and marital status, suggesting these two as the primary drivers for the prediction. Relationship as a feature has been excluded in this step, which might indicate a possible correlation with marital status.", "Mask 2": "At this step, the model seems to be looking beyond just personal and work nature to more financial and demographic aspects. Capital gain, which is directly related to financial health, is considered along with hours-per-week, potentially pointing to a relationship between longer working hours and higher income. The inclusion of race might indicate underlying societal or economic factors at play.", "Mask 3": "This combination brings back features from both the initial and second masks. Workclass and relationship are revisited, which further underscores their importance. Capital gain remains consistent from the second step, emphasizing its importance as a predictor for income.", "Aggregate": "The features that consistently appear across multiple masks are: workclass (masks 0, 1, and 3), relationship (masks 0 and 3), and capital gain (masks 2 and 3). This highlights that these features play a pivotal role in predicting whether an individual’s income exceeds \$50,000 per year. Marital status is seen in the initial stages but disappears later, while hours-per-week and race come into play at the second step. This could indicate that while some features provide foundational predictive power, others may provide more nuanced insight or refinement to the model's predictions. Work-related aspects, personal relationships, and financial metrics seem to be the main predictors of an individual's income level in this dataset. It would be crucial to further analyze these features' exact impacts and understand any potential correlations among them for a more comprehensive understanding."\}

The Forest Cover Type dataset is considered which is a classification of pixels into 7 forest cover types based on attributes such as elevation, aspect, slope, and more. There are 50 test samples and 54 features. 

At the 0th step of feature selection, we observe mask 0 with the main features highlighted as 24, 29 and 34 which are different soil types. At the 1st step of feature selection, we observe mask 1 with the main feature highlighted as 34 which is soil type. At the 2nd step of feature selection, we observe mask 2 with the main features highlighted as 10, 34 and 37 which are wilderness area and soil types. Lastly, at the 3rd step of feature selection, we observe mask 3 with the main features highlighted as 0 and 29 which are elevation, and soil type.

Output: \{"Mask 0": "The initial feature selection identifies that various soil types are significant predictors for the forest cover types. It appears that the characteristics of the soil play a vital role in determining the type of forest cover.", "Mask 1": "At this step, the model emphasizes even more on one particular soil type (34), suggesting it as a primary driver for the classification. This specific soil type may have unique characteristics making it a distinctive feature for certain forest cover types.", "Mask 2": "Here, besides the emphasis on soil type 34, another soil type (37) and wilderness area are considered. This suggests that both the nature of the terrain (wilderness) and the characteristics of the soil together contribute to the forest cover type.", "Mask 3": "This step introduces elevation as a significant predictor while still retaining a soil type. Elevation, combined with soil characteristics, might influence specific flora and subsequently determine forest cover types.", "Aggregate": "Soil type 34 is the most consistent feature across all the masks, appearing in every step, emphasizing its importance in the forest cover type classification. Another soil type, 29, is also consistent, appearing in both mask 0 and 3. The presence of different features across the steps, like wilderness area (mask 2) and elevation (mask 3), indicates that the landscape's nature and the physical height of the terrain are also important but might offer secondary or nuanced insights into forest cover types. The dominant predictor seems to be soil types, particularly soil type 34, suggesting that specific soil characteristics play a crucial role in determining forest cover types. However, aspects like elevation and the nature of the terrain (wilderness area) also contribute to the classification and shouldn't be overlooked."\}

\subsubsection*{Higgs Prompt}

Conduct aggregate analysis on the description of the following feature mask. Please output ONLY a dictionary and no other natural language generation when generating the sentence as shown in the in-context example below. Please use single-word classification that encapsulates the meaning of the features if possible. 

The Higgs dataset is considered, which is a classification of events into signal and background processes based on attributes like jet kinematics, lepton momentum, and other high-energy physics variables. There are 50 test samples and 28 features. At the 0th step of feature selection, we observe mask 0 with the main features highlighted as 22 and 25 which are the invariant mass of the lepton, the missing energy, and the leading jet, as well as the four leading jets. At the 1st step of feature selection, we observe mask 1 with the main feature highlighted as 6 which is the Azimuthal angle for the second leading jet. At the 2nd step of feature selection, we observe mask 2 with the main features highlighted as 15 and 25 which are the transverse momentum for the second leading jet and the invariant mass of the lepton, the missing energy, and the leading jet.. Lastly, at the 3rd step of feature selection, we observe mask 3 with the main features highlighted as 8 and 16 which are the B-tagging score for the second leading jet and the pseudorapidity of the lepton. 

Here are in-context examples for few-shot learning. 

The Adult Census Income dataset is considered which predicts whether an individual’s income exceeds \$50,000 per year. There are 50 test samples and 14 features. At the 0th step of feature selection, we observe mask 0 with the main features highlighted as 1, 5 and 7 which are workclass, marital status and relationship. At the 1st step of feature selection, we observe mask 0 with the main features highlighted as 1 and 5 which are workclass and marital status. At the 2nd step of feature selection, we observe mask 2 with the main features highlighted as 8, 10 and 12 which are race, capital gain and hours-per-week. Lastly, at the 3rd step of feature selection, we observe mask 3 with the main features highlighted as 1, 7 and 10 which are workclass, relationship and capital-gain.

Output: \{"Mask 0": "This suggests that early in the feature selection process, the model finds that work-related factors (like the type of work) and personal relationships (both marital status and other relationships) are significant predictors for income exceeding \$50,000.", "Mask 1": "Here, the model emphasizes even more on work nature and marital status, suggesting these two as the primary drivers for the prediction. Relationship as a feature has been excluded in this step, which might indicate a possible correlation with marital status.", "Mask 2": "At this step, the model seems to be looking beyond just personal and work nature to more financial and demographic aspects. Capital gain, which is directly related to financial health, is considered along with hours-per-week, potentially pointing to a relationship between longer working hours and higher income. The inclusion of race might indicate underlying societal or economic factors at play.", "Mask 3": "This combination brings back features from both the initial and second masks. Workclass and relationship are revisited, which further underscores their importance. Capital gain remains consistent from the second step, emphasizing its importance as a predictor for income.", "Aggregate": "The features that consistently appear across multiple masks are: workclass (masks 0, 1, and 3), relationship (masks 0 and 3), and capital gain (masks 2 and 3). This highlights that these features play a pivotal role in predicting whether an individual’s income exceeds \$50,000 per year. Marital status is seen in the initial stages but disappears later, while hours-per-week and race come into play at the second step. This could indicate that while some features provide foundational predictive power, others may provide more nuanced insight or refinement to the model's predictions. Work-related aspects, personal relationships, and financial metrics seem to be the main predictors of an individual's income level in this dataset. It would be crucial to further analyze these features' exact impacts and understand any potential correlations among them for a more comprehensive understanding."\}

The Forest Cover Type dataset is considered which is a classification of pixels into 7 forest cover types based on attributes such as elevation, aspect, slope, and more. There are 50 test samples and 54 features. 

At the 0th step of feature selection, we observe mask 0 with the main features highlighted as 24, 29 and 34 which are different soil types. At the 1st step of feature selection, we observe mask 1 with the main feature highlighted as 34 which is soil type. At the 2nd step of feature selection, we observe mask 2 with the main features highlighted as 10, 34 and 37 which are wilderness area and soil types. Lastly, at the 3rd step of feature selection, we observe mask 3 with the main features highlighted as 0 and 29 which are elevation, and soil type.

Output: \{"Mask 0": "The initial feature selection identifies that various soil types are significant predictors for the forest cover types. It appears that the characteristics of the soil play a vital role in determining the type of forest cover.", "Mask 1": "At this step, the model emphasizes even more on one particular soil type (34), suggesting it as a primary driver for the classification. This specific soil type may have unique characteristics making it a distinctive feature for certain forest cover types.", "Mask 2": "Here, besides the emphasis on soil type 34, another soil type (37) and wilderness area are considered. This suggests that both the nature of the terrain (wilderness) and the characteristics of the soil together contribute to the forest cover type.", "Mask 3": "This step introduces elevation as a significant predictor while still retaining a soil type. Elevation, combined with soil characteristics, might influence specific flora and subsequently determine forest cover types.", "Aggregate": "Soil type 34 is the most consistent feature across all the masks, appearing in every step, emphasizing its importance in the forest cover type classification. Another soil type, 29, is also consistent, appearing in both mask 0 and 3. The presence of different features across the steps, like wilderness area (mask 2) and elevation (mask 3), indicates that the landscape's nature and the physical height of the terrain are also important but might offer secondary or nuanced insights into forest cover types. The dominant predictor seems to be soil types, particularly soil type 34, suggesting that specific soil characteristics play a crucial role in determining forest cover types. However, aspects like elevation and the nature of the terrain (wilderness area) also contribute to the classification and shouldn't be overlooked."\}

\newpage
\subsection{InterpreTabNet: GPT-4 Output Text on All Datasets for Mask Interpretation}
\label{appendix_output}

\begin{center}
\begin{longtable}{cp{10cm}} 
\toprule
 \textbf{Dataset} & \textbf{Output from GPT-4} \\  
\midrule
 Adult Census Income & \{"Mask 0": "This suggests that early in the feature selection process, the model finds that work-related factors (like the type of work) and personal relationships (both marital status and other relationships) are significant predictors for income exceeding \$50,000.", 
 
 "Mask 1": "Here, the model emphasizes even more on work nature and marital status, suggesting these two as the primary drivers for the prediction. Relationship as a feature has been excluded in this step, which might indicate a possible correlation with marital status.", 
 
 "Mask 2": "At this step, the model seems to be looking beyond just personal and work nature to more financial and demographic aspects. Capital gain, which is directly related to financial health, is considered along with hours-per-week, potentially pointing to a relationship between longer working hours and higher income. The inclusion of race might indicate underlying societal or economic factors at play.", 
 
 "Mask 3": "This combination brings back features from both the initial and second masks. Workclass and relationship are revisited, which further underscores their importance. Capital gain remains consistent from the second step, emphasizing its importance as a predictor for income.", 
 
 "Aggregate": "The features that consistently appear across multiple masks are: workclass (masks 0, 1, and 3), relationship (masks 0 and 3), and capital gain (masks 2 and 3). This highlights that these features play a pivotal role in predicting whether an individual’s income exceeds \$50,000 per year. Marital status is seen in the initial stages but disappears later, while hours-per-week and race come into play at the second step. This could indicate that while some features provide foundational predictive power, others may provide more nuanced insight or refinement to the model's predictions. Work-related aspects, personal relationships, and financial metrics seem to be the main predictors of an individual's income level in this dataset. It would be crucial to further analyze these features' exact impacts and understand any potential correlations among them for a more comprehensive understanding."\} \\
\midrule
\newpage
\midrule
 Forest Cover Type & \{"Mask 0": "The initial feature selection identifies that various soil types are significant predictors for the forest cover types. It appears that the characteristics of the soil play a vital role in determining the type of forest cover.", 
 
 "Mask 1": "At this step, the model emphasizes even more on one particular soil type (34), suggesting it as a primary driver for the classification. This specific soil type may have unique characteristics making it a distinctive feature for certain forest cover types.", 
 
 "Mask 2": "Here, besides the emphasis on soil type 34, another soil type (37) and wilderness area are considered. This suggests that both the nature of the terrain (wilderness) and the characteristics of the soil together contribute to the forest cover type.", 
 
 "Mask 3": "This step introduces elevation as a significant predictor while still retaining a soil type. Elevation, combined with soil characteristics, might influence specific flora and subsequently determine forest cover types.", 
 
 "Aggregate": "Soil type 34 is the most consistent feature across all the masks, appearing in every step, emphasizing its importance in the forest cover type classification. Another soil type, 29, is also consistent, appearing in both mask 0 and 3. The presence of different features across the steps, like wilderness area (mask 2) and elevation (mask 3), indicates that the landscape's nature and the physical height of the terrain are also important but might offer secondary or nuanced insights into forest cover types. The dominant predictor seems to be soil types, particularly soil type 34, suggesting that specific soil characteristics play a crucial role in determining forest cover types. However, aspects like elevation and the nature of the terrain (wilderness area) also contribute to the classification and shouldn't be overlooked."\} \\
\midrule
 Poker Hand & \{"Mask 0": "Initially, the rank of card 2 is recognized as a significant predictor in the classification of poker hands. The rank of individual cards seems crucial at this stage.", 
 
 "Mask 1": "At this step, the rank of card 3 is highlighted, further emphasizing the importance of individual card ranks in determining poker hand types.", 
 
 "Mask 2": "Here, besides the rank of card 3, the suit of the same card is considered, suggesting that both rank and suit play roles in classifying poker hands.", 
 
 "Mask 3": "This step combines multiple features, both ranks (cards 1 and 3) and suits (cards 3 and 5), indicating that a combination of these factors provides a broader understanding of poker hand classifications.", 
 
 "Aggregate": "The rank of cards, specifically card 2 and card 3, appears consistently across different masks, indicating their primary role in poker hand classification. Suits, particularly of card 3, also have importance but are introduced later in the feature selection. The combined importance of both ranks and suits in the final step suggests that while individual card ranks are pivotal in initial classifications, understanding the relationship between card ranks and their respective suits provides a more comprehensive insight into the poker hand types. The consistent presence of card 3's attributes (both rank and suit) underscores its pivotal role in determining poker hand categories."\}
 \\
\midrule
\newpage
\midrule
 Mushroom & 
\{"Mask 0": "At the outset of the feature selection process, the model identifies physical attributes such as bruises, odor, and veil-type as significant indicators for classifying mushrooms into edible or poisonous categories. This suggests that visible physical changes and inherent characteristics play an essential role in the initial classification stages.", 

"Mask 1": "In this step, stalk-shape is introduced while retaining bruises and odor from the previous mask. Here, the model seems to emphasize the physical appearance (bruises and stalk-shape) in conjunction with inherent characteristics (odor) for the classification. Veil-type, although significant in the initial step, is omitted here, hinting at its reduced relevance or a possible correlation with other selected features.", 

"Mask 2": "The model shifts its attention here, emphasizing gill-attachment and stalk-surface-below-ring while still retaining the veil-type. This shift indicates the significance of features directly related to the mushroom's growth structure and its underlying layers. The absence of previously highlighted features suggests a refined approach to feature importance at this step.", 

"Mask 3": "This combination merges the significant features from the earlier masks, highlighting bruises, odor, gill-attachment, and veil-type. The recurrence of these features underscores their consistent importance in classifying mushrooms. The introduction of gill-attachment here, combined with its presence in the previous mask, further emphasizes its role.", 

"Aggregate": "The features that consistently appear across multiple masks are: bruises (masks 0, 1, and 3), odor (masks 0, 1, and 3), and veil-type (masks 0, 2, and 3). This suggests that these attributes play a crucial role in distinguishing between edible and poisonous mushrooms in this dataset. Gill-attachment, although appearing in the latter stages (masks 2 and 3), indicates its significance in the classification process. The dataset appears to highlight the importance of physical attributes (like bruises and stalk-shape), inherent characteristics (like odor), and growth-related features (like gill-attachment and stalk-surface-below-ring) in determining the edibility of mushrooms. Further investigation into these features' interactions and exact impacts would provide deeper insights into the classification process."\} 
 \\
\midrule
\newpage
\midrule
 Blastchar & \{
"Mask 0": "The initial feature selection suggests that aspects related to customer service preferences such as StreamingTV and billing and payment methods like PaperlessBilling and PaymentMethod are significant predictors for customer churn. These features might be pointing towards a customer's comfort and adaptability with the service's digital offerings.",

"Mask 1": "In this step, demographic aspects come into play with the inclusion of SeniorCitizen, suggesting that age or seniority might have a bearing on customer churn. The continued emphasis on PaperlessBilling also underlines its importance in this prediction.",

"Mask 2": "At this juncture, the model considers the type of service subscribed by the customer, as seen from the inclusion of InternetService and DeviceProtection. MonthlyCharges also comes into the picture, indicating that the amount being charged could play a role in a customer's decision to stay or leave the service. It's possible that customers assess the value they receive from these services against what they are being charged.",

"Mask 3": "In this combination, there's a mix of service-related (DeviceProtection), demographic (Partner), and payment (PaymentMethod) features. The appearance of Partner might indicate the influence of family decisions or joint subscriptions on customer churn.",

"Aggregate": "PaperlessBilling (masks 0 and 1) and DeviceProtection (masks 2 and 3) consistently appear in multiple steps, highlighting their significance in predicting customer churn. PaymentMethod also appears in masks 0 and 3, suggesting the importance of billing methods in this context. On the other hand, unique features like SeniorCitizen and InternetService give insight into the potential impact of demographics and service type respectively. In essence, customer service preferences, billing methods, and charges, along with certain demographic elements, are key in predicting customer churn in this dataset. For a comprehensive understanding, a deeper dive into the interaction of these features and their respective weights in the prediction model would be beneficial."
\}
\\
\midrule
\newpage
\midrule
Diabetes & 
\{
"Mask 0": "In the initial step, the model identifies discharge disposition id, number of diagnoses, and specific medications (glyburide, and citoglipton) as key features. This suggests a combination of administrative data (discharge process), clinical data (diagnoses), and prescribed medications play a significant role in predicting patient readmission.",

"Mask 1": "Here, the model focuses on the patient’s admission type and discharge disposition, emphasizing the importance of the hospital admission and discharge process in determining the likelihood of readmission. The absence of clinical and medication features in this step indicates a potential higher influence of administrative processes at this stage of feature selection.",

"Mask 2": "The model reintroduces clinical data (number of diagnoses) and adds a new medication (metformin), while retaining admission type id. This combination underscores a nuanced interplay between the nature of hospital admission, the complexity of patient’s health conditions, and specific treatments in predicting readmissions.",

"Mask 3": "This step focuses on the number of diagnoses and two medications (glipizide and metformin). The consistency of number of diagnoses across masks highlights its pivotal role, while the introduction of glipizide and retention of metformin point towards an evolving emphasis on medication in conjunction with clinical data.",

"Aggregate": "Number diagnoses (masks 0, 2, and 3) and discharge disposition id (masks 0 and 1) are recurrent, underscoring their persistent importance in predicting readmissions. Admission type id also appears consistently (masks 1 and 2), linking the administrative process tightly with readmission predictions. The changing set of medications across the masks (glyburide, citoglipton, metformin, glipizide) points to the dynamic nature of treatment's role in the prediction, warranting a closer look into specific medication combinations and their interactions with clinical and administrative data. The aggregate analysis reveals a complex interplay between administrative processes, clinical data, and medication in predicting patient readmissions, suggesting a multifaceted approach for model refinement and interpretation."
\}
\\
\midrule
\newpage
\midrule
 Higgs & \{
"Mask 0": "The initial feature selection emphasizes on variables related to kinematics and energy in the Higgs dataset, particularly invariant mass of the lepton, missing energy, and jets, implying these measurements play a fundamental role in distinguishing signal from background processes.",

"Mask 1": "At this step, the model focuses on the Azimuthal angle for the second leading jet, suggesting that the orientation or direction of this jet may have a unique significance in the classification process.",

"Mask 2": "Here, the model considers the transverse momentum of the second leading jet alongside the invariant mass of the lepton, missing energy, and leading jet. This suggests an interplay between momentum and energy variables as critical determinants in the classification.",

"Mask 3": "In this step, the B-tagging score for the second leading jet and the pseudorapidity of the lepton are highlighted. B-tagging, used to identify the presence of a b-quark, combined with the pseudorapidity, indicates the model's attempt to understand particle behavior and properties for classification.",

"Aggregate": "Over the feature selection steps, emphasis is consistently placed on high-energy physics variables, particularly those related to jets and leptons. The invariant mass of the lepton and missing energy are repeated features, appearing in both masks 0 and 2, showcasing their importance in the classification process. Momentum and angular measurements, like the Azimuthal angle and transverse momentum, also play a pivotal role. As the steps progress, there's an evident shift from energy-related features to more particle-specific characteristics, such as B-tagging. Overall, the classification in the Higgs dataset relies heavily on a combination of energy measurements, momentum, and particle properties."\}
\\
\bottomrule
\end{longtable}
\end{center}

\newpage
\subsection{TabNet: Prompt and GPT-4 Output on Adult Dataset}
\label{tabnet_prompt}

\textbf{Prompt} Conduct aggregate analysis on the description of the following feature mask. Please output ONLY a dictionary and no other natural language generation when generating the sentence, as shown in the in-context example below. Please use single-word classification that encapsulates the meaning of the features if possible.

The Adult Census Income dataset is considered which predicts whether an individual’s income exceeds \$50,000 per year. There are 50 test samples and 14 features. 

At the 0th step of feature selection, we observe mask 0 with barely any main features highlighted. At most feature 3 which is education is somewhat highlighted. At the 1st step of feature selection, we observe mask 1 with the main features highlighted as somewhat 7 and mainly 11 which are relationship and capital loss. At the 2nd step of feature selection, we observe mask 2 with the main features highlighted as 3 and 5 which are education and marital-status. Lastly, at the 3rd step of feature selection, we observe mask 3 with difficulties in discerning the important features.

Here are in-context examples for few-shot learning. 

The Poker Hand dataset is considered, which is a classification of poker hands into 10 distinct types based on attributes like rank of individual cards, suits, and card combinations. There are 50 test samples and 10 features. 

At the 0th step of feature selection, we observe mask 0 with the main feature highlighted as 3 which is the rank of the card 2. At the 1st step of feature selection, we observe mask 1 with the main feature highlighted as 7 which is the rank of card 3. At the 2nd step of feature selection, we observe mask 2 with the main feature highlighted as 6 which is the suit of card 3. Lastly, at the 3rd step of feature selection, we observe mask 3 with the main features highlighted as 3, 6, 7 and 10 which are the rank of cards 1 and 3 as well as the suits of cards 3 and 5.

Output: \{"Mask 0": "Initially, the rank of card 2 is recognized as a significant predictor in the classification of poker hands. The rank of individual cards seems crucial at this stage.", "Mask 1": "At this step, the rank of card 3 is highlighted, further emphasizing the importance of individual card ranks in determining poker hand types.", "Mask 2": "Here, besides the rank of card 3, the suit of the same card is considered, suggesting that both rank and suit play roles in classifying poker hands.", "Mask 3": "This step combines multiple features, both ranks (cards 1 and 3) and suits (cards 3 and 5), indicating that a combination of these factors provides a broader understanding of poker hand classifications.", "Aggregate": "The rank of cards, specifically card 2 and card 3, appears consistently across different masks, indicating their primary role in poker hand classification. Suits, particularly of card 3, also have importance but are introduced later in the feature selection. The combined importance of both ranks and suits in the final step suggests that while individual card ranks are pivotal in initial classifications, understanding the relationship between card ranks and their respective suits provides a more comprehensive insight into the poker hand types. The consistent presence of card 3's attributes (both rank and suit) underscores its pivotal role in determining poker hand categories."\}

The Forest Cover Type dataset is considered which is a classification of pixels into 7 forest cover types based on attributes such as elevation, aspect, slope, and more. There are 50 test samples and 54 features. 

At the 0th step of feature selection, we observe mask 0 with the main features highlighted as 24, 29 and 34 which are different soil types. At the 1st step of feature selection, we observe mask 1 with the main feature highlighted as 34 which is soil type. At the 2nd step of feature selection, we observe mask 2 with the main features highlighted as 10, 34 and 37 which are wilderness area and soil types. Lastly, at the 3rd step of feature selection, we observe mask 3 with the main features highlighted as 0 and 29 which are elevation, and soil type.

Output: \{"Mask 0": "The initial feature selection identifies that various soil types are significant predictors for the forest cover types. It appears that the characteristics of the soil play a vital role in determining the type of forest cover.", "Mask 1": "At this step, the model emphasizes even more on one particular soil type (34), suggesting it as a primary driver for the classification. This specific soil type may have unique characteristics making it a distinctive feature for certain forest cover types.", "Mask 2": "Here, besides the emphasis on soil type 34, another soil type (37) and wilderness area are considered. This suggests that both the nature of the terrain (wilderness) and the characteristics of the soil together contribute to the forest cover type.", "Mask 3": "This step introduces elevation as a significant predictor while still retaining a soil type. Elevation, combined with soil characteristics, might influence specific flora and subsequently determine forest cover types.", "Aggregate": "Soil type 34 is the most consistent feature across all the masks, appearing in every step, emphasizing its importance in the forest cover type classification. Another soil type, 29, is also consistent, appearing in both mask 0 and 3. The presence of different features across the steps, like wilderness area (mask 2) and elevation (mask 3), indicates that the landscape's nature and the physical height of the terrain are also important but might offer secondary or nuanced insights into forest cover types. The dominant predictor seems to be soil types, particularly soil type 34, suggesting that specific soil characteristics play a crucial role in determining forest cover types. However, aspects like elevation and the nature of the terrain (wilderness area) also contribute to the classification and shouldn't be overlooked."\}

\textbf{GPT-4 Output} \{"Mask 0": "Initially, only education is somewhat highlighted, suggesting its early recognition as a potential factor in predicting income levels.",
  
"Mask 1": "This step brings relationship and capital loss into focus, indicating their significant roles in determining income above \$50,000.",
  
"Mask 2": "Education and marital-status are emphasized, reinforcing the importance of educational background and marital status in income prediction.",
  
"Mask 3": "Difficulties in discerning important features suggest a complexity in determining clear predictors by this stage.",
  
"Aggregate": "Across the steps, education appears consistently, underscoring its critical role in income prediction. Relationship and capital loss also emerge as key factors, with their introduction in mask 1 highlighting their importance. The repeated emphasis on marital status alongside education in mask 2 further indicates the significance of personal and social factors in determining income levels. However, the challenges faced in mask 3 reveal the complexities in identifying definitive predictors as feature selection progresses."\}

\newpage
\section{Synthetic Dataset Experiments}
\label{appen:syn_dataset}

\begin{table}[htbp] 
    \centering
    \caption{AUC (classification task) mean and standard deviation (20 random seed trails) scores on 6 synthetic datasets from \citep{chen2018learning}. Results from the other models are reported from the TabNet paper which includes TabNet \citep{arik_pfister_2020} itself, INVASE \citep{yoon2018invase}, L2X \citep{chen2018learning} and Tree Ensembles \citep{tree_ensemble}. "Global" refers to the use of only globally salient features. "No Selection" refers to the use of all features without any feature selection. $\uparrow$ indicates that the higher the score, the better the performance. The corresponding InterpreTabNet $r_M$ values for each Syn1-Syn6 datasets are as follows: $5250$, $4750$, $4750$, $3250$, $5000$, $3000$. Red bolded numbers denote the best performance for each dataset.} 
    \label{tbl:exp-syn}
    \begin{threeparttable}
    {\small
        \begin{tabular}{lcccccc}
            \toprule[0.8pt]
            \multirow{2}{*}{Methods} & {\textbf{Syn1}} &{\textbf{Syn2}} & \textbf{Syn3} & {\textbf{Syn4}} &   {\textbf{Syn5}} & {\textbf{Syn6}} \\
            \cmidrule{2-7} 
            & AUC $\uparrow$ & AUC $\uparrow$ &  AUC $\uparrow$ &  AUC $\uparrow$  & AUC $\uparrow$ &  AUC $\uparrow$ \\
            \midrule 
            InterpreTabNet & $\redbf{.696}${\tiny$\pm \redbf{.005}$} & $.885${\tiny$\pm .003$} & $.899${\tiny$\pm .004$}  & $\redbf{.790}${\tiny$\pm\redbf{.008}$}  & $\redbf{.791}${\tiny$\pm\redbf{.006}$} & $\redbf{.880}${\tiny$\pm\redbf{.003}$} \\
            TabNet & $.682${\tiny$\pm .005$} & $\redbf{.892}${\tiny$\pm \redbf{.004}$} & $.897${\tiny$\pm .003$} & $.776${\tiny$\pm .017$} & $.789${\tiny$\pm .009$} & $.878${\tiny$\pm .004$}  \\
            INVASE & $.690${\tiny$\pm .006$} & $.877${\tiny$\pm .003$} & $\redbf{.902}${\tiny$\pm \redbf{.003}$} & $.787${\tiny$\pm .004$}    & $.784${\tiny$\pm .005$} & $.877${\tiny$\pm .003$} \\
            L2X & $.498${\tiny$\pm .005$} & $.823${\tiny$\pm .029$} & $.862${\tiny$\pm .009$} & $.678${\tiny$\pm .024$} & $.709${\tiny$\pm .008$} & $.827${\tiny$\pm .017$} \\
            Lasso-regularized & $.498${\tiny$\pm .006$} & $.555${\tiny$\pm .061$} & $.886${\tiny$\pm .003$} & $.512${\tiny$\pm .031$} & $.691${\tiny$\pm .024$} & $.727${\tiny$\pm .025$} \\
            Tree & $.574${\tiny$\pm .101$} & $.872${\tiny$\pm .003$} & $.899${\tiny$\pm .001$} & $.684${\tiny$\pm .017$} & $.741${\tiny$\pm .004$} & $.771${\tiny$\pm .031$} \\
            Global & $.686${\tiny$\pm .005$} & $.873${\tiny$\pm .003$} & $.900${\tiny$\pm .003$} & $.774${\tiny$\pm .006$} & $.784${\tiny$\pm .005$} & $.858${\tiny$\pm .004$} \\
            No Selection & $.578${\tiny$\pm .004$} & $.789${\tiny$\pm .003$} & $.854${\tiny$\pm .004$} & $.558${\tiny$\pm .021$} & $.662${\tiny$\pm .013$} & $.692${\tiny$\pm .015$} \\
		\bottomrule[1.0pt] 
		\end{tabular}
    }
    \end{threeparttable}
\end{table}

Synthetic datasets are crucial in determining whether the feature selection process is faithful towards the model's predictions. We use the same synthetic data generation model in TabNet \citep{arik_pfister_2020} and L2X \citep{chen2018learning}, as well as a sample size of 10K training and 10K testing. The datasets are formulated where a subset of features determines the prediction. For Syn1-Syn3, they only depend on their specified salient features i.e. Syn1 only depends on features 0 and 1. For Syn4-Syn6, the salient features depend on instance-wise features i.e. the output of Syn4, relies on either features 0-1 or features 2-5 depending on the value of feature 10.

Table \ref{tbl:exp-syn} indicates that InterpreTabNet with $r_M$ values $5250$, $4750$, $4750$, $3250$, $5000$, $3000$ for each Syn1-Syn6 datasets outperforms existing methods in 4 out of 6 of the synthetic datasets. More notably in all Syn4-Syn6 while remaining competitive in the other two Syn2 and Syn3 datasets. With respect to Syn1-Syn3, InterpreTabNet's competitiveness indicates that it is able to accurately achieve global feature selection. Additionally, for Syn4-Syn6, the SOTA performance indicate that InterpreTabnet is the best at performing global feature selection when instance-wise redundant features are removed.

\begin{figure}[htb!]
    \centering
    
    \begin{minipage}[b]{0.43\textwidth}
        \includegraphics[width=\textwidth, height=5cm]{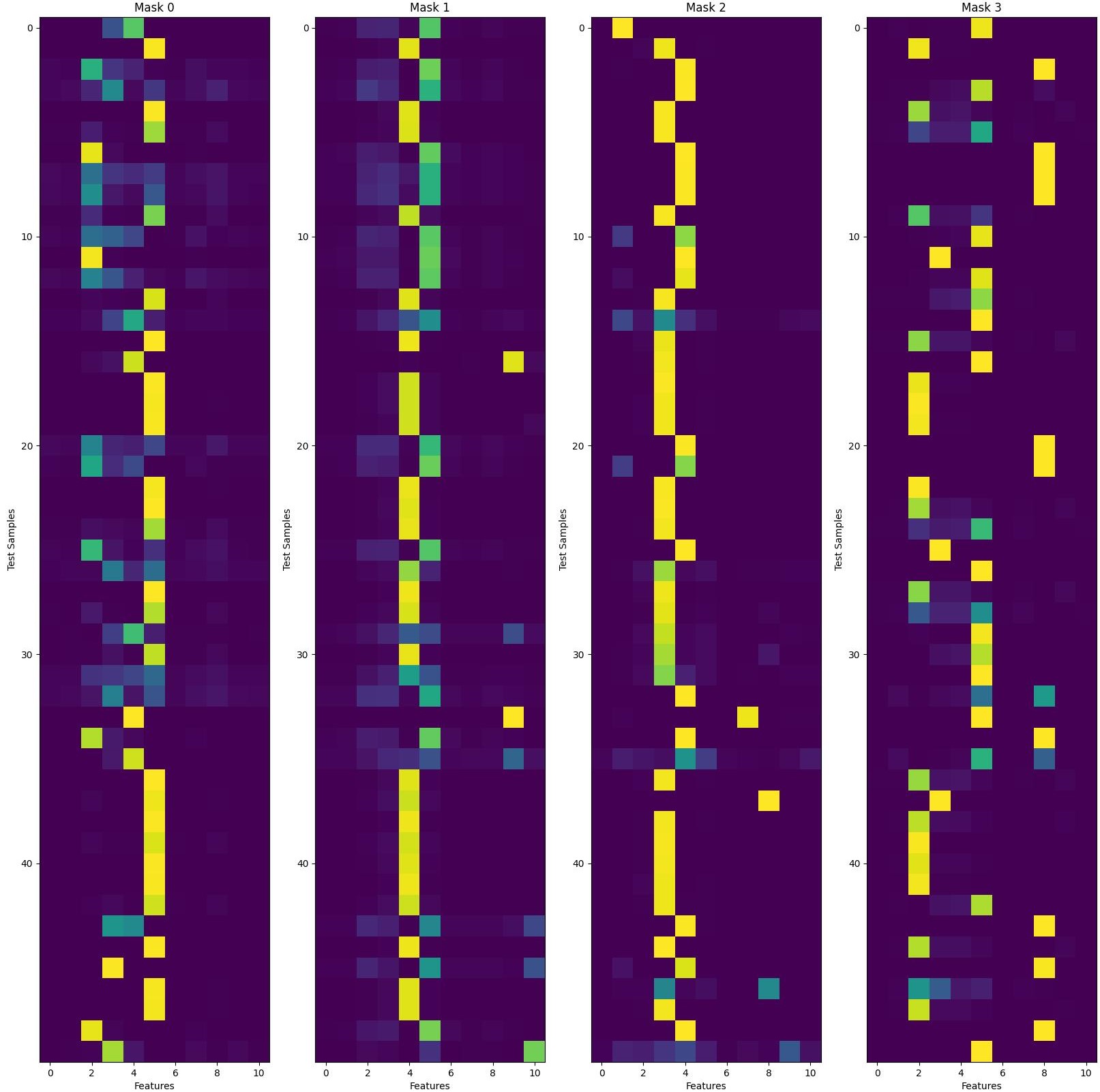}
        \caption*{(a) InterpreTabNet} 
    \end{minipage}
    \hfill
    \begin{minipage}[b]{0.43\textwidth}
        \includegraphics[width=\textwidth, height=5cm]{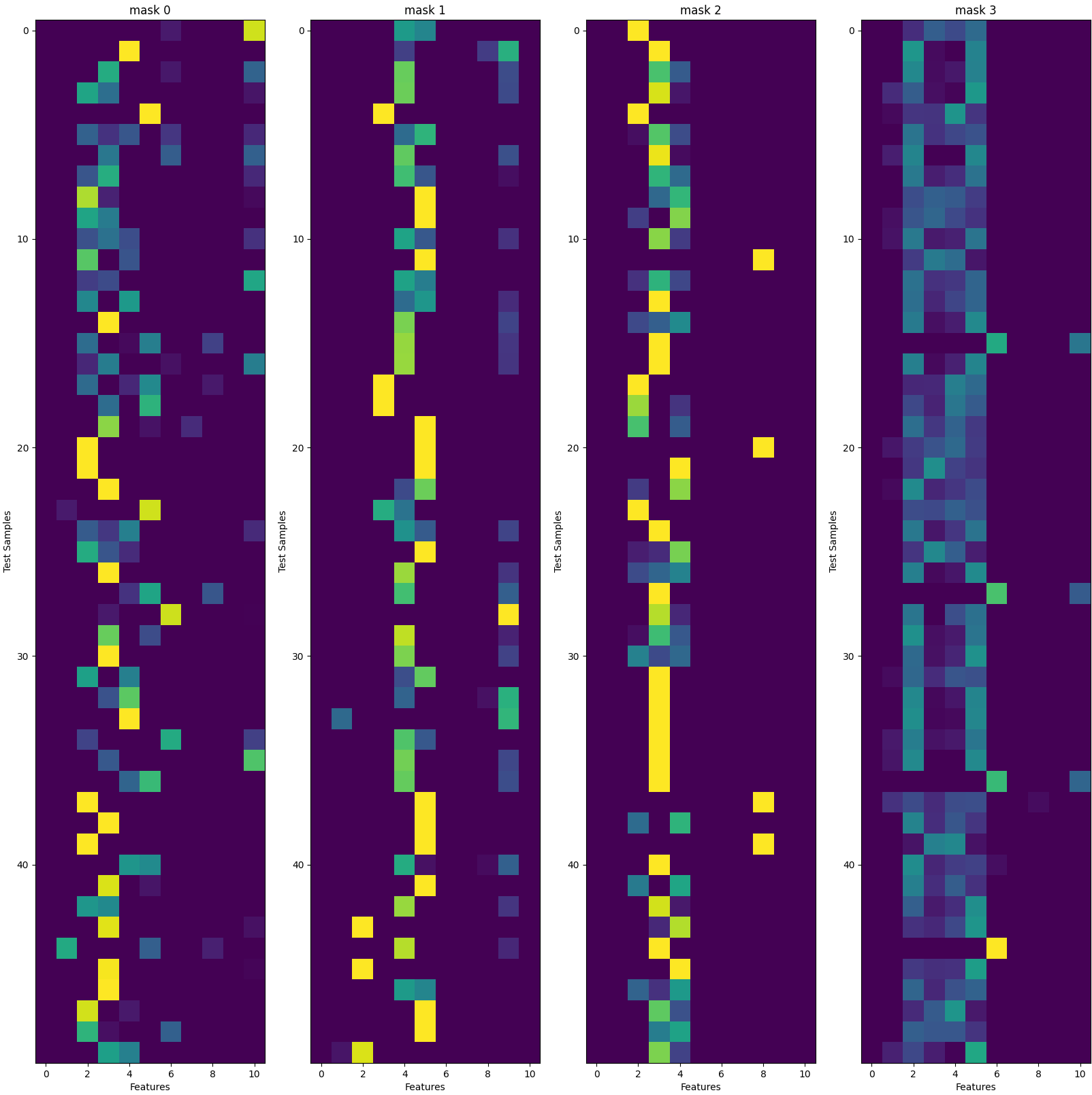}
        \caption*{(b) Original TabNet Model} 
    \end{minipage}
    
    \caption{(a) Best performing model using InterpreTabNet $r_M^* = 4,750$ with an AUC of 0.899 on Syn3 Dataset. (b) The baseline performance using the Original TabNet model, attaining an AUC of 0.897.}
    \label{fig:syn3_comparison}
\end{figure}

Syn3 uses features 2-5 to generate its predictions. As observed in Figure \ref{fig:syn3_comparison}, InterpreTabNet does indeed determine features 2-5 to be the most salient features as shown by the yellow bars in most of its masks. On the other hand, TabNet depicts lower salience levels when conducting its decision process for the prediction, indicating its uncertainty in reasoning when predicting the ground truth.

\newpage
\section{Justifications and Sanity Checks for Faithfulness and Robustness}
\subsection{InterpreTabNet Subgroup Feature Importance Analysis for Varying Subpopulation Characteristics}
\subsubsection*{Adult Census Income Mask Figures}
\label{appendix_mask_subgroups}

\begin{figure}[htb!]
    \centering
    
    \begin{minipage}[b]{0.45\textwidth}
        \includegraphics[width=\textwidth]{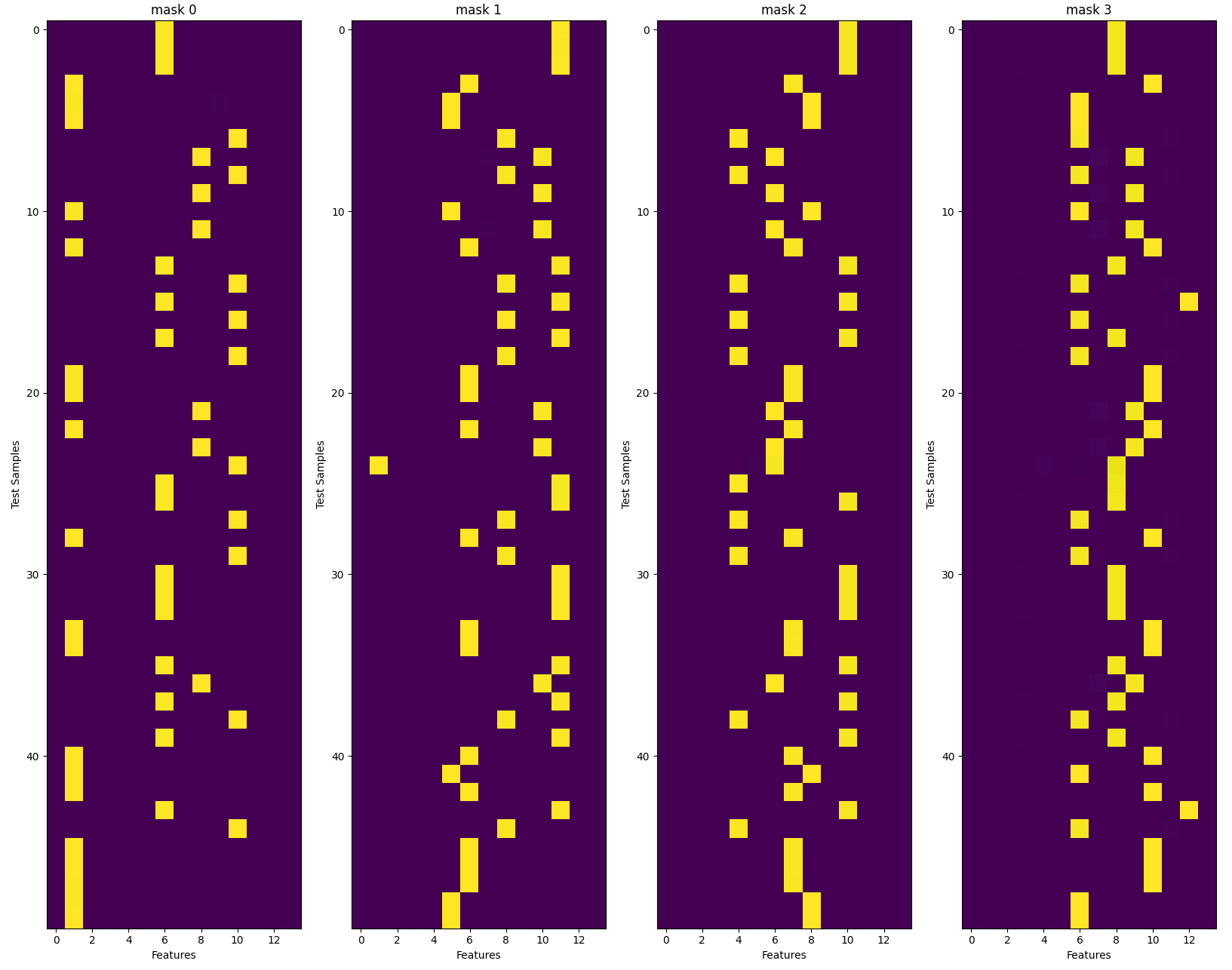}
        \caption*{(a) InterpreTabNet (Sex Subgroup: Male)} 
    \end{minipage}
    \hfill
    \begin{minipage}[b]{0.45\textwidth}
        \includegraphics[width=\textwidth]{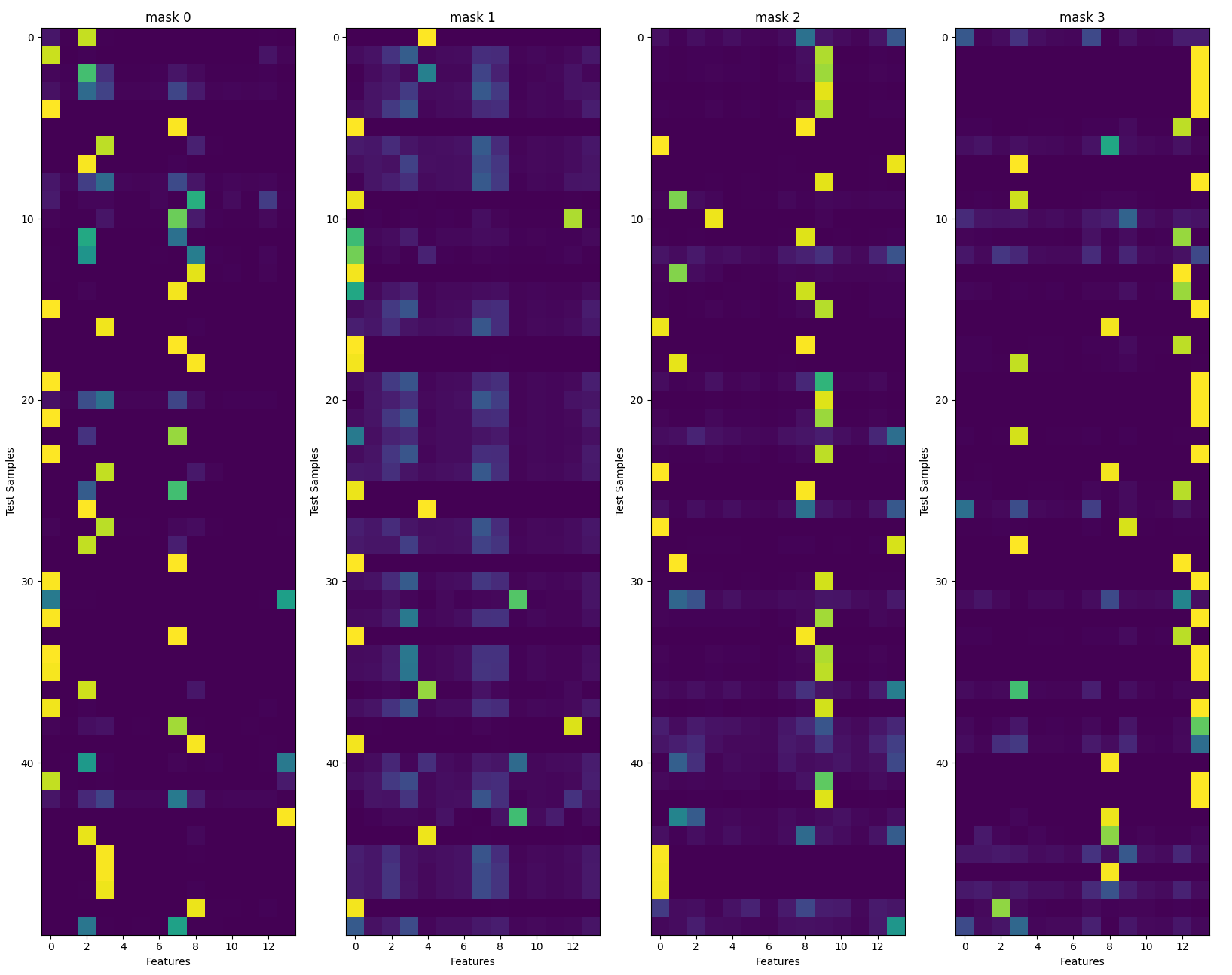}
        \caption*{(b) InterpreTabNet (Sex Subgroup: Female)} 
    \end{minipage}
    
    \caption{Feature importance visualization across different education subgroups in InterpreTabNet.}
    \label{fig:sex_subgroups_importance}
\end{figure}

To assess InterpreTabNet's robustness and reliability, we conducted a subgroup analysis, examining if salient feature identification differs across subgroups. We divided the adult census income dataset into male and female subpopulations, with the feature importance visualization depicted in Figure \ref{fig:sex_subgroups_importance}. For males, mask 0 highlighted work class, occupation, race, and capital gain (features 1, 6, 8, and 10, respectively) as most influential for income prediction. Conversely, mask 0 for females pinpointed age, education, and marital status (features 0, 3, and 5) as salient features contributing to prediction.

The salient features match well with known socio-economic factors affecting income. For males, occupation and capital gains highlight the importance of job roles and investments. For females, marital status notably influences income, likely due to the socio-economic dynamics and potential household income sharing. 

Subsequent masks for males consistently highlight features 6 (occupation), 8 (race), and 10 (capital gain). For females, mask 1 identifies feature 0 (age), mask 2 both feature 0 (age) and 8 (race), and mask 3 points feature 13 (native country). The minimum overlap in identified features suggests our model's robustness in identifying differences in important features between different subpopulations of the data.

\newpage
\subsection{Rationale of Maximising Feature Diversity}
\label{appen_diversity}

\begin{figure}[htb!]
    \centering
    
    \begin{minipage}[b]{0.4\textwidth}
        \includegraphics[width=\textwidth, height=7cm]{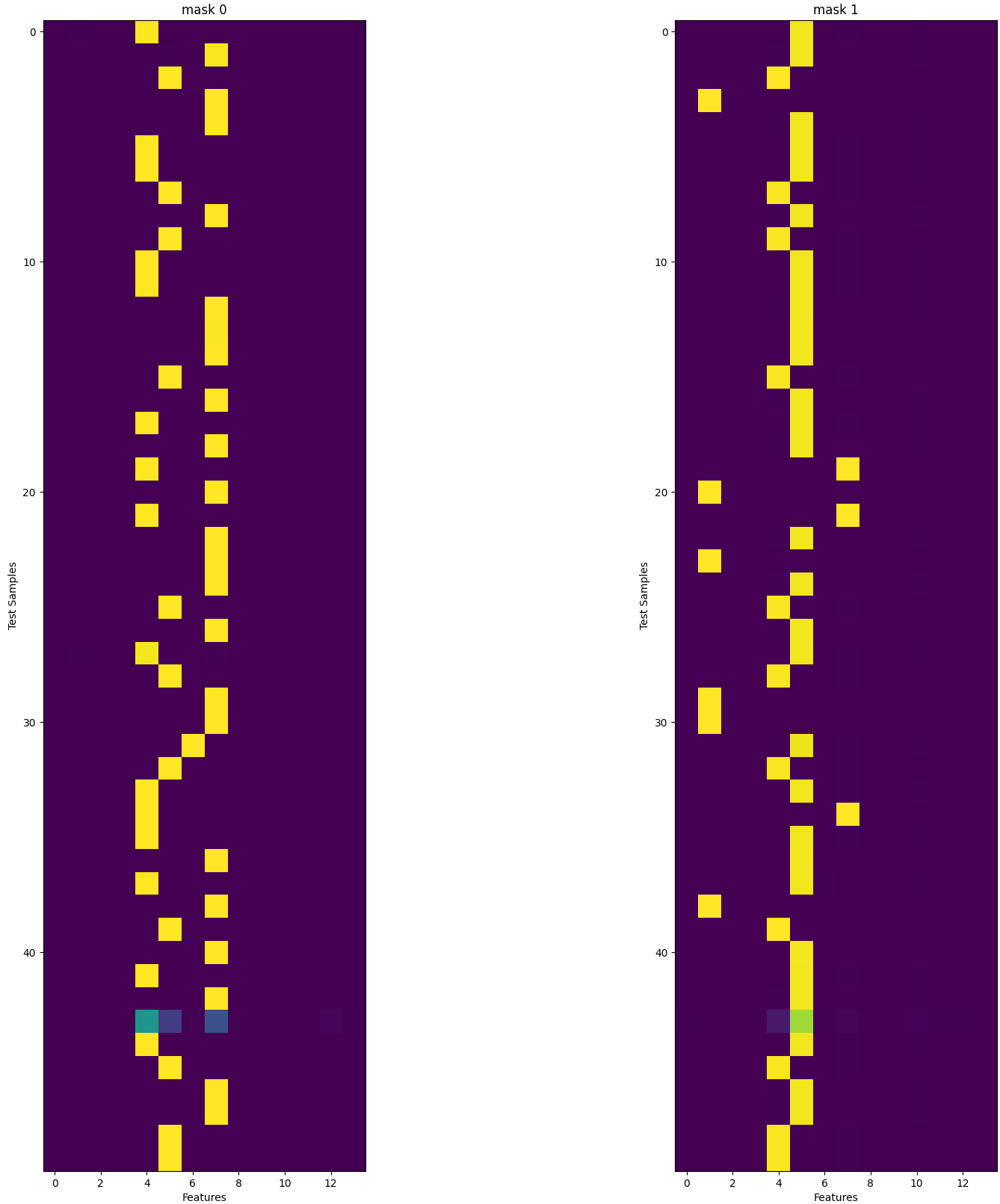}
        \caption*{(a) InterpreTabNet} 
    \end{minipage}
    \hfill
    \begin{minipage}[b]{0.4\textwidth}
        \includegraphics[width=\textwidth, height=7cm]{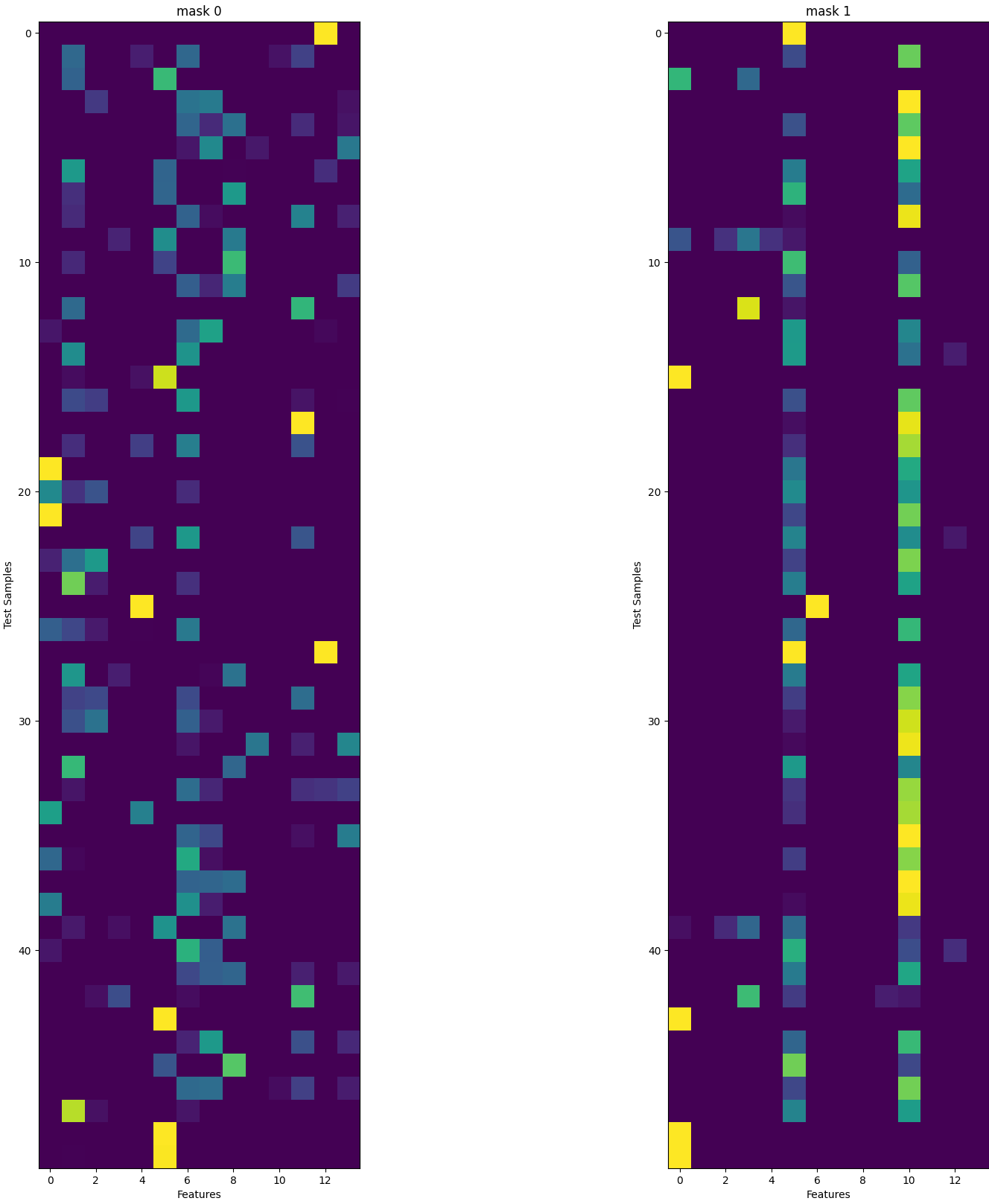}
        \caption*{(b) Original TabNet Model} 
    \end{minipage}
    
    \caption{(a) Best performing model using $r_M^* = 4$ for InterpreTabNet has an accuracy of 87.48 on the Adult Dataset. (b) The baseline performance using the Original TabNet model, attains an accuracy of 86.87\%.}
    \label{appen_fig_diversity}
\end{figure}

We observe in Figure \ref{appen_fig_diversity} that even when we select the least possible decision steps $N_{steps}=2$, TabNet utilizes every single feature aggregated across the two masks. On the other hand, InterpreTabNet selects clear salient features in its decision-making process.

\newpage
\subsection{Training Stability of InterpreTabNet vs. TabNet}
\label{append:training_stability}

\begin{figure*}[htb!]
    \centering
    \includegraphics[width=\textwidth, height=9cm]{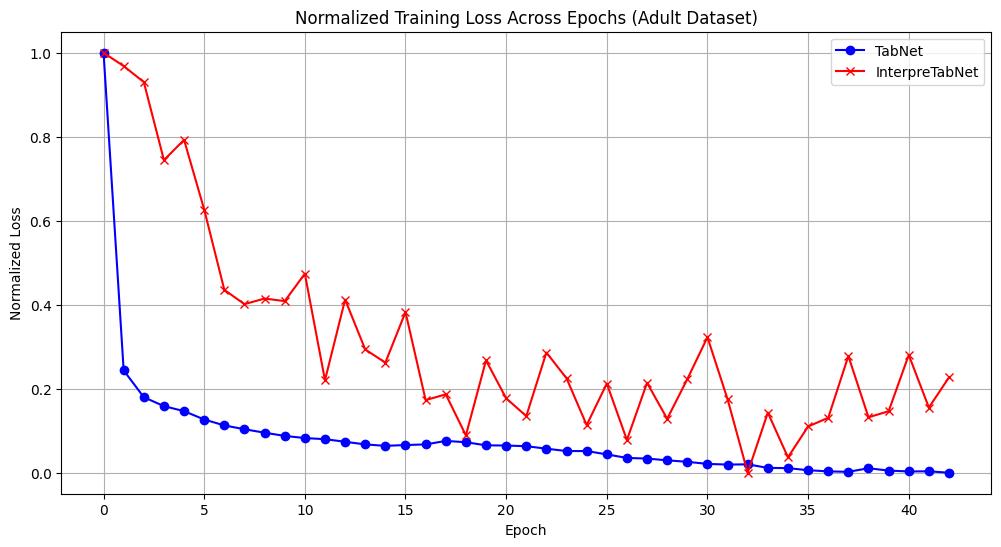}
    \caption{Normalized Training Loss of InterpreTabNet vs. TabNet for the Adult Income Dataset}
    \label{fig:training_stability}
\end{figure*}

Figure \ref{fig:training_stability} demonstrates the normalized training loss across epochs for two models: TabNet and InterpreTabNet. TabNet’s loss decreases rapidly in the initial epochs and stabilizes quickly. In comparison, InterpreTabNet exhibits higher variability in loss reduction across epochs, which suggests a more exploratory learning process to determine the salient features. Despite the fluctuations, InterpreTabNet’s loss shows a general downward trend, which is a positive indicator of learning. Overall, while TabNet appears to have a smoother training process, InterpreTabNet remains competitive.

\newpage

\subsection{Verifying Reliability of GPT-4's Analysis on Synthetic Dataset.}
\label{appendix_gpt4_check}

 We verify the reliability of GPT-4's analysis by prompting it to interpret synthetic datasets generated in the manner from \citep{yoon2018invase}. A synthetic dataset ($Syn1$) is considered where input features are generated from an 11-dimensional Gaussian distribution with no correlations across the features $(X\sim N(0,I))$. The label $Y$ is sampled as a Bernoulli random variable with $\mathbb{P}(Y=1|X)=\frac{1}{1+logit(X)}$, where $logit(X)$ is varied to create a synthetic dataset as follows: $Syn1 =exp(X_0, X_1)$. As shown in the GPT-4 output below, the analysis is indeed robust and trustworthy since it does not show any signs of hallucinations or mistakes hence, verifying the integrity of GPT-4's ability to interpret feature masks.

\textbf{Prompt.}
Conduct aggregate analysis on the description of the following feature mask. Please output ONLY a dictionary and no other natural language generation when generating the sentence as shown in the in-context example below. Please use single-word classification that encapsulates the meaning of the features if possible.

A synthetic dataset (Syn1) is considered where input features are generated from an 11-dimensional Gaussian distribution with no correlations across the features $X\sim N(0,I)$. The label Y is sampled as a Bernoulli random variable with $\mathbb{P}(Y=1|X)=\frac{1}{1+logit(X)}$, where logit(X) is varied to create a synthetic dataset as follows: Syn1 = $exp(X_0, X_1)$. Make sure to analyze the effect of Syn1 = $exp(X_0, X_1)$ in your aggregate analysis.

At the 0th step of feature selection, we observe mask 0 with the main feature highlighted as 0 which is $X_0$. At the 1st step of feature selection, we observe mask 1 with the main feature highlighted as 1 which is $X_1$. Lastly, at the 2nd step of feature selection, we observe mask 2 with the main features highlighted as 0 and 1 which are $X_1$ and $X_2$.

Here are in-context examples for few-shot learning. 

The Poker Hand dataset is considered, which is a classification of poker hands into 10 distinct types based on attributes like rank of individual cards, suits, and card combinations. There are 50 test samples and 10 features. 

At the 0th step of feature selection, we observe mask 0 with the main feature highlighted as 3 which is the rank of the card 2. At the 1st step of feature selection, we observe mask 1 with the main feature highlighted as 7 which is the rank of card 3. At the 2nd step of feature selection, we observe mask 2 with the main feature highlighted as 6 which is the suit of card 3. Lastly, at the 3rd step of feature selection, we observe mask 3 with the main features highlighted as 3, 6, 7 and 10 which are the rank of cards 1 and 3 as well as the suits of cards 3 and 5.

Output: \{"Mask 0": "Initially, the rank of card 2 is recognized as a significant predictor in the classification of poker hands. The rank of individual cards seems crucial at this stage.", "Mask 1": "At this step, the rank of card 3 is highlighted, further emphasizing the importance of individual card ranks in determining poker hand types.", "Mask 2": "Here, besides the rank of card 3, the suit of the same card is considered, suggesting that both rank and suit play roles in classifying poker hands.", "Mask 3": "This step combines multiple features, both ranks (cards 1 and 3) and suits (cards 3 and 5), indicating that a combination of these factors provides a broader understanding of poker hand classifications.", "Aggregate": "The rank of cards, specifically card 2 and card 3, appears consistently across different masks, indicating their primary role in poker hand classification. Suits, particularly of card 3, also have importance but are introduced later in the feature selection. The combined importance of both ranks and suits in the final step suggests that while individual card ranks are pivotal in initial classifications, understanding the relationship between card ranks and their respective suits provides a more comprehensive insight into the poker hand types. The consistent presence of card 3's attributes (both rank and suit) underscores its pivotal role in determining poker hand categories."\}

The Forest Cover Type dataset is considered which is a classification of pixels into 7 forest cover types based on attributes such as elevation, aspect, slope, and more. There are 50 test samples and 54 features. 

At the 0th step of feature selection, we observe mask 0 with the main features highlighted as 24, 29 and 34 which are different soil types. At the 1st step of feature selection, we observe mask 1 with the main feature highlighted as 34 which is soil type. At the 2nd step of feature selection, we observe mask 2 with the main features highlighted as 10, 34 and 37 which are wilderness area and soil types. Lastly, at the 3rd step of feature selection, we observe mask 3 with the main features highlighted as 0 and 29 which are elevation, and soil type.

Output: \{"Mask 0": "The initial feature selection identifies that various soil types are significant predictors for the forest cover types. It appears that the characteristics of the soil play a vital role in determining the type of forest cover.", "Mask 1": "At this step, the model emphasizes even more on one particular soil type (34), suggesting it as a primary driver for the classification. This specific soil type may have unique characteristics making it a distinctive feature for certain forest cover types.", "Mask 2": "Here, besides the emphasis on soil type 34, another soil type (37) and wilderness area are considered. This suggests that both the nature of the terrain (wilderness) and the characteristics of the soil together contribute to the forest cover type.", "Mask 3": "This step introduces elevation as a significant predictor while still retaining a soil type. Elevation, combined with soil characteristics, might influence specific flora and subsequently determine forest cover types.", "Aggregate": "Soil type 34 is the most consistent feature across all the masks, appearing in every step, emphasizing its importance in the forest cover type classification. Another soil type, 29, is also consistent, appearing in both mask 0 and 3. The presence of different features across the steps, like wilderness area (mask 2) and elevation (mask 3), indicates that the landscape's nature and the physical height of the terrain are also important but might offer secondary or nuanced insights into forest cover types. The dominant predictor seems to be soil types, particularly soil type 34, suggesting that specific soil characteristics play a crucial role in determining forest cover types. However, aspects like elevation and the nature of the terrain (wilderness area) also contribute to the classification and shouldn't be overlooked."\}

\textbf{GPT-4 Output.} \{"Mask 0": "Initially, the feature $X_0$ is identified as a significant predictor for the dataset. This step suggests the individual effect of $X_0$ on the label Y is crucial.",

"Mask 1": "At this step, feature $X_1$ is highlighted, indicating its individual significance in influencing the label Y, similar to $X_0$'s initial impact.",

"Mask 2": "Here, both features $X_0$ and $X_1$ are highlighted together, suggesting that the combined effect of these features significantly impacts the label Y, beyond their individual contributions.",

"Aggregate": "The progression of feature selection indicates a transition from individual to combined feature importance. Initially, $X_0$ and then $X_1$ are identified as significant on their own. However, by the final step, the combination of $X_0$ and $X_1$ is recognized as crucial, highlighting the importance of interactions between features in influencing the label Y. This suggests that while individual features have their significance, the interaction between $X_0$ and $X_1$, especially considering the synthetic dataset's nature (Syn1 = $exp(X_0, X_1)$), is vital for predicting Y. The synthetic dataset's structure emphasizes the multiplicative interaction between $X_0$ and $X_1$, illustrating the complex relationship between features in modeling."\}

\newpage

\subsection{Prompt Variability}
\label{appendix_prompt_variability}
\begin{table}[htb!]
\centering
\caption{GPT-4 Prompts in different formats}
\label{appen:prompt_format}
\footnotesize 
\begin{tabularx}{\columnwidth}{l*{1}{X}}
\toprule
\textbf{Section} & \textbf{Description} \\
\midrule
Goal \& Format & Generate a dictionary analysis of the feature masks described below. The output should consist solely of a dictionary, with no additional explanatory text. Keys should be the mask identifiers, and values should be lists containing single-word descriptors that capture the essence of the features identified at each step of feature selection. The descriptors should be as succinct as possible. \\
\addlinespace
Dataset Description & For the Adult Census Income dataset, which predicts whether an individual’s income exceeds \$50,000 per year based on 50 test samples and 14 features, the following observations were made:
 \\
\addlinespace
Mask Description & At the 0th step of feature selection, we observe mask 0 with the main features highlighted as 1, 5 and 7 which are workclass, marital status and relationship. At the 1st step of feature selection, we observe mask 0 with the main features highlighted as 1 and 5 which are workclass and marital status. At the 2nd step of feature selection, we observe mask 2 with the main features highlighted as 8, 10 and 12 which are race, capital gain and hours-per-week. Lastly, at the 3rd step of feature selection, we observe mask 3 with the main features highlighted as 1, 7 and 10 which are workclass, relationship and capital-gain. \\
\midrule
\textit{GPT-4 Output} & \textit{\{ "Mask 0": ["Employment", "Marital", "Relationship"], "Mask 1": ["Employment", "Marital"], "Mask 2": ["Ethnicity", "Income", "Worktime"], "Mask 3": ["Employment", "Relationship", "Income"] \}} \\
\bottomrule
\end{tabularx}
\end{table}

\begin{table}[htb!]
\centering
\caption{GPT-4 Prompts in varying detail levels}
\label{appen:prompt_detail}
\footnotesize 
\begin{tabularx}{\columnwidth}{l*{1}{X}}
\toprule
\textbf{Section} & \textbf{Description} \\
\midrule
Less Detailed Goal & Generate a dictionary with key insights from feature masks. Output a dictionary with keys as 'Mask <step>' and values as single-word descriptors of the most significant feature(s) at each step. The descriptors should summarize the feature's impact on predicting an individual’s income in the Adult Census Income dataset. \\
\addlinespace
Moderately Detailed Goal & Construct a dictionary summarizing the significance of feature masks in predicting income levels in the Adult Census Income dataset. For each mask, output a dictionary with a list of descriptors, each a concise term that captures the importance of the features at that step of selection.
 \\
\addlinespace
More Detailed Goal & Create a detailed dictionary explaining the role of each feature mask in the context of the Adult Census Income dataset. Each dictionary entry should include the step number as the key and a list of phrases as the value, capturing the nuanced significance of each feature in predicting whether an individual’s income exceeds \$50,000 per year. \\
\midrule
\textit{GPT-4 Less Detailed Output} & \textit{\{ "Mask 0": ["Primary"], "Mask 1": ["Secondary"], "Mask 2": ["Tertiary"], "Mask 3": ["Quaternary"] \}} \\
\addlinespace
\textit{GPT-4 Moderately Detailed Output} & \textit{\{"Mask 0": ["Work", "Family", "Relations"], "Mask 1": ["Work", "Family"], "Mask 2": ["Ethnicity", "Investment", "Labor"], "Mask 3": ["Work", "Relations", "Investment"]\}} \\
\addlinespace
\textit{GPT-4 More Detailed Output} & \textit{\{"Mask 0": ["Employment class", "Marital bonds", "Social ties"], "Mask 1": ["Employment status", "Marital status"], "Mask 2": ["Ethnicity diversity", "Capital growth", "Work duration"], "Mask 3": ["Employment nature", "Social relationship", "Capital investment"]\}} \\
\addlinespace
\bottomrule
\end{tabularx}
\end{table}

\newpage
\subsection{Prompting GPT-4 with Persona of a Domain Expert}
\label{domain_expert}

We conditioned GPT-4 on a prompt where the model was asked to comment on the feature masks as if they were an expert in the broad topic that the predictive model was built for i.e. "Assume that you are an Economist - Adult Dataset", "Assume that you are a Physician - Diabetes Dataset". The goal of this experiment is to give the reader a qualitative assessment of the variability in the interpretations based on what an expert in the domain might care about. 

We summarize the variation of results in Sections \ref{economist} and \ref{diabetes} — what we find is that the expert interpretations do not vary significantly compared to the original interpretations given different "expert" prompts. Therefore, we can anticipate that the interpretations are "robust".

\subsubsection*{Adult Dataset - Economist}
\label{economist}

\begin{center}
\begin{longtable}{cp{10cm}} 
\toprule
 \textbf{Dataset \& Expert Persona} & \textbf{Output from GPT-4} \\  
\midrule
Adult Dataset, Economist &  \{
"Mask 0": "Initial selection highlights workclass, marital status, and relationship as key predictors of income levels, indicating that employment type, marital situation, and family roles are crucial for income classification.",

"Mask 1": "This step maintains focus on workclass and marital status, reaffirming their importance in predicting income levels and suggesting a consistent pattern in the significance of employment type and marital situation.",

"Mask 2": "The focus shifts to race, capital gain, and hours-per-week, introducing new dimensions such as ethnic background, financial investment returns, and work hours as significant factors influencing income.",

"Mask 3": "Reintroduces workclass and relationship along with capital gain, indicating that while employment type and family roles remain pertinent, financial investments emerge as a critical predictor.",

"Aggregate": "Workclass and marital status appear as consistent predictors across multiple masks, emphasizing the role of employment type and marital situation in income determination. Relationship and capital gain also emerge as significant, with the latter highlighting financial investments' importance. The inclusion of race and hours-per-week in one mask suggests a broader perspective, considering ethnic background and work dedication. Overall, employment type, marital status, financial investments, and to a lesser extent, family roles, ethnic background, and work hours, are key to predicting income levels."
\} 
\\
\midrule
\newpage
\midrule
Adult Dataset, No Persona &  \{"Mask 0": "This suggests that early in the feature selection process, the model finds that work-related factors (like the type of work) and personal relationships (both marital status and other relationships) are significant predictors for income exceeding \$50,000.", 
 
 "Mask 1": "Here, the model emphasizes even more on work nature and marital status, suggesting these two as the primary drivers for the prediction. Relationship as a feature has been excluded in this step, which might indicate a possible correlation with marital status.", 
 
 "Mask 2": "At this step, the model seems to be looking beyond just personal and work nature to more financial and demographic aspects. Capital gain, which is directly related to financial health, is considered along with hours-per-week, potentially pointing to a relationship between longer working hours and higher income. The inclusion of race might indicate underlying societal or economic factors at play.", 
 
 "Mask 3": "This combination brings back features from both the initial and second masks. Workclass and relationship are revisited, which further underscores their importance. Capital gain remains consistent from the second step, emphasizing its importance as a predictor for income.", 
 
 "Aggregate": "The features that consistently appear across multiple masks are: workclass (masks 0, 1, and 3), relationship (masks 0 and 3), and capital gain (masks 2 and 3). This highlights that these features play a pivotal role in predicting whether an individual’s income exceeds \$50,000 per year. Marital status is seen in the initial stages but disappears later, while hours-per-week and race come into play at the second step. This could indicate that while some features provide foundational predictive power, others may provide more nuanced insight or refinement to the model's predictions. Work-related aspects, personal relationships, and financial metrics seem to be the main predictors of an individual's income level in this dataset. It would be crucial to further analyze these features' exact impacts and understand any potential correlations among them for a more comprehensive understanding."\}
\\
\bottomrule
\end{longtable}
\end{center}

\textbf{Prompt} \textit{Assume that you are an Economist.} Conduct aggregate analysis on the description of the following feature mask. Please output ONLY a dictionary and no other natural language generation when generating the sentence as shown in the in-context example below. Please use single-word classification that encapsulates the meaning of the features if possible.

The Adult Census Income dataset is considered which predicts whether an individual’s income exceeds \$50,000 per year. There are 50 test samples and 14 features. 

At the 0th step of feature selection, we observe mask 0 with the main features highlighted as 1, 5 and 7 which are workclass, marital status and relationship. At the 1st step of feature selection, we observe mask 1 with the main features highlighted as 1 and 5 which are workclass and marital status. At the 2nd step of feature selection, we observe mask 2 with the main features highlighted as 8, 10 and 12 which are race, capital gain and hours-per-week. Lastly, at the 3rd step of feature selection, we observe mask 3 with the main features highlighted as 1, 7 and 10 which are workclass, relationship and capital-gain.

Here are in-context examples for few-shot learning. 

The Poker Hand dataset is considered, which is a classification of poker hands into 10 distinct types based on attributes like rank of individual cards, suits, and card combinations. There are 50 test samples and 10 features. 

At the 0th step of feature selection, we observe mask 0 with the main feature highlighted as 3 which is the rank of the card 2. At the 1st step of feature selection, we observe mask 1 with the main feature highlighted as 7 which is the rank of card 3. At the 2nd step of feature selection, we observe mask 2 with the main feature highlighted as 6 which is the suit of card 3. Lastly, at the 3rd step of feature selection, we observe mask 3 with the main features highlighted as 3, 6, 7 and 10 which are the rank of cards 1 and 3 as well as the suits of cards 3 and 5.

Output: \{"Mask 0": "Initially, the rank of card 2 is recognized as a significant predictor in the classification of poker hands. The rank of individual cards seems crucial at this stage.", "Mask 1": "At this step, the rank of card 3 is highlighted, further emphasizing the importance of individual card ranks in determining poker hand types.", "Mask 2": "Here, besides the rank of card 3, the suit of the same card is considered, suggesting that both rank and suit play roles in classifying poker hands.", "Mask 3": "This step combines multiple features, both ranks (cards 1 and 3) and suits (cards 3 and 5), indicating that a combination of these factors provides a broader understanding of poker hand classifications.", "Aggregate": "The rank of cards, specifically card 2 and card 3, appears consistently across different masks, indicating their primary role in poker hand classification. Suits, particularly of card 3, also have importance but are introduced later in the feature selection. The combined importance of both ranks and suits in the final step suggests that while individual card ranks are pivotal in initial classifications, understanding the relationship between card ranks and their respective suits provides a more comprehensive insight into the poker hand types. The consistent presence of card 3's attributes (both rank and suit) underscores its pivotal role in determining poker hand categories."\}

The Forest Cover Type dataset is considered which is a classification of pixels into 7 forest cover types based on attributes such as elevation, aspect, slope, and more. There are 50 test samples and 54 features. 

At the 0th step of feature selection, we observe mask 0 with the main features highlighted as 24, 29 and 34 which are different soil types. At the 1st step of feature selection, we observe mask 1 with the main feature highlighted as 34 which is soil type. At the 2nd step of feature selection, we observe mask 2 with the main features highlighted as 10, 34 and 37 which are wilderness area and soil types. Lastly, at the 3rd step of feature selection, we observe mask 3 with the main features highlighted as 0 and 29 which are elevation, and soil type.

Output: \{"Mask 0": "The initial feature selection identifies that various soil types are significant predictors for the forest cover types. It appears that the characteristics of the soil play a vital role in determining the type of forest cover.", "Mask 1": "At this step, the model emphasizes even more on one particular soil type (34), suggesting it as a primary driver for the classification. This specific soil type may have unique characteristics making it a distinctive feature for certain forest cover types.", "Mask 2": "Here, besides the emphasis on soil type 34, another soil type (37) and wilderness area are considered. This suggests that both the nature of the terrain (wilderness) and the characteristics of the soil together contribute to the forest cover type.", "Mask 3": "This step introduces elevation as a significant predictor while still retaining a soil type. Elevation, combined with soil characteristics, might influence specific flora and subsequently determine forest cover types.", "Aggregate": "Soil type 34 is the most consistent feature across all the masks, appearing in every step, emphasizing its importance in the forest cover type classification. Another soil type, 29, is also consistent, appearing in both mask 0 and 3. The presence of different features across the steps, like wilderness area (mask 2) and elevation (mask 3), indicates that the landscape's nature and the physical height of the terrain are also important but might offer secondary or nuanced insights into forest cover types. The dominant predictor seems to be soil types, particularly soil type 34, suggesting that specific soil characteristics play a crucial role in determining forest cover types. However, aspects like elevation and the nature of the terrain (wilderness area) also contribute to the classification and shouldn't be overlooked."\}

\newpage
\subsubsection*{Diabetes Dataset - Physician}
\label{diabetes}

\begin{center}
\begin{longtable}{cp{10cm}} 
\toprule
 \textbf{Dataset \& Expert Persona} & \textbf{Output from GPT-4} \\  
\midrule
Diabetes Dataset, Physician & \{"Mask 0": "This initial selection highlights the importance of discharge disposition, the number of diagnoses, and specific medications (glyburide and citoglipton) in predicting readmissions. It suggests that patient outcomes and the specific treatment protocols, including medication use, play a crucial role in the likelihood of a patient being readmitted.",

"Mask 1": "At this stage, the model focuses on the admission type and discharge disposition, indicating the significance of the nature of admission and the patient’s discharge status in predicting readmissions. This could reflect how initial hospitalization circumstances and post-discharge plans impact patient outcomes.",

"Mask 2": "Here, the inclusion of admission type, number of diagnoses, and metformin (another medication) suggests a continued emphasis on the circumstances of admission and the complexity of the patient's health condition, as well as the role of specific diabetes medication in influencing readmission risks.",

"Mask 3": "By focusing on the number of diagnoses and medications (glipizide and metformin), this step underscores the critical nature of the patient’s health status and the treatment of diabetes as key factors in predicting hospital readmissions. It highlights the significance of managing complex health conditions and diabetes treatment.",

"Aggregate": "Across the selection process, the consistent features are the number of diagnoses and the emphasis on specific diabetes medications (metformin appearing in masks 2 and 3, and other diabetes-related medications in masks 0 and 3). This indicates a strong relationship between the complexity of a patient's health (as represented by the number of diagnoses) and the treatment of diabetes in predicting readmissions. Discharge disposition and admission type are also significant but might offer insights into the logistical aspects of care that affect readmission risks. The aggregate analysis underscores the complexity of managing diabetes within the hospital setting and its impact on readmission rates, highlighting the importance of both clinical and administrative factors."\}
\\
\midrule
\newpage
\midrule
Diabetes Dataset, No Persona &  \{
"Mask 0": "In the initial step, the model identifies discharge disposition id, number of diagnoses, and specific medications (glyburide, and citoglipton) as key features. This suggests a combination of administrative data (discharge process), clinical data (diagnoses), and prescribed medications play a significant role in predicting patient readmission.",

"Mask 1": "Here, the model focuses on the patient’s admission type and discharge disposition, emphasizing the importance of the hospital admission and discharge process in determining the likelihood of readmission. The absence of clinical and medication features in this step indicates a potential higher influence of administrative processes at this stage of feature selection.",

"Mask 2": "The model reintroduces clinical data (number of diagnoses) and adds a new medication (metformin), while retaining admission type id. This combination underscores a nuanced interplay between the nature of hospital admission, the complexity of patient’s health conditions, and specific treatments in predicting readmissions.",

"Mask 3": "This step focuses on the number of diagnoses and two medications (glipizide and metformin). The consistency of number of diagnoses across masks highlights its pivotal role, while the introduction of glipizide and retention of metformin point towards an evolving emphasis on medication in conjunction with clinical data.",

"Aggregate": "Number diagnoses (masks 0, 2, and 3) and discharge disposition id (masks 0 and 1) are recurrent, underscoring their persistent importance in predicting readmissions. Admission type id also appears consistently (masks 1 and 2), linking the administrative process tightly with readmission predictions. The changing set of medications across the masks (glyburide, citoglipton, metformin, glipizide) points to the dynamic nature of treatment's role in the prediction, warranting a closer look into specific medication combinations and their interactions with clinical and administrative data. The aggregate analysis reveals a complex interplay between administrative processes, clinical data, and medication in predicting patient readmissions, suggesting a multifaceted approach for model refinement and interpretation."
\}
\\
\bottomrule
\end{longtable}
\end{center}

\textbf{Prompt} \textit{Assume you are a Physician.} Conduct aggregate analysis on the description of the following feature masks. Start off with an analysis of the individual masks, followed by an aggregate analysis of all masks combined. Please format the output into a dictionary as shown in the in-context examples. The output should only contain the formatted output, no other natural language generation is required.

The Diabetes 130-US hospitals for years 1999-2008 dataset is considered, which is a classification of patient encounters into readmitted or not readmitted categories based on attributes like the number of laboratory tests performed, the number of medications prescribed, diagnoses, and other clinical and administrative data. There are 50 test samples and 50 features.

At the 0th step of feature selection, we observe mask 0 with the main features highlighted as 7, 21, 31, and 40 which are discharge disposition id, number diagnoses, glyburide, and citoglipton. At the 1st step of feature selection, we observe mask 1 with the main feature highlighted as 6 and 7 which are admission type id and discharge disposition id. At the 2nd step of feature selection, we observe mask 2 with the main features highlighted as 6, 21, and 45 which are admission type id, number diagnoses, and metformin. Lastly, at the 3rd step of feature selection, we observe mask 3 with the main features highlighted as 21, 30, 45 which are number diagnoses, glipizide, and metformin.

Here are in-context examples for few-shot learning. 

The Adult Census Income dataset is considered which predicts whether an individual’s income exceeds \$50,000 per year. There are 50 test samples and 14 features. At the 0th step of feature selection, we observe mask 0 with the main features highlighted as 1, 5 and 7 which are workclass, marital status and relationship. At the 1st step of feature selection, we observe mask 0 with the main features highlighted as 1 and 5 which are workclass and marital status. At the 2nd step of feature selection, we observe mask 2 with the main features highlighted as 8, 10 and 12 which are race, capital gain and hours-per-week. Lastly, at the 3rd step of feature selection, we observe mask 3 with the main features highlighted as 1, 7 and 10 which are workclass, relationship and capital-gain.

Output: \{"Mask 0": "This suggests that early in the feature selection process, the model finds that work-related factors (like the type of work) and personal relationships (both marital status and other relationships) are significant predictors for income exceeding \$50,000.", "Mask 1": "Here, the model emphasizes even more on work nature and marital status, suggesting these two as the primary drivers for the prediction. Relationship as a feature has been excluded in this step, which might indicate a possible correlation with marital status.", "Mask 2": "At this step, the model seems to be looking beyond just personal and work nature to more financial and demographic aspects. Capital gain, which is directly related to financial health, is considered along with hours-per-week, potentially pointing to a relationship between longer working hours and higher income. The inclusion of race might indicate underlying societal or economic factors at play.", "Mask 3": "This combination brings back features from both the initial and second masks. Workclass and relationship are revisited, which further underscores their importance. Capital gain remains consistent from the second step, emphasizing its importance as a predictor for income.", "Aggregate": "The features that consistently appear across multiple masks are: workclass (masks 0, 1, and 3), relationship (masks 0 and 3), and capital gain (masks 2 and 3). This highlights that these features play a pivotal role in predicting whether an individual’s income exceeds \$50,000 per year. Marital status is seen in the initial stages but disappears later, while hours-per-week and race come into play at the second step. This could indicate that while some features provide foundational predictive power, others may provide more nuanced insight or refinement to the model's predictions. Work-related aspects, personal relationships, and financial metrics seem to be the main predictors of an individual's income level in this dataset. It would be crucial to further analyze these features' exact impacts and understand any potential correlations among them for a more comprehensive understanding."\}

The Forest Cover Type dataset is considered which is a classification of pixels into 7 forest cover types based on attributes such as elevation, aspect, slope, and more. There are 50 test samples and 54 features. 

At the 0th step of feature selection, we observe mask 0 with the main features highlighted as 24, 29 and 34 which are different soil types. At the 1st step of feature selection, we observe mask 1 with the main feature highlighted as 34 which is soil type. At the 2nd step of feature selection, we observe mask 2 with the main features highlighted as 10, 34 and 37 which are wilderness area and soil types. Lastly, at the 3rd step of feature selection, we observe mask 3 with the main features highlighted as 0 and 29 which are elevation, and soil type.

Output: \{"Mask 0": "The initial feature selection identifies that various soil types are significant predictors for the forest cover types. It appears that the characteristics of the soil play a vital role in determining the type of forest cover.", "Mask 1": "At this step, the model emphasizes even more on one particular soil type (34), suggesting it as a primary driver for the classification. This specific soil type may have unique characteristics making it a distinctive feature for certain forest cover types.", "Mask 2": "Here, besides the emphasis on soil type 34, another soil type (37) and wilderness area are considered. This suggests that both the nature of the terrain (wilderness) and the characteristics of the soil together contribute to the forest cover type.", "Mask 3": "This step introduces elevation as a significant predictor while still retaining a soil type. Elevation, combined with soil characteristics, might influence specific flora and subsequently determine forest cover types.", "Aggregate": "Soil type 34 is the most consistent feature across all the masks, appearing in every step, emphasizing its importance in the forest cover type classification. Another soil type, 29, is also consistent, appearing in both mask 0 and 3. The presence of different features across the steps, like wilderness area (mask 2) and elevation (mask 3), indicates that the landscape's nature and the physical height of the terrain are also important but might offer secondary or nuanced insights into forest cover types. The dominant predictor seems to be soil types, particularly soil type 34, suggesting that specific soil characteristics play a crucial role in determining forest cover types. However, aspects like elevation and the nature of the terrain (wilderness area) also contribute to the classification and shouldn't be overlooked."\}

\newpage
\subsection{GPT-4 Analysis without InterpreTabNet}
\label{appendix_gpt_without_interpretabnet}
\begin{table}[htb!]
\centering
\caption{GPT-4 Analysis without InterpreTabNet}
\label{appen:gpt_no_intabnet}
\footnotesize 
\begin{tabularx}{\columnwidth}{XX}
\toprule
\textbf{Prompts} & \textbf{Outputs} \\
\midrule
Identify the most important features of the following dataset and conduct aggregate text analysis on it. Please output ONLY a dictionary and no other natural language generation when generating the sentence. The Adult Census Income dataset is considered which predicts whether an individual’s income exceeds \$50,000 per year. The features are as follows: Age. Workclass. Final Weight. Education. Education Number of Years. Marital-status. Occupation. Relationship. Race. Sex. Capital-gain. Capital-loss. Hours-per-week. Native-country. & \{ "Important Features": { "Numerical": ["Age", "Final Weight", "Education Number of Years", "Capital-gain", "Capital-loss", "Hours-per-week"], "Categorical": ["Workclass", "Education", "Marital-status", "Occupation", "Relationship", "Race", "Sex", "Native-country"] }, "Aggregate Text Analysis": \{ "Income Prediction": "The Adult Census Income dataset is used to predict if an individual's income exceeds \$50,000 per year based on various demographic and financial features." \} \} \\
\addlinespace
Indicate the most important features of the following dataset. Please output ONLY a dictionary and no other natural language generation when generating the sentence. The Adult Census Income dataset is considered which predicts whether an individual’s income exceeds \$50,000 per year. The features are as follows: Age. Workclass. Final Weight. Education. Education Number of Years. Marital-status. Occupation. Relationship. Race. Sex. Capital-gain. Capital-loss. Hours-per-week. Native-country. & \{ "Important Features": ["Age", "Education Number of Years", "Marital-status", "Occupation", "Relationship", "Capital-gain", "Capital-loss", "Hours-per-week" ] \} \\
\addlinespace
\bottomrule
\end{tabularx}
\end{table}

\newpage
\subsection{Feature Mask Definition Check}
\label{appendix_feature_mask_definition_check}
\begin{figure}[htb!]
    \centering
    \includegraphics[width=0.75\textwidth]{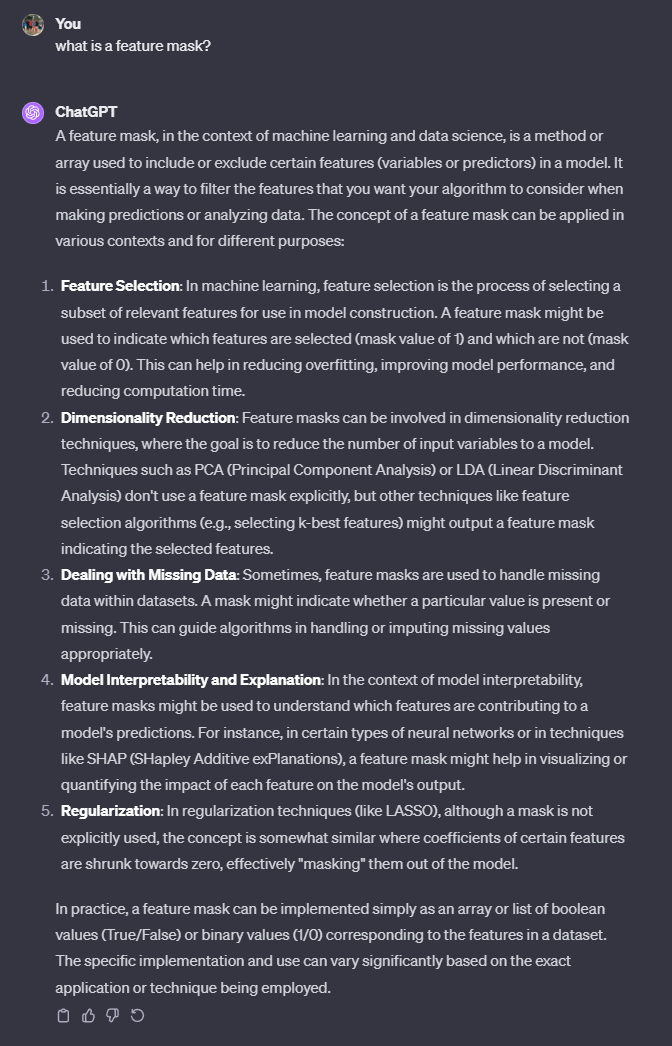}
    \caption{Feature Mask Definition Check}
    \label{fig:mask_definition_check}
\end{figure}

\end{document}